\documentclass[default, iicol]{sn-jnl}%

\usepackage{graphicx}%
\usepackage{multirow}%
\usepackage{amsmath,amssymb,amsfonts}%
\usepackage{amsthm}%
\usepackage{mathrsfs}%
\usepackage[title]{appendix}%
\usepackage{xcolor}%
\usepackage{textcomp}%
\usepackage{manyfoot}%
\usepackage{booktabs}%
\usepackage{algorithm}%
\usepackage{algorithmicx}%
\usepackage{algpseudocode}%
\usepackage{listings}%
\usepackage{bbding}
\usepackage{comment}
\usepackage{float}
\usepackage{makecell}
\usepackage{ragged2e}

\usepackage{array}

\usepackage{booktabs}
\usepackage{color}
\usepackage{pifont}
\usepackage{subcaption}

\usepackage{notoccite}

\definecolor{responsecolor}{RGB}{255, 0, 0}

\def\ie{\emph{i.e.}}
\def\eg{\emph{e.g.}}
\newcommand{\etal}{\textit{et al}.}

\theoremstyle{thmstyleone}%

\theoremstyle{thmstyletwo}%

\theoremstyle{thmstylethree}%

\title[MMInvertFill]{High-Fidelity Image Inpainting with Multimodal Guided GAN Inversion}

\author[1]{\fnm{Libo} \sur{Zhang}}\email{libo@iscas.ac.cn}\equalcont

\author[1]{\fnm{Yongsheng} \sur{Yu}}\email{yuyongsheng19@mails.ucas.ac.cn}\equalcont

\author[2]{\fnm{Jiali} \sur{Yao}}\email{yaojiali3000@hotmail.com}

\author*[3]{\fnm{Heng} \sur{Fan}}\email{heng.fan@unt.edu}

\affil[1]{Institute of Software Chinese Academy of Sciences, Beijing, China}
\affil[2]{Hangzhou Institute for Advanced Study, University of Chinese Academy of Sciences, Hangzhou, China}
\affil[3]{Department of Computer Science and Engineering, University of North Texas, Denton, TX, USA}

\begin{document}
\removelastskip\vskip36pt\vskip0pt

\abstract{
Generative Adversarial Network (GAN) inversion have demonstrated excellent performance in image inpainting that aims to restore lost or damaged image texture using its unmasked content. Previous GAN inversion-based methods usually utilize well-trained GAN models as effective priors to generate the realistic regions for missing holes. Despite excellence, they ignore a hard constraint that the unmasked regions in the input and the output should be the same, resulting in a gap between GAN inversion and image inpainting and thus degrading the performance. Besides, existing GAN inversion approaches often consider a single modality of the input image, neglecting other auxiliary cues in images for improvements. Addressing these problems, we propose a novel GAN inversion approach, dubbed \emph{MMInvertFill}, for image inpainting. MMInvertFill contains primarily a multimodal guided encoder with a pre-modulation and a GAN generator with $\mathcal{F}\&\mathcal{W}^+$ latent space. Specifically, the multimodal encoder aims to enhance the multi-scale structures with additional semantic segmentation edge texture modalities through a gated mask-aware attention module. Afterwards, a pre-modulation is presented to encode these structures into style vectors. To mitigate issues of conspicuous color discrepancy and semantic inconsistency, we introduce the $\mathcal{F}\&\mathcal{W}^+$ latent space to bridge the gap between GAN inversion and image inpainting. Furthermore, in order to reconstruct faithful and photorealistic images, we devise a simple yet effective Soft-update Mean Latent module to capture more diversified in-domain patterns for generating high-fidelity textures for massive corruptions. In our extensive experiments on six challenging datasets, including CelebA-HQ~\cite{DBLP:conf/cvpr/Lee0W020}, Places2~\cite{DBLP:journals/pami/ZhouLKO018}, OST~\cite{DBLP:conf/cvpr/WangYDL18}, CityScapes~\cite{DBLP:conf/cvpr/CordtsORREBFRS16}, MetFaces~\cite{DBLP:conf/nips/KarrasAHLLA20} and Scenery~\cite{yang2019very}, we show that our MMInvertFill qualitatively and quantitatively outperforms other state-of-the-arts and it supports the completion of out-of-domain images effectively. Our project webpage including code and results will be available at \href{https://yeates.github.io/mm-invertfill}{https://yeates.github.io/mm-invertfill}.
}


\maketitle

\section{Introduction}\label{sec1}
Image inpainting looks for a semantically compatible way to recover a masked image using its unmasked content. It has been widely used to manipulate photographs, such as fixing corrupted images~\cite{DBLP:conf/cvpr/ZhengCC19}, removing unwanted objects~\cite{DBLP:conf/cvpr/CriminisiPT03,DBLP:conf/nips/ShettyFS18}, or changing the position of objects~\cite{DBLP:conf/aaai/SongCSHH19}.

Currently, image inpainting approaches can be roughly categorized into two types: end-to-end U-Net~\cite{DBLP:conf/miccai/RonnebergerFB15} architectures and large-scale trained Generative Adversarial Networks (GANs)~\cite{DBLP:journals/corr/GoodfellowPMXWOCB14}. The former approaches work on the premise that masked images provide enough information for inpainting. When dealing with small holes or removing foreground objects, these models have demonstrated good outcomes~\cite{DBLP:conf/iccv/YuLYSLH19,li2022misf}. However, when the holes get bigger, they may suffer from the ill-posed nature of inpainting and produces repetitive patterns and artifacts. Different from the U-Net like architectures, the latter GAN-based approaches are trained on large-scale benchmarks and aim to produce high-fidelity images that do not exist~\cite{karras2019style}. These methods can fix severe mask (\eg, masked region exceeds even 70\%) because of variations in model structure, training strategy, and objective functions. However, they are not suitable for faithful image restoration and perform poorly in the evaluation of pixel-level similarity. For existing image inpainting methods, it remains challenging to yield outputs with high fidelity while being faithful to original input.

\begin{figure}[t]
  \centering
  \includegraphics[width=0.85\columnwidth]{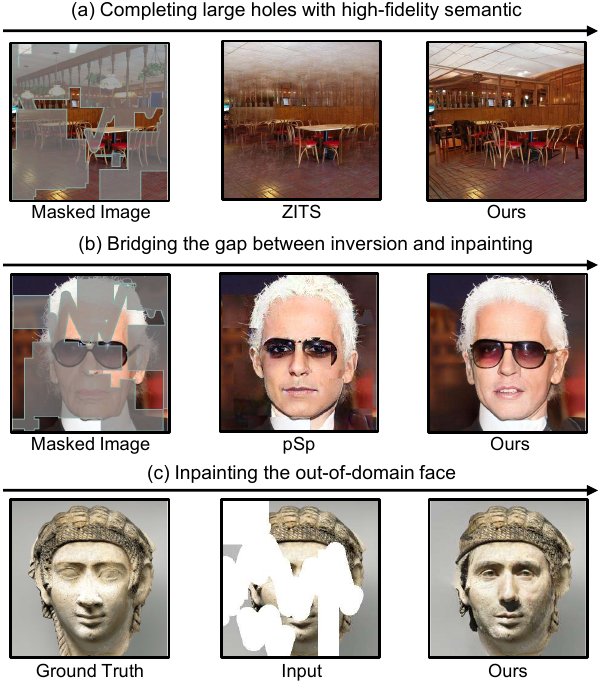}
  \caption{The proposed method supports seamless high-fidelity large hole image inpainting and out-of-domain restoration.}\label{fig:teaser}
\end{figure}

It has been demonstrated recently that GAN models can effectively create high-resolution photo-realistic images~\cite{karras2020analyzing,abdal2019image2stylegan,abdal2020image2stylegan}. In these models, GAN inversion~\cite{zhu2020domain} is an efficient way to bridge conditional generation and GAN. Specifically, when simply fed with stochastic vectors of the latent space, GAN cannot be applied for image-to-image translation. To deal with this issue, GAN inversion approach applies a pre-trained GAN as a prior, and encodes the inputs into latent space that represents the objective images and produces high-fidelity translation results. This serves as an inspiration for several ways~\cite{richardson2021encoding,gu2020mganprior} that make a great effort to develop GAN inversion for image inpainting. However, these approaches have ignored a hard constraint, that {\it the unmasked regions in the input and the output should be the same}, in image inpainting, resulting in color discrepancy and semantic inconsistency (see Fig.~\ref{fig:teaser} (b)). In this work, we call this ``gapping'' problem. To alleviate this gapping problem, previous methods usually require post-processing such as image blending, making them complicated and inefficient.

Faithful inpainting builds on the assumption that unmasked regions provide sufficient knowledge. The incorporation of auxiliary priors from images is intuitive when the supplied known information is trivial. Recently, there are attempts~\cite{DBLP:conf/bmvc/SongYSWHK18,DBLP:conf/cvpr/XiongYLYLBL19,DBLP:conf/iccvw/NazeriNJQE19,DBLP:conf/eccv/LiaoXWLS20,DBLP:conf/icpr/ArdinoL0LN20,DBLP:conf/cvpr/Liao00L021} to introducing auxiliary priors such as \textit{edges} and \textit{segmentation} for image inpainting with improved performance. However, they still suffer from the biased prior issue due to employing predicted priors for guidance. 
The term \emph{biased prior} refers to directly using predicted auxiliary structure maps to guide inpainting, such as inpainted canny edge estimation and inpainted segmentation estimation. However, since the predicted structures are biased relative to the ground truth, using them directly can introduce cumulative error and mistakes during inpainting.
 In addition, they rarely study the use of masked multi-modal structures as an input to the model but refer to it as terms of supervising. The approach may be used more broadly by taking into account multi-modal auxiliary priors.

In this paper, we introduce a novel multimodal guided GAN inversion model, dubbed \emph{MMInvertFill}, for image inpainting. MMInvertFill follows the encoder-based inversion fashion architecture~\cite{richardson2021encoding} and consists of an multimodal guided encoder and a GAN generator. We first develop a new latent space $\mathcal{F}\&\mathcal{W}^+$ that encodes the original images into style vector to enable the accessibility of the generator backbone to inputs, decreasing color discrepancy and semantic inconsistency. In the encoder, we develop the adaptive contextual bottlenecks for better context reasoning. To adapt to the current image content and missing region, the gating mask is updated to weight different dilated convolutions to enhance base features. 
Then, the multimodal mutual decoder is proposed to decode the enhanced features into three modalities, \ie, RGB image, and corresponding semantic segmentation and edge textures. It consists of one image inpainting branch and two auxiliary branches for semantic segmentation and edge textures. Unlike existing approaches based on direct guidance from predicted auxiliary structures, we focus on jointly learning the unbiased discriminative interplay information among the three branches.
Specifically, the proposed gated mask-aware attention mechanism integrates multi-modal feature maps via auxiliary denormalization to reduce duplicated and noisy content for image inpainting. Supervised by ground-truth RGB images, semantic segmentation and edge maps, the whole network is trained in an end-to-end fashion efficiently. 
Besides, to make full use of the encoder, we present pre-modulation networks to amplify the reconstruction signals of the style vector based on the predicted multi-scale structures, further enhancing the discriminative semantic. Then, we propose a simple yet effective soft-update mean latent technique to sample a dynamic in-domain code for the generator. Compared to using a fixed code, our method is able to facilitate diverse downstream goals while reconstructing faithfully and photo-realistically, even in the task of unseen domain. To verify the superiority of our method, we conduct extensive experiments on four datasets, including CelebA-HQ~\cite{DBLP:conf/cvpr/Lee0W020}, Places2~\cite{DBLP:journals/pami/ZhouLKO018}, OST~\cite{DBLP:conf/cvpr/WangYDL18}, CityScapes~\cite{DBLP:conf/cvpr/CordtsORREBFRS16}, MetFaces~\cite{DBLP:conf/nips/KarrasAHLLA20}, and Scenery~\cite{yang2019very}. The results demonstrate that our method achieves favorable performance, especially for images with large corruptions. Furthermore, our approach can handle images and masks from unseen domains by optimizing a lightweight encoder without retraining the GAN generator on a large-scale dataset. Fig.~\ref{fig:teaser} shows several visual results of our approach.

The contributions of our work are summarized as three-fold: 
\begin{itemize}
    \item We introduce a novel $\mathcal{F}\&\mathcal{W}^+$ latent space to resolve the problems of color discrepancy and semantic inconsistency and thus bridge the gap between image inpainting and GAN inversion.
    \item We propose an end-to-end multi-modality guided transformer to learn interplay information from multiple modalities including RGB image, edge textures and semantic segmentation, within the multi-scale spatial-aware attention mechanism with auxiliary denormalization to capture compact and discriminative multi-modal features to guide unbiased image inpainting.
    \item We suggest pre-modulation networks to encode more discriminative semantic from compact multi-scale structures and soft-update mean latent to synthesize more semantically reasonable and visually realistic patches by leveraging diverse patterns.
    \item Extensive experiments on six benchmarks show that the proposed approach outperforms current state-of-the-arts, evidencing its effectiveness, especially for large holes.
\end{itemize}

This paper builds upon our preliminary conference publications~\cite{yu2022high} and~\cite{yu2022unbiased} and extends them in different aspects. {\bf (1)} We combine GAN inversion and Unbiased Multi-modal Guidance networks for image completion, which incorporates the more semantic structure of diverse modals and realizing implicitly guidance of image completion. This way, we are able to obtain user-guide results by the input of auxiliary structure. {\bf (2)} We propose an effective Gated Mask-aware Attention for fusing multi-modal feature maps, which modifies the MSSA module in~\cite{yu2022unbiased} for improvements. {\bf (3)} We include more in-depth analysis of the proposed approach for better understanding. {\bf (4)} We supplement more comprehensive experiments, \eg, the latest baselines, benchmarks, and evaluation metrics. 

\begin{figure}[t]
    \centering
    \includegraphics[width=\linewidth]{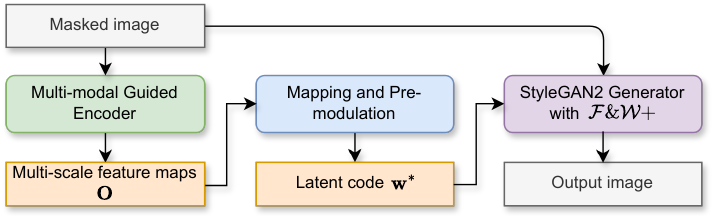}
    \caption{A high-level schematic diagram illustrating the full pipeline of the proposed method.}
    \label{fig:workflow}
\end{figure}

The remaining parts of this work are structured as described below. Section~\ref{sec:related_work} discusses approaches related to this paper. Our approach is elaborated in Section~\ref{sec:methodology}. In Section~\ref{sec:experiments}, we introduce the experiments settings and implementation details. Experiments results are demonstrated in~\ref{sec:results}, including comparisons with state-of-the-arts, ablation studies, and visual analysis, followed by conclusions in Section~\ref{sec:conclusion}. Figure \ref{fig:workflow} aids in visualizing the interconnections between the individual components of the system, providing a clearer and more immediate understanding of the overall approach.

\section{Related Work}
\label{sec:related_work}

\subsection{Image Inpainting}

Image inpainting could be treated as a conditional translation task with hard constraint. Present image inpainting methods mainly employ the encoder-decoder architecture since Pathak \etal ~\cite{DBLP:conf/cvpr/PathakKDDE16} integrate U-Net and GAN discriminator~\cite{DBLP:journals/corr/GoodfellowPMXWOCB14} that help train the U-Net and mitigate the blurring caused by the pixel-level averaging property of a reconstruction loss.
Follow-up techniques specifically discard invalid signals in masked areas. Liu \etal~\cite{DBLP:conf/eccv/LiuRSWTC18} add a heuristic step to vanilla convolution that update a mask each layer after the convolutional transformation, and Yu \etal~\cite{DBLP:conf/iccv/YuLYSLH19} technically replace the heuristic update technique with a learnable sigmoid-activated convolution layer. Multiple residual modules of dilation convolution~\cite{DBLP:conf/cvpr/YuKF17} are introduced by GLILC~\cite{DBLP:journals/tog/IizukaS017} as the encoder bottleneck in order to more effectively use context between missing and uncorrupted areas. Due to just sampling non-zero places, it can cause the ``gridding'' issue~\cite{DBLP:conf/wacv/WangCYLHHC18}. In other words, a single constant dilation rate causes either problematic mask crossing over or sparse convolution kernels. Wang \etal~\cite{DBLP:conf/nips/WangTQSJ18} create a generative multi-column network for image inpainting to achieve this. Zeng \etal~\cite{DBLP:journals/corr/abs-2104-01431} suggest the AOT blocks to combine contextual modifications from diverse receptive fields, which capture both rich patterns of attention and instructive distant visual contexts. In contrast to the approaches mentioned above, we provide a novel adaptive contextual bottleneck in the encoder, where dynamic gating update weights various paths of dilated convolutions based on various masks and image contents.

A range of attention~\cite{DBLP:conf/eccv/YanLLZS18,DBLP:conf/cvpr/Yu0YSLH18,DBLP:conf/iccv/LiuJX019,DBLP:conf/cvpr/LiWZDT20,zeng2021cr} initiatives have recently been presented, using references as helpful signals that are comparable to the traditional exemplar-based method. In order to direct a patch-swap operation, RFR~\cite{DBLP:conf/cvpr/LiWZDT20} specifically applies gradually restoration at the bottleneck while sharing the attention scores. ProFill~\cite{zeng2020high} uses the confidence map generated by spatial attention to iteratively conduct inpainting. An objective function for contextual reconstruction that learns query-reference feature similarity is produced by CRFill~\cite{zeng2021cr}. In order to maintain consistency, MEDFE~\cite{liu2020rethinking} uses spatial and channel equalization while learning to jointly represent structures and textures. Through parallel routes, CTSDG~\cite{guo2021image} connects texture and structure, and then employs bidirectional gated layers to combine them. While taking into consideration both mask-awareness and long-dependency, we include a multi-scale spatial-aware attention method that combines multi-modal feature maps through auxiliary denormalization to decrease redundant and noisy material for image inpainting.

Researchers have suggested various modules to extend the aforementioned U-Net style approach to more complex scenarios, but due to the restrictions of the U-Net model structure, these mechanisms have difficulty filling in missing regions at extreme scales and instead produce blurry mosaics or artifacts. As a result, lately, various alternatives have appeared. Score-SDE~\cite{DBLP:conf/iclr/0011SKKEP21} is one well-known example. It proposes a scoring model that saves the gradient computation of energy-based models for effective sampling. Another more prevalent one is GAN, which trained on millions of images, and the neural representations of the model's layers include levels of image semantics and patterns, allowing the synthetic images to produce photo-quality visuals.

\begin{figure*}[t]
  \centering
  \includegraphics[width=0.91\textwidth]{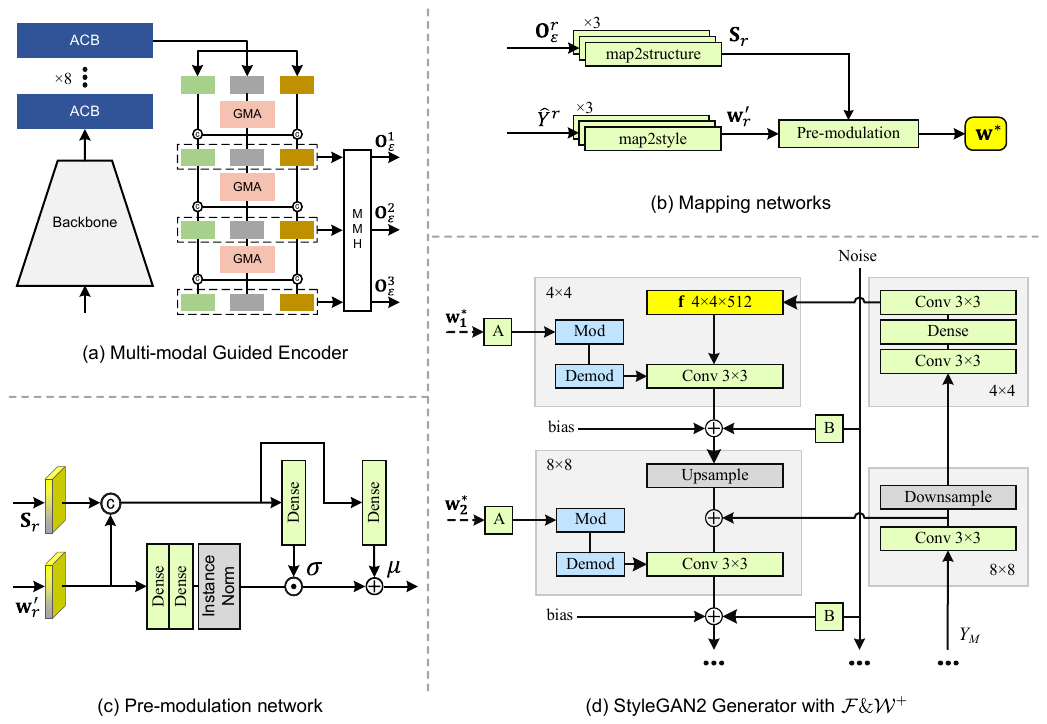}
  \caption{Illustration of our MMInvertFill, including multi-modal guided encoder (image (a)), feature pyramid-based mapping networks (image (b)),  mapping network with pre-modulation network (image (c)) and StyleGAN2 generator with our proposed $\mathcal{F} \& \mathcal{W}^+$ latent space (image (d)).}\label{fig:mminvertfill}
\end{figure*}

\subsection{Inpainting with GAN Inversion}
Instead of using the initial stochastic vector, StyleGAN~\cite{karras2019style} implicitly learns hierarchical latent styles, giving style-modulation. control over the output style at coarse-to-fine degrees of detail. For enhanced quality, StyleGAN2~\cite{karras2020analyzing} also suggests weight demodulation, route length regularization, and generator redesign. Although they may generate well without any provided images, they also need paired training data, specialized networks~\cite{mao2017least}, or regularization~\cite{gulrajani2017improved,DBLP:conf/iclr/MiyatoKKY18}. A typical technique is GAN inversion~\cite{zhu16generative}, which uses the inherent statistics of a well trained large-scale GAN as the prior for general applications~\cite{zhu2020domain,abdal2019image2stylegan}. The two main types of GAN inversion techniques now in use are optimized-based~\cite{cheng2022inout,gu2020mganprior} and encoder-based~\cite{richardson2021encoding,zhu2020domain,yang2022vtoonify}. The approach mGANprior~\cite{gu2020mganprior}, which has applicability in a variety of tasks, including inpainting, uses multiple latent codes and adaptive channel significance. pSp~\cite{richardson2021encoding} synthesizes with the mapping network in order to extract images into style vectors of latent space as the input of StyleGAN. The ``gapping'' problem is ignored, resulting in inconsistent color and semantic.

\subsection{Multi-task image inpainting}
Attempting to re-create a lost area is difficult because of its ill-posed nature. Techniques~\cite{DBLP:conf/iccvw/NazeriNJQE19,liu2020rethinking,DBLP:conf/cvpr/Liao00L021,guo2021image} construct intermediate structures to generate more faithful output by using auxiliary labels (\eg, \textit{edges}, \textit{segmentation}, and \textit{contours}). 

To get finer inpainting results, Edge Connect (EC)~\cite{DBLP:conf/iccvw/NazeriNJQE19} makes use of a corrupted canny edge image. For the purpose of avoiding canny edge's pool coherence, Cao \etal~\cite{DBLP:journals/corr/abs-2103-15087} introduce an additional encoder. DeepLabv3+~\cite{DBLP:conf/eccv/ChenZPSA18} is used to predict the segmentation of a corrupted image, and the Segmentation Prediction and Guidance network (SPG)~\cite{DBLP:conf/bmvc/SongYSWHK18} is a two-stage inpainting model based on semantic segmentation. Foreground object and its contour can be located and filled using a novel three-stage model~\cite{DBLP:conf/cvpr/XiongYLYLBL19} that breaks up the inter-object intersections.

Semantic Guidance and Evaluation (SGE) network~\cite{DBLP:conf/eccv/LiaoXWLS20} couples with segmentation and inpainting at various layers of the decoder. In this scenario, the segmentation, after being completed and receiving a confidence score, directs the image inpainting through the use of semantic normalization~\cite{DBLP:conf/cvpr/Park0WZ19}. In order to capture the semantic relevance that exists between segmentation and textures while performing non-local operations, Liao \etal~\cite{DBLP:conf/cvpr/Liao00L021} propose the use of the Semantic-wise Attention Propagation (SWAP). Yang \etal~\cite{DBLP:conf/aaai/YangQS20} predict explicit edge embedding with an attention mechanism in order to facilitate image inpainting through the use of the multi-task learning technique. It is important to point out that the majority of the works that have been discussed thus far employ estimated auxiliary structures as the direct guide for image inpainting. Differently, we propose a Gated mask-aware attention module to guide inpainting based on jointly acquired discriminative features using unbiased multi-modal priors.

\subsection{Transformers in Image Inpainting} Recently developed techniques~\cite{DBLP:conf/mm/DengHZMW21,DBLP:conf/mm/YuZWPCLMXM21} based on Vision Transformer\cite{DBLP:conf/iclr/DosovitskiyB0WZ21} that provides long-range dependencies between input features have improved picture inpainting. Deng \etal~\cite{DBLP:conf/mm/DengHZMW21} take advantage of the intimate ties between the corrupted and uncorrupted regions. An autoregressive transformer that can simulate missing regions in both directions has been introduced by Yu \etal~\cite{DBLP:conf/mm/YuZWPCLMXM21}. A new multi-modality guided transformer is proposed in our method in order to capture the interplay information between three modalities.

\subsection{Our approach}
In this paper we focus on encoder-based GAN inversion to improve generation fidelity for image completion. The suggested  Multi-modal Invertfill is related to but significantly different from previous studies. In specific, Multi-modal Invertfill is relevant to the methods in~\cite{richardson2021encoding,zhu2020domain,xu2021generative} where encoder-based architecture is adopted. However, differing from them, we introduce a new $\mathcal{F}\&\mathcal{W}^+$ latent space to explicitly handle the ``gapping'' issue which is ignored in previous algorithms. Our method also shares a similar spirit with the works of~\cite{DBLP:conf/iccvw/NazeriNJQE19,guo2021image} that adopt auxiliary priors for image inpainting. The difference is that these approaches may suffer from unbiased guidance, while the proposed method exploits Multi-modal Guided Decoder with Gated Mask-aware Attention and can achieve image inpainting with high-fidelity semantics.

\section{Methodology}\label{sec:methodology}

\subsection{Overview}

Suppose we are given a degenerated image $Y_M = Y \odot (1 - M), Y \in \mathbb{R}^{H\times W\times 3}$, where $Y$ and $M$ stand for an original image and mask. Pixel values in the missing region $M$ equal to $0$ are defined as invisible pixels. Solving image completion is an inverse problem of degeneration, and a solver yields reconstructed synthesis $\hat{Y}$. The goal of the solver lies in faithful restoration or high-fidelity generation considering the ill-posed of image completion.

The proposed method is derived from encoder-based GAN inversion~\cite{xia21inversionsurvey}, which consists of a multi-modal multi-task transformer encoder and a pre-trained GAN generator. The encoder is named Multi-modal Guided Encoder (MGE), which takes masked multi-modal input of segmentation map $S_M$, edge map $E_M$ and RGB image $Y_M$, and restores them as well as extracts the intermediate style vector/latent code.

The pre-trained GAN generator synthesizes a high-fidelity image that depends on style vector.
This method generalizes the image inpainting where multi-modal inputs provide the synthesis more auxiliary guidance. Specifically, The input edge and segmentation maps have individual masks different from $M$ and could be zero matrices (standard image inpainting). We will introduce each stage in the following sections.

\begin{figure}[!t]
  \centering
  \includegraphics[width=\columnwidth]{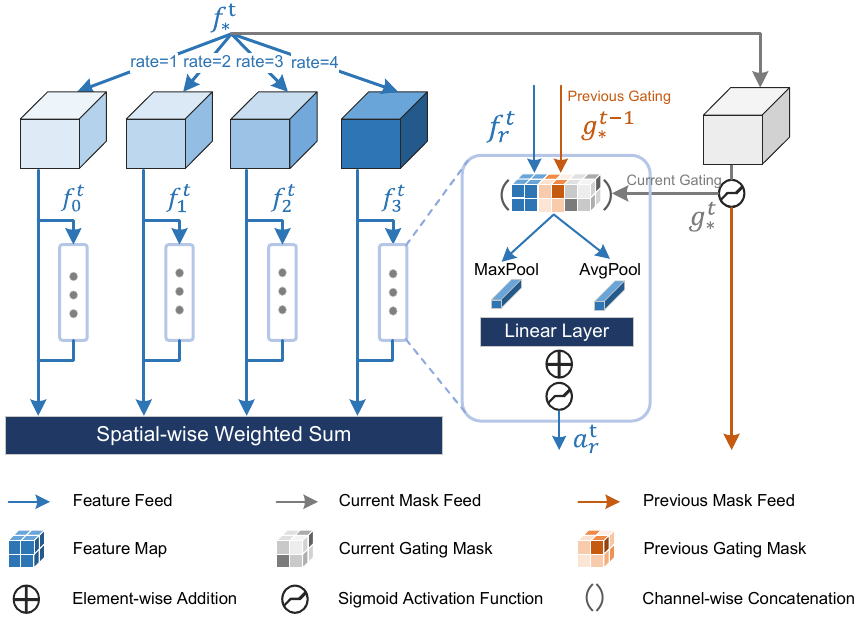}
  \caption{Illustration of Adaptive Contextual Bottleneck.}\label{fig:ACB}
\end{figure}

\subsection{Multi-Modality Guided Encoder}

In order to extract multi-modal inputs hierarchically, we maintain the U-Net architecture in place of feature pyramid networks~\cite{richardson2021encoding}. As illustrated in Fig.~\ref{fig:mminvertfill}(a), the Multi-Modality Guided Encoder is made up of a downsampling backbone with tail gated conv~\cite{DBLP:conf/iccv/YuLYSLH19}, an adaptive contextual bottlenecks, and a Multi-modal Mutual Decoder (MMD) with gated mask-aware attention. The backbone has a five-layer convolution and self-attention module. The feature maps from each upsampling layer of the decoder flow to the associated lateral head, dubbed Multi-modal Multi-scale Head (MMH), which produces hierarchically predicted restoration. Predicted segmentation, edge map, and RGB image are denoted as $\hat{S}^r$, $\hat{E}^r$, $\hat{Y}^r$, respectively. $r$ corresponds to three scales of MMD. $\mathbf{O}^{r}_{\mathcal{E}}$ stands for concatenation of these prediction restoration, \ie, $[\hat{S}^r;\hat{E}^r;\hat{Y}^r]$. In Fig.~\ref{fig:mminvertfill}(d), $Y_M$ is not explicitly annotated; it serves as the input to the RGB branch. The 'F' in $F\&W+$ represents feature maps at each scale from the RGB branch, as shown on the right side of Figure~\ref{fig:mminvertfill}(d), connected via skip-connections. The left side of Figure~\ref{fig:mminvertfill}(d) corresponds to the W+ latent code at each scale.

\subsubsection{Adaptive Contextual Bottlenecks}

Between the downsampling and upsampling of MGE, the bottleneck employs multi-stream structures to weight dilated convolutions and encodes current image content and missing region for enhanced context reasoning. We construct a stack of Adaptive Contextual Bottlenecks (ACBs) that use dynamic gating to adjust to mask form size and image context. As can be seen in Fig.~\ref{fig:ACB}, the ACB module makes use of four parallel paths of convolutional layers with adjustable dilation rates and a single gating mask to weight these convolutions. Therefore, the encoder can widen its field of convolutional perception and identify the most likely route for the masked region.

Suppose that the base features $f_{*}^{0}$ and gating $g_{*}^{0}$ are initialized by the last layer of downsampling, and the last layer of downsampling uses gated convolution~\cite{DBLP:conf/iccv/YuLYSLH19} that could extract and determine a dynamic gating mask. Then $f_{*}^{t}$ and $g_{*}^{t}$ at each bottleneck are updated by the ACB block.
The gating mask $g^l_\ast$ is used to estimate the probability of missing region based on the feature map at the $t$-th layer ($t=1,\cdots,T$), \ie, $g^t_\ast=\mathrm{gconv}(f^t_\ast)$, where $\mathrm{gconv}$ denotes the gated convolution operation~\cite{DBLP:conf/iccv/YuLYSLH19}. In term of each pathway with dilation rate $r$, we compute the dilated feature maps $f^t_r$ based on $f^t_\ast$ and corresponding weight $a^t_r$. Similar to \cite{DBLP:conf/eccv/WooPLK18}, the spatial-wise weight $a^t_r$ is calculated based on both average and max pooling of concatenation of dilated feature maps $f^t_r$ and gating masks $g^t_\ast,g^{t-1}_\ast$, \ie, 
$a^t_r = \phi(\mathrm{fc}(\mathrm{avg}(g^t_r))+\mathrm{fc}(\mathrm{max}(g^t_r)))$, where $\phi(\cdot)$ is the sigmoid function, and $\mathrm{avg}$ and $\mathrm{max}$ are the average and maximal pooling respectively. $\mathrm{fc}$ denotes the fully-connected layer, and the gating mask for each pathway is calculated as $g^t_r = \mathrm{conv}([f^t_r;g^t_\ast;g^{t-1}_\ast])$.
Finally, the feature map at the $(t+1)$-th ACB layer is updated by the spatial-wise weighted summation of $f^t_r$ as
\begin{equation}
f^{t+1}_\ast = \sum_{r \in R} \frac{\exp(a^t_r)}{\sum_{r\in R}\exp(a^t_r)} \cdot f^t_r + f^t_\ast,
\end{equation}
where $R$ denotes the set of different dilation rates. The fractional term denotes element-wise product between dilated feature map $f_{r}^{t}$ and attention vector $a_{r}^{t}$, weighting dilation block based on mask and image context. For simplicity, we omit the subscript $t$ in the following sections.

\subsubsection{Multi-modal Mutual Decoder}

In order to learn the multi-modal structure concurrently, the decoder utilizes stacks of transformer blocks equipped with enhanced features $f_\ast$. One of its forks is used for inpainting, which restores the masked image, and the other two are auxiliary and supply unbiased segmentation and edge priors. Skipping connections aids three forks in recalling the encoded features of the input and keep networks from degrading.

\begin{figure*}[htbp]
  \centering
  \includegraphics[width=0.93\textwidth]{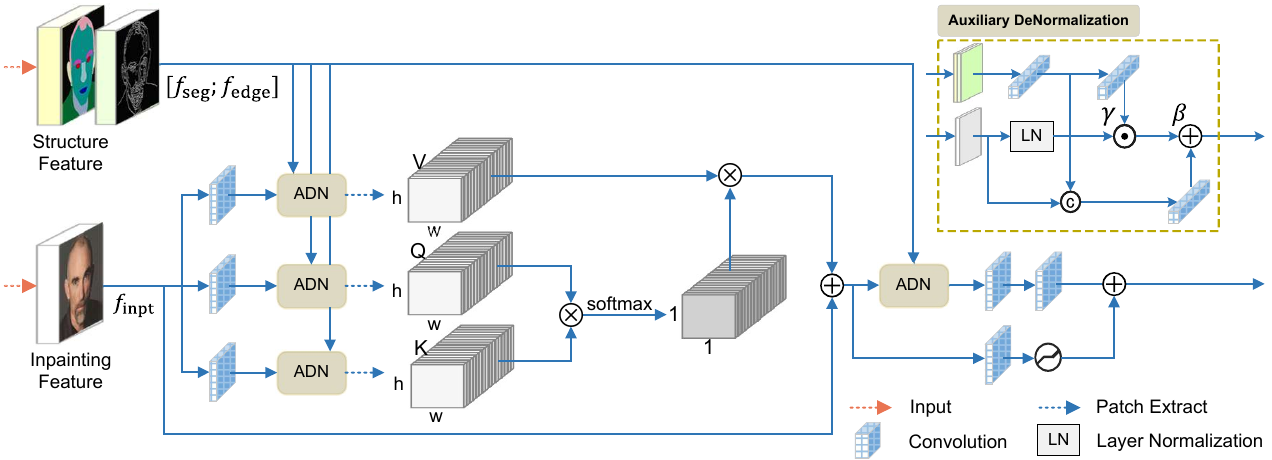}
  \caption{Illustration of Gated Mask-aware Attention.}\label{fig:gma}
\end{figure*}

To learn mutual features from different modalities, it makes logical to simply concatenate or add their feature maps. However, such approaches may introduce redundant and noisy content for image inpainting. In order to efficiently incorporate compact features from auxiliary branches, we introduce a new Gated Mask-aware Attention (GMA) technique as shown in Fig.~\ref{fig:gma}. The GMA block blends feature maps from the three outlets in each transformer block. Each three stage of GMA indicates a upsampling, as depicted in Fig.~\ref{fig:mminvertfill} (a). We hierarchically obtain inpainted image, edge map, and segmentation map by employing corresponding lateral heads.

{\noindent {\bf Gated Mask-aware Attention.}}
Based on the enhanced feature maps $f_\ast$, we use $f_\text{inpt}, f_\text{edge}, f_\text{seg}$ to denote the input feature maps for the inpainting branch, edge branch, and segmentation branch, respectively. We propose the following Auxiliary DeNormalization (ADN) to integrate the feature maps instead of directing by predicted restorations: 
\begin{equation}
\mathrm{ADN}(f_\text{inpt} \| [f_\text{edge}; f_\text{seg}]) = \gamma \odot \mathrm{LN}(f_\text{inpt}) + \beta,
\end{equation}
where $[;]$ denotes the matrix concatenation along channel dimension, and $\odot$ the element-wise multiplication. $\mathrm{LN}$ denotes layer normalization~\cite{DBLP:journals/corr/BaKH16}. $\gamma$ and $\beta$ are the affine transformation parameters learned by two convolutional layers based on $[f_\text{edge}; f_\text{seg}]$. This allows for the integration of multi-modal features based on context from auxiliary structures that shift in composition depending on location.

Then, the merged features are embedded into query $Q$, key $K$ and value $V$. The embedded feature map is spatially split into $N$ patches, \ie, $P_i\in\mathbb{R}^{h\times w\times c} (i=1,\dots,N)$, where $h,w,c$ denote the height, width and channel of patches respectively. The normalized self-attention $\alpha_{i,j}$ between patches $i$ and $j$ can be calculated as,
\begin{equation}
    \alpha_{i,j}= \mathrm{softmax}(\frac{Q_i \cdot K_j^T}{\sqrt{h\cdot w\cdot c}})\cdot M^\prime, \quad i, j \in {1, \dots, N} ,
\end{equation}
where $M^\prime$ learned by a sigmoid-activated convolution stands for dynamic gating that makes the self-attention mask-aware. The insight is inspired of previous findings~\cite{DBLP:conf/iccv/YuLYSLH19,DBLP:conf/eccv/LiuRSWTC18}.

Note that we can perform multi-head self-attention like~\cite{DBLP:conf/iclr/DosovitskiyB0WZ21}. Thus the feature map of each patch is updated in a non-local form, \ie, $\hat{P}_i = \sum_{j=1}^{N} \alpha_{i,j} V_j$. Compared to the previous MSSA~\cite{yu2022unbiased}, Gated Mask-aware Attention allows each token to dynamically adjust its weight according to the mask. This is crucial because the input image is a masked image, and tokens in the masked region do not provide valuable information. In contrast, the self-attention mechanism in the entire MSSA treats all tokens equally, which is not reasonable in image inpainting~\cite{li2022mat}.

{\noindent {\bf Gated Feed-forward.}}
Finally, we piece all feature maps $\hat{P}_i$ together and reshape them with the original scale of input inpainting features $f_\text{inpt}$. Following the gated feed-forward layer, we can output the final feature maps for inpainted image prediction. Similar to gated conv~\cite{DBLP:conf/iccv/YuLYSLH19}, the gated feed-forward layer can ease the color discrepancy problem by detecting potentially corrupted and uncorrupted regions.

\subsection{Seamless GAN Inversion}

Our architecture is derived from encoder-based GAN inversion~\cite{xia21inversionsurvey} including Multi-modality Guided Encoder $\mathcal{E}$ that extracts input images and provides hierarchical reconstructed RGB images and auxiliary structures. The mapping networks with pre-modulation produce extended style vector $\mathbf{w}^*$ for the input of StyleGAN generator $\mathcal{G}$. We attach three prediction heads to the encoder $\mathcal{E}$ for generating prediction restoration $\mathbf{O}_{\mathcal{E}}^r=\{\mathbf{O}_{\mathcal{E}}^1,\mathbf{O}_{\mathcal{E}}^2,\mathbf{O}_{\mathcal{E}}^3\}$ in correspondence to three different scale. We follow the \textit{map2style}~\cite{richardson2021encoding} for the mapping network and cut down the network number from $18$ to $3$, each of which corresponds to the disentanglement level of image representation (\ie, coarse, middle and fine~\cite{karras2019style,richardson2021encoding}). Three \textit{map2style} networks encode the output feature map of the encoder into the intermediate latent code $\mathbf{w}^\prime \in \mathbb{R}^{3 \times 512}$. Similarly, we replicate \textit{map2style} as \textit{map2structure} to project reconstructed RGB images $\mathbf{O}_{\mathcal{E}}^r$ gradually into structure vector $\mathbf{S}_r = \{\mathbf{S}_{1}, \mathbf{S}_{2}, \mathbf{S}_{3}\}$.

\subsubsection{Pre-modulation}

As depicted as Fig.~\ref{fig:mminvertfill} (b), before executing the style modulation in the generator, we perform $L$ pre-modulation networks to project the semantic structure $\mathbf{S}_r$ into the style vector $\mathbf{w}^*$ in latent space $\mathcal{F} \& \mathcal{W}^+$, i.e., $\mathbf{w}^* = \mathcal{E}(\mathbf{I}_m), \mathbf{w}^* \in \mathbb{R}^{L \times 512}$, where $L = \log_2(s) \cdot 2 - 2$ denotes number of style-modulation layers of StyleGAN2 generator, and is adjusted by the image resolution $s$ on the generator side. 
As shown in Fig~\ref{fig:mminvertfill}, we adopt Instance Normalization (IN)~\cite{ulyanov2016instance} to regularize the $\mathbf{w}^\prime$ latent code, then carry out denormalization according to multi-scale structure vector $\mathbf{S}_r$, 
\begin{equation}
    \mathbf{w}_{l}^* = \mathbf{\sigma} \odot \text{IN}\left(\mathbf{w}^{\prime}_{r}\right) + \mathbf{\mu} ,
\end{equation}
where $l \in \{1,2, \dots ,L\}$ denotes the index of style vectors, $r \in [1,3]$ indicates three vectors $\mathbf{w}^{\prime}$ correspond to level of coarse to fine, $(\sigma, \mu)$ is a pair of the affine transformation parameters learned by two-layer convolutions. Different than previous methods in only using intermediate latent code from a network, the proposed pre-modulation is a lightweight network and novel in applying more discriminative multi-scale features to help latent code perceive uncorrupted prior and better guide image generation.

\subsubsection{$\mathcal{F} \& \mathcal{W}^+$ Latent Space}

The GAN is initially fed with a stochastic vector $z \in \mathcal{Z}$, and previous works~\cite{abdal2019image2stylegan,gu2020mganprior,richardson2021encoding,zhu2020domain} invert the source images into the intermediate latent space $\mathcal{W}$ or $\mathcal{W}^+$,  which is a less entangled representation than latent space $\mathcal{Z}$. The style vectors $w \in \mathcal{W}$ or $w^+ \in \mathcal{W}^+$ are sent to the style-modulation layers of pre-trained StyleGAN2 to synthesize target images. These approaches can be formulated mathematically as follows,
\begin{equation}
	\mathbf{O}_\mathcal{G} = \mathcal{G}(\mathcal{E}(\mathbf{I}_m)),\mathcal{E}(\mathbf{I}_m) \sim W^+ ,
	\label{equ:ori}
\end{equation}
where $\mathcal{E}(\cdot)$ and $\mathcal{G}(\cdot)$ represent the encoder that maps source images into latent space and the pre-trained GAN generator, respectively. 

Nevertheless, the above formulation in Equ.~(\ref{equ:ori}) may encounter the ``gapping'' issue in image translation tasks with hard constraint, \eg, image inpainting. The hard constraint requires that parts of the source and recovered image remain the same. We formally defined the hard constraint in image inpainting as $Y \odot (1-M) \equiv \mathbf{O} \odot (1 - M)$. Intuitively, we argue that the ``gapping'' issue is caused by that the GAN model cannot directly access pixels of the input image but the intermediate latent code. To avoid the semantic inconsistency and color discrepancy caused by this problem, we utilize the corrupted image $Y_M$ as one of the inputs to assist with the GAN generator inspired by skip connection of U-Net~\cite{DBLP:conf/miccai/RonnebergerFB15}. In detail, $Y_M$ is fed into the RGB branch as shown in Fig.~\ref{fig:mminvertfill}(d), the feature map between RGB branch and the generator are connected by element-wise addition.
Hence, the previous formulation in Equ. \eqref{equ:ori} is updated as:
\begin{equation}
    \mathbf{O}_\mathcal{G} = \mathcal{G}(\mathcal{E}(\mathbf{I}_M), \mathbf{I}_M), \mathcal{E}(\mathbf{I}_M) \sim \mathcal{F} \& \mathcal{W}^+ .
\end{equation}

\subsubsection{Soft-update Mean Latent}

Pixels closer to the mask boundary are more accessible to inpainting, but conversely the model is hard to predict specific content missing. We find that the encoder learns a trick to averaging textures to reconstruct the region away from unmasked region. It causes blurring or mosaic in some areas of the output image, mainly located away from the mask borders, as shown in Fig~\ref{fig:ablation_sml}. Drawing inspiration from L2 regularization and motivated by the intuition that fitting diverse domains works better than fitting a preset static domain, a feasible solution is to make style code $\mathbf{w}^*$ be bounded by the mean latent code of pre-trained GAN.

The mean latent code is obtained from abundant random samples that restrict the encoder outputs to the average style hence lossy the diversity of output distribution of encoder. In addition, it introduces additional hyperparameters and a static mean latent code that requires loading when training the model.

We adopt dynamic mean latent code instead of static one by stochastically fluctuating the mean latent code while training. 
Further, we smooth the effect of fluctuating variance for convergence inspired by a reinforcement learning~\cite{DBLP:journals/corr/LillicrapHPHETS15}. 
For initialization, target mean latent code $\mathbf{\overline{w}}_t$ and online mean latent code $\mathbf{\overline{w}}_o$ are sampled. 
$\overline{\mathbf{w}}_o$ is used in image generation instead of $\overline{\mathbf{w}}_{t}$, which is fixed until $\overline{\mathbf{w}}_o = \overline{\mathbf{w}}_t$ and then resampled. 
Between two successive sampled mean latent codes, $\overline{\mathbf{w}}_o$ is updated by $\mathbf{\overline{w}}_t \leftarrow \tau \mathbf{\overline{w}}_o + (1- \tau)$ per iteration during training, where $\tau$ denotes updating factor and $\mathbf{\overline{w}}_t $ for soft updating target mean latent code. The soft-update mean latent degraded to static mean latent~\cite{richardson2021encoding} when the parameter $\tau$ of soft-update mean latent approaching zero. 

\subsection{Optimization}

Following prior work in inpainting~\cite{DBLP:conf/eccv/LiuRSWTC18,DBLP:conf/cvpr/LiWZDT20}, our architecture is supervised by regular inpainting loss $\mathcal{L}_{\text{ipt}}$, which consists of the high receptive field perceptual loss $\mathcal{L}_\text{P}$ and the adversarial loss $\mathcal{L}_\mathcal{G}$:
\begin{equation}
    \mathcal{L}_{\text{ipt}} = \mathcal{L}_\text{P} + \mathcal{L}_\mathcal{G},
    \label{eq:ipt}
\end{equation}
where all above distance are calculated between $Y$ and $\mathbf{O}_\mathcal{G}$. The perceptual loss $\mathcal{L}_\text{P}=\sum_{p=0}^{P-1} \frac{\| \Psi_{p}^{\mathbf{O}_\mathcal{G}} - \Psi_{p}^{Y} \|_2}{N}$ computes Euclidean norm of encoded feature maps, where activation map of selected layer $p^{th}$ is $\Psi_{p}$, $N$ is the number of elements in $\Psi_{p}^{Y}$. The feature extractor~\cite{DBLP:conf/wacv/SuvorovLMRASKGP22} is a dilated ResNet-50 pretrained on ADE50K~\cite{zhou2017scene}. The adversarial loss $\mathcal{L}_\text{adv}=-\mathbb{E}\left[\mathbf{D}\left([\mathbf{O}_\mathcal{G};\hat{E}^{3}]\right)\right]$ justifies if $O_{G}$ and predicted edge $\hat{E}^{3}$ is fake or real. $\mathbf{D}$ denotes the spectral normalization discriminator \cite{DBLP:conf/iclr/MiyatoKKY18} that is composed of five convolutional layers.

To directly optimize our encoder, the multi-scale reconstruction loss $\mathcal{L}_{\text{msr}}$ is utilized to penalize the deviation of $\mathbf{O}_{\mathcal{E}}^{r}$ at each scale:
\begin{equation}
    \mathcal{L}_{\text{msr}} = \sum_{r=1}^{3} (\mathcal{L}_{\text{rec}}^r + \mathcal{L}_{\text{P}}^{r}
    + \mathcal{L}_{edge}^{r}
    + \mathcal{L}_{seg}^{r}),
    \label{eq:loss_msr}
\end{equation}
where $\mathcal{L}_{\text{rec}}$ and $\mathcal{L}_{\text{P}}^{r}$ are represented as $\ell_1$ and perceptual loss between $Y$ and $\hat{Y}$. $\mathcal{L}_{edge}^{r}$ stands for the binary cross-entropy loss for the edge branch, $\mathcal{L}_{edge}^{r} = \frac{1}{N}\sum_{i=1}^N -[E^i \log \hat{E}^i + (1-E^i) \log (1-\hat{E}^i)]$ predicts the edge structure, and 
$\hat{E}$ is the probability map between $0$ and $1$ for the reconstructed edge while $E$ is the ground-truth edge based on the canny operator~\cite{DBLP:conf/iccvw/NazeriNJQE19}. $\mathcal{L}_{seg}^{r}$ indicates the cross-entropy loss for the segmentation branch, $\mathcal{L}_{seg}^{r}=\frac{1}{N}\sum_{i=1}^N -S^i \log \hat{S}^i$, where $S^i$ and $\hat{S}^i$ denote the ground-truth category and predicted probability for pixel $i$. The role of $\mathcal{L}_{\mathrm{msr}}$ is to supervise the generated image from decoder and make final generation close to the original image.

The soft-update mean latent is used to prevent the encoder from falling into a problematic approach during GAN inversion training. Specifically, the Multi-modal Guided Encoder might reduce L2 loss by overfitting to the average patterns of the training set. We refer to this as a ``trick way" because, although it achieves a lower L2 loss, it gradually shifts the encoded latent code away from the pre-trained StyleGAN2’s latent distribution. This shift decreases the diversity and realism of the generated results.

We adopt the following fidelity loss $\mathcal{L}_{\text{fid}}$ for improving the quality and diversity of output images:
\begin{equation}
    \mathcal{L}_{\text{fid}} = \| \mathbf{w}^* - \overline{\mathbf{w}} \|_2, \text{ } (\mathbf{w}^*, \overline{\mathbf{w}}) \in \mathbb{R}^{L \times 512} .
\end{equation}

The fidelity loss $\mathcal{L}_{\mathrm{fid}}$ is designed as a mean squared loss of style vectors $\mathrm{\mathbf{w}}^{*}$ and mean latent code $\mathrm{\bar{\mathbf{w}}}$. Its role is to improve the quality and diversity of the output images.

Overall, the loss of our networks is defined as the weighted sum of the inpainting loss, the multi-scale reconstruction loss, and the fidelity loss.
\begin{equation}
    \mathcal{L} = \mathcal{L}_{\text{ipt}} + \lambda_{\text{msr}} \mathcal{L}_{\text{msr}} + \lambda_{\text{fid}} \mathcal{L}_{\text{fid}} ,
    \label{eq:loss_all}
\end{equation}
where $\lambda_{\text{msr}}$ and $\lambda_{\text{fid}}$ are the balancing factors for the multi-scale reconstruction loss and the fidelity loss, respectively.

\begin{table*}[t]
\centering
\setlength{\tabcolsep}{7pt}
\caption{Quantitative comparison with the state-of-the-art approaches on CelebA-HQ. $\uparrow$ higher is better, and $\downarrow$ lower
is better. Best and second results are \textbf{highlighted}
and \underline{underlined}.}
\resizebox{\linewidth}{!}{  
\begin{tabular}{lcccccccccc}
\toprule &
           & \multicolumn{3}{c}{FID$\downarrow$}             & \multicolumn{3}{c}{LPIPS$\downarrow$}            & \multicolumn{3}{c}{SSIM$\uparrow$}              \\ \cmidrule(r){3-5}  \cmidrule(r){6-8} \cmidrule(r){9-11}
 & where & 10-30\% & 40-60\% & 70-90\% & 10-30\% & 40-60\% & 70-90\% & 10-30\% & 40-60\% & 70-90\% \\ \midrule
GC~\cite{DBLP:conf/iccv/YuLYSLH19}   & ICCV'19 &            
17.0318&   30.6609         &   53.3808 &
0.1174&    0.2179        &    0.3443        &           
0.8830 &    0.7714        &    0.6386   \\
RFR~\cite{DBLP:conf/cvpr/LiWZDT20}     & CVPR'20 &    
21.8687 &     35.0555       &    52.0651        &           
0.1058 & 0.1897           &   0.2861           &          
\underline{0.9008} &  \underline{0.7993}          &   \underline{0.6884}  \\
ICT~\cite{Wan_2021_ICCV} & ICCV'21 &
26.1688 & 38.5847 & 62.5949  &            
0.1328 & 0.2290 & 0.3467 & 
0.8828 & 0.7721 & 0.6436\\
CTSDG~\cite{guo2021image} &  ICCV'21 &      
23.5799 &    35.0791        &      56.1095     &            
0.1355 &    0.2237        &      0.3309      &            
0.8918 &    0.7946        &   0.6863      \\
Score-SDE~\cite{DBLP:conf/iclr/0011SKKEP21} & ICLR'21 &
23.8945 & 30.1031  & 40.0529   &            
0.1829 & 0.2819  & 0.4279 &           
0.8287 & 0.7140  & 0.5857   \\
LAMA~\cite{DBLP:conf/wacv/SuvorovLMRASKGP22} & WACV'22 &        
18.5979 &      26.8567      &  39.4042         &            
0.1145 &      0.1914      &      0.2858      &           
0.8912 &      \textbf{0.7994}      &     \textbf{0.6979}  \\
MISF~\cite{li2022misf}  &   CVPR'22 & 
17.4082 &    25.7009        &       43.4309      &            
0.1079 &    0.1999        &        0.3260    &            
\textbf{0.9026} &    0.7992        &        0.6773   \\\midrule
MMT~\cite{yu2022unbiased}  &   ECCV'22 & 
21.5339 & 27.9715 & 35.3631 & 
0.1573  & 0.2238 & 0.3041    &            
0.8468  & 0.7572 & 0.6579   \\
InvertFill~\cite{yu2022high}  &   ECCV'22 & 
\underline{12.2064} & \underline{17.8915}   & \underline{25.9103} &            
\underline{0.0821} & \underline{0.1705}  & \underline{0.2780}  & 
0.8909 & 0.7684  & 0.6493   \\
Ours  & - &
\textbf{10.1012}     &  \textbf{16.5700}          &        \textbf{23.6047}  &    
\textbf{0.0743}   &  \textbf{0.1635}     &   \textbf{0.2667}    &   
0.8956 &  0.7790      &   0.6569    \\ \bottomrule
\end{tabular}
}
\label{tab:main_compare_1}
\end{table*}

\begin{table*}[t]
\centering
\setlength{\tabcolsep}{7pt}
\caption{Quantitative comparison with the state-of-the-art approaches on Places2. $\uparrow$ higher is better, and $\downarrow$ lower
is better. Best and second results are \textbf{highlighted}
and \underline{underlined}.
}
\resizebox{\linewidth}{!}{  
\begin{tabular}{lcccccccccc}
\toprule & &  \multicolumn{3}{c}{FID$\downarrow$} &
           \multicolumn{3}{c}{P-IDS (\%) $\uparrow$} &
           \multicolumn{3}{c}{U-IDS (\%) $\uparrow$}   \\ \cmidrule(r){3-5}  \cmidrule(r){6-8} \cmidrule(r){9-11}
& where & 10-30\% & 40-60\% & 70-90\% & 10-30\% & 40-60\% & 70-90\% & 10-30\% & 40-60\% & 70-90\%  \\ \midrule
EC~\cite{DBLP:conf/iccvw/NazeriNJQE19}  & ICCVW'19 &
 9.6572 & 18.6686 &  57.1499&            
0.08 & 0.04 &  0.0    &            
5.72 & 2.57  &  0.0 \\
GC~\cite{DBLP:conf/iccv/YuLYSLH19}  &  ICCV'19 &
3.0548 & 12.2463  & 34.8357  &
1.95 & 0.24 & 0.01 & 
21.23 & 6.83 & 0.93 \\
RFR~\cite{DBLP:conf/cvpr/LiWZDT20}  & CVPR'20 &
6.1249 & 16.2174  & 43.2024 &            
0.15 & 0.02 & 0.0 &            
8.74 & 2.78  & 0.0 \\
CTSDG~\cite{guo2021image} & ICCV'21 &
6.8684 &  22.8353 & 63.2084  &
0.10 & 0.01  & 0.0  &
7.42 & 1.13 & 0.0 \\
LAMA~\cite{DBLP:conf/wacv/SuvorovLMRASKGP22}  & WACV'22 &
1.0816 & 7.4601 & 21.9619 &            
10.50 & 2.71 & 0.32 &            
36.65 & 20.88 & 6.62 \\
MISF~\cite{li2022misf} & CVPR'22 &
4.2603 & 9.2160  &  30.0080  &
4.86 & 0.22  & 0.01 &
12.82 & 7.80 & 1.28 \\
ZITS~\cite{DongZITS} & CVPR'22 &
\underline{1.0244} & \underline{7.1675}   & 27.2599 &
14.67 & 2.27 & 0.25 &            
38.64 & 19.63 & 5.33 \\\midrule
InvertFill~\cite{yu2022high}  &   ECCV'22 & 
1.1701 & 7.1938 & \underline{13.1982} &
\underline{18.36} & \underline{9.59} & \underline{3.10} &
\underline{35.21} & \underline{24.78} & \underline{15.59} \\
Ours  & - &      
\textbf{0.9570} & \textbf{6.6223}  & \textbf{10.9116} & 
\textbf{21.55} &  \textbf{10.46} & \textbf{6.84} & 
\textbf{41.25} &  \textbf{28.45} & \textbf{19.35}  \\ \bottomrule
\end{tabular}
}
\label{tab:main_compare_2}
\end{table*}

\section{Experiment Setup}\label{sec:experiments}

\subsection{Datasets}

{\noindent {\bf CelebA-HQ~\cite{DBLP:conf/cvpr/Lee0W020}.}} A large-scale face image dataset with $30K$ HD face images, where each image has a semantic segmentation mask corresponding to $19$ facial categories. We merge some labels into one, \eg, left-eye and right-eye into eyes, resulting in $15$ labels.

{\noindent {\bf Places2~\cite{DBLP:journals/pami/ZhouLKO018}.}} It contains $10$ million real-world photos covering more than $400$ different types of scenes. We evaluate approaches on 36,500 images of the official evaluation set.

{\noindent {\bf OST~\cite{DBLP:conf/cvpr/WangYDL18} and CityScapes~\cite{DBLP:conf/cvpr/CordtsORREBFRS16}.}} OST includes $9,900$ training images and $300$ testing images for $8$ semantic categories, which are obtained from the outdoor scene photography collection. And CityScapes dataset contains $5,000$ street view images belonging to $20$ categories. We expand the number of training images in this dataset, \ie, $2,975$ images from the training set and $1,525$ images from the test set are used for training, and $500$ images from the validation set are used for testing.

{\noindent {\bf MetFaces~\cite{DBLP:conf/nips/KarrasAHLLA20} and Scenery~\cite{yang2019very}.}} MetFaces consists of 1,336 human faces extracted from works of art. Scenery is a common benchmark for recent image outpainting tasks and contains 6,040 landscape photographs. The robustness of out-of-domain is evaluated on these two datasets. Specifically, we pre-train the CelebA-HQ model to test on MetFaces, and the Places2 model to test on Scenery.

\subsection{Implementation details}

We utilize eight A100 GPUs for pre-training the GAN generator, and one A100 GPU for optimizing the MGE and other experiments.
In the main qualitative and quantitative comparisons on CelebA-HQ and Places2, the images are resized as $512 \times 512$ resolution. As for other benchmarks, we following previous work~\cite{DBLP:conf/cvpr/LiWZDT20} to scale the image size to $256 \times 256$ as the input. In the light of the mask coverage, we classify the test masks into three difficulty levels: 10\% $\sim$ 30\%, 40\% $\sim$ 60\%, 70\% $\sim$ 90\%. All methods are tested using the identical image and mask pair to ensure a level playing field.

In this work, the updating factor $\tau$ of soft-update mean latent is set to $0.001$. In respect of the overall loss in Eq.~\ref{eq:loss_all}, we use $\lambda_{\text{msr}}=0.5, \lambda_{\text{fid}}=0.005$. We train the encoder using Adam optimizer and set the batch size to 8 and the initial learning rate to $1e^{-4}$. Further, we can see that noise is a insignificant factor in this study. The number of variables was cut by inserting Gaussian-distributed random noise into each image generation.

In this work, the layers of adaptive contexual bottlenecks in the encoder are set to $T=8$, where the dilation rates of convolutions in each bottleneck are empirically set to $r \in R=\{1,2,3,4\}$. In multi-scale spatial-aware attention, the patch size is set to $h=w=\{2,4\}$.

\subsection{Evaluation Metrics}

Similar to the previous works \cite{DBLP:conf/eccv/LiaoXWLS20,DBLP:journals/corr/abs-2104-01431}, we employ three metrics, and Structural Similarity Index (SSIM)~\cite{DBLP:journals/tip/WangBSS04}, and LPIPS~\cite{DBLP:conf/cvpr/ZhangIESW18}. Unfortunately, this kind of metrics suitable for evaluating the pixel-wise similarity, but fails to assess the fidelity in such ill-posed generation task, especially for extensive mask coverage.

\newcommand{\MM}{0.133}
\renewcommand\arraystretch{0.8}
\begin{figure*}[!th]
	\centering
	\begin{tabular}{c@{\hspace{1.0mm}}c@{\hspace{1.0mm}}c@{\hspace{1.mm}}c@{\hspace{1.0mm}}c@{\hspace{1.0mm}}c@{\hspace{1.0mm}}c}
	\small{Masked} & \small{DeepFillv2} & \small{CTSDG} & \small{ICT} & \small{Score-SDE} & \small{LAMA} & \small{Ours} \\

	\frame{\includegraphics[width=\MM\linewidth]{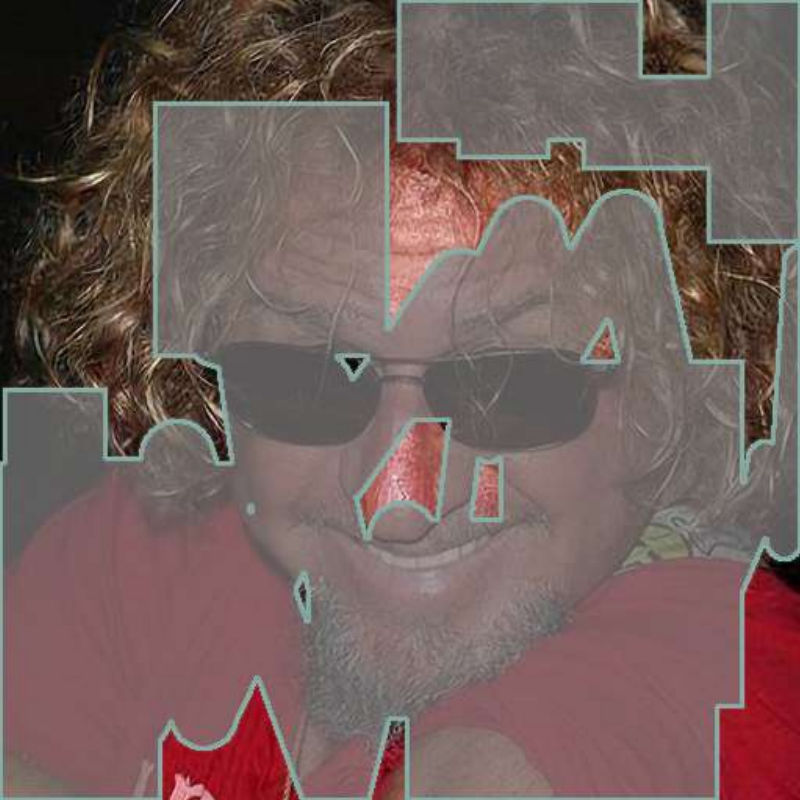}} &
	\frame{\includegraphics[width=\MM\linewidth]{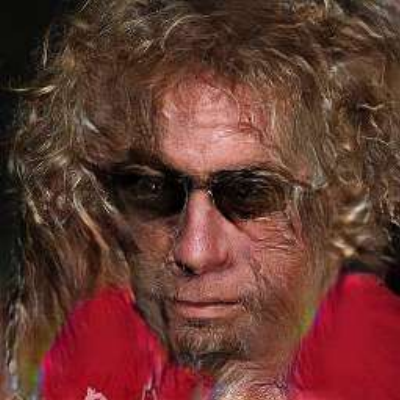}} &
	\frame{\includegraphics[width=\MM\linewidth]{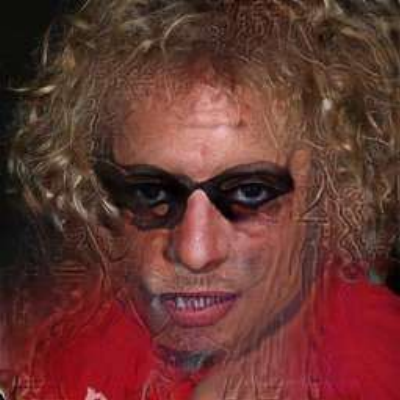}} &
	\frame{\includegraphics[width=\MM\linewidth]{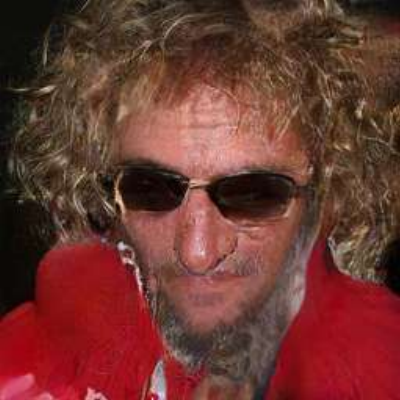}} &
	\frame{\includegraphics[width=\MM\linewidth]{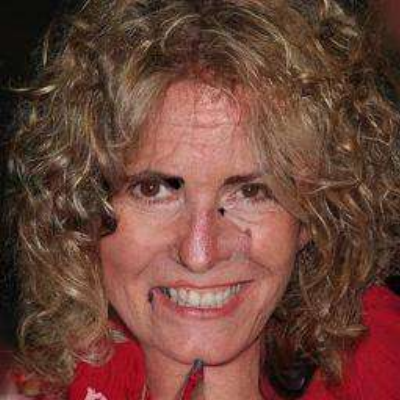}} &
	\frame{\includegraphics[width=\MM\linewidth]{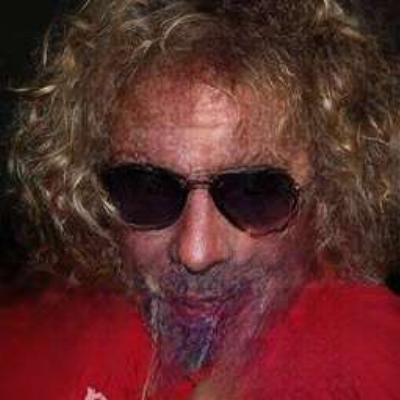}} &
	\frame{\includegraphics[width=\MM\linewidth]{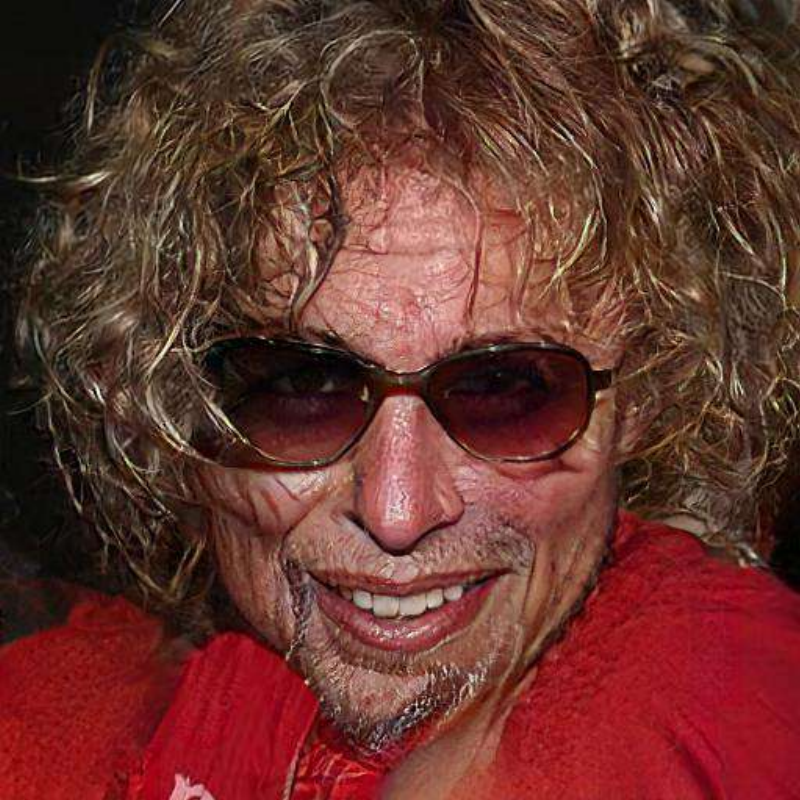}} \\
	
	\frame{\includegraphics[width=\MM\linewidth]{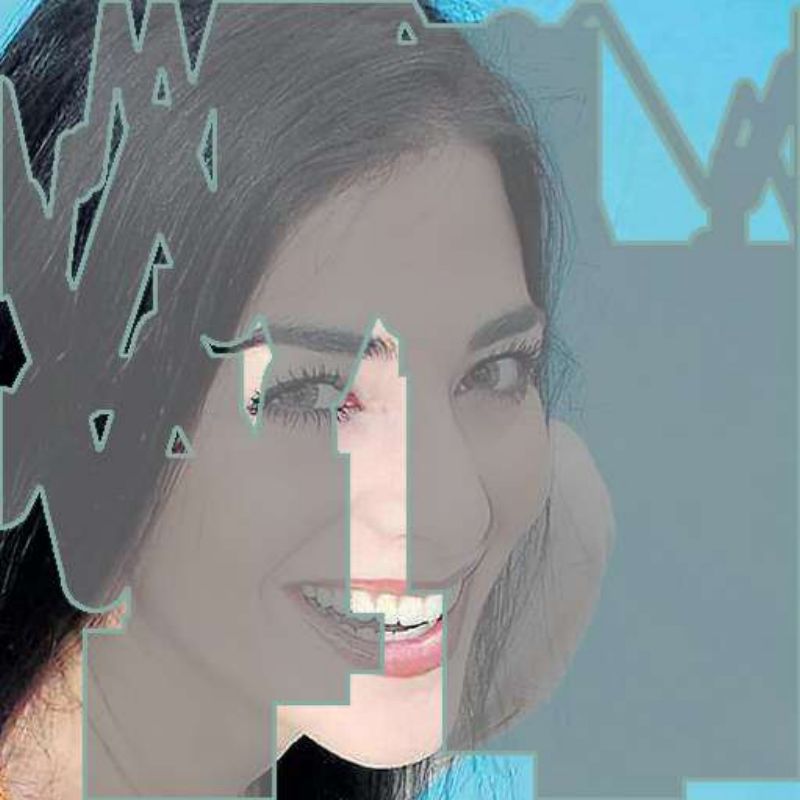}} &
	\frame{\includegraphics[width=\MM\linewidth]{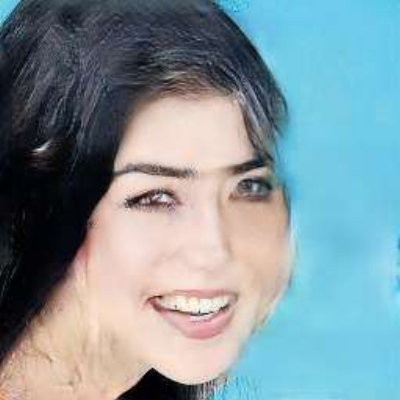}} &
	\frame{\includegraphics[width=\MM\linewidth]{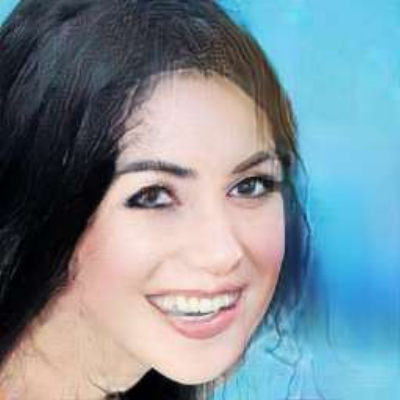}} &
	\frame{\includegraphics[width=\MM\linewidth]{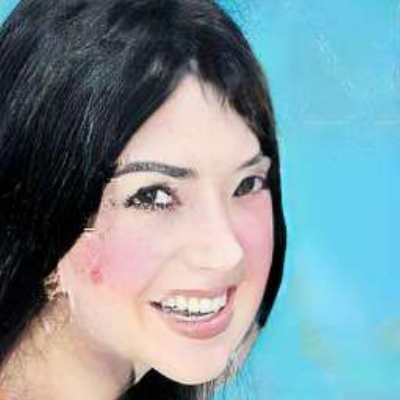}} &
	\frame{\includegraphics[width=\MM\linewidth]{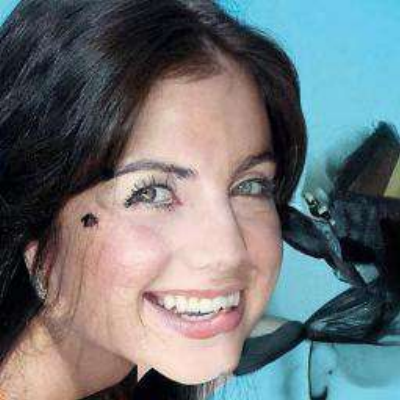}} &
	\frame{\includegraphics[width=\MM\linewidth]{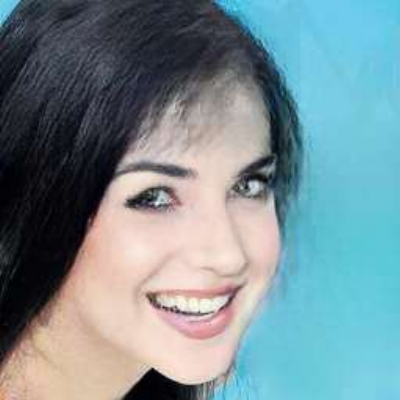}} &
	\frame{\includegraphics[width=\MM\linewidth]{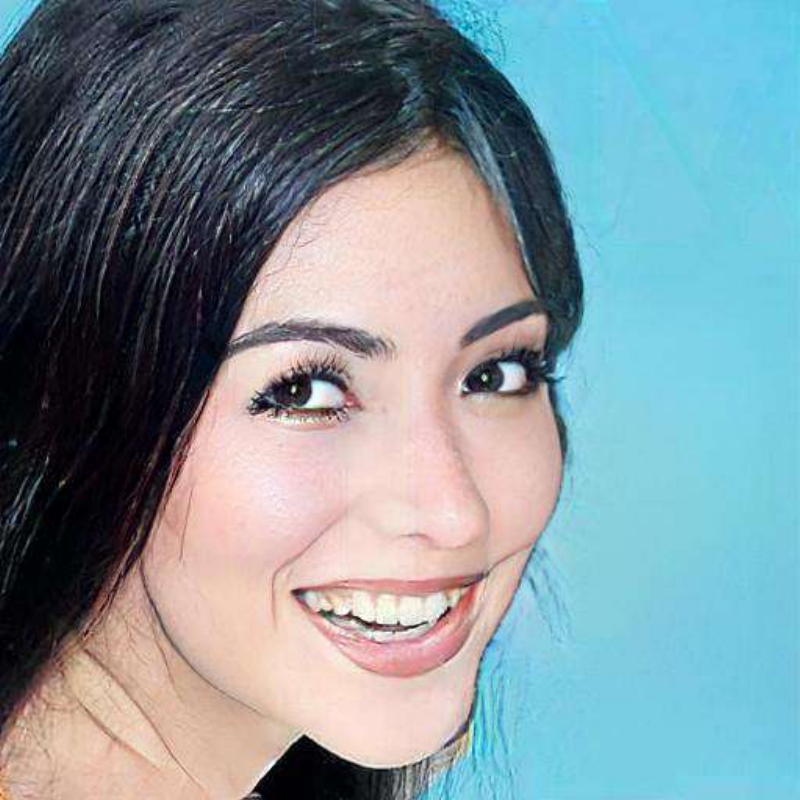}}  \\
	
	\frame{\includegraphics[width=\MM\linewidth]{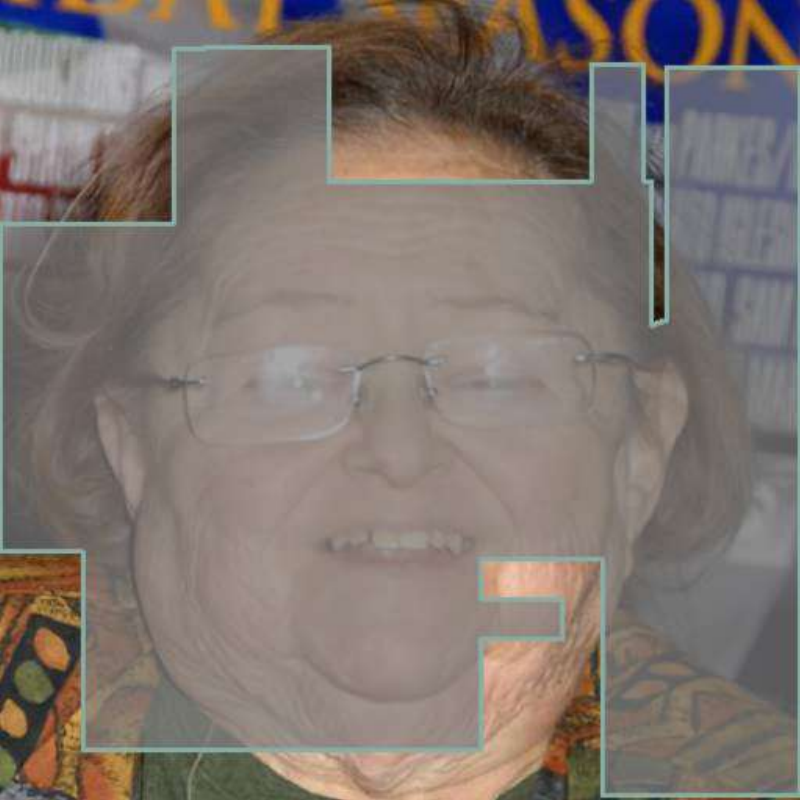}} &
	\frame{\includegraphics[width=\MM\linewidth]{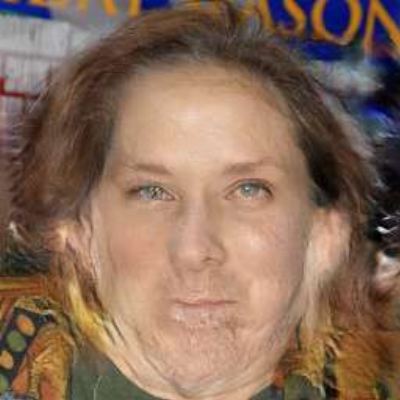}} &
	\frame{\includegraphics[width=\MM\linewidth]{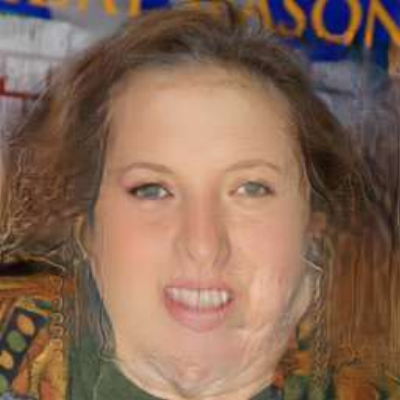}} &
	\frame{\includegraphics[width=\MM\linewidth]{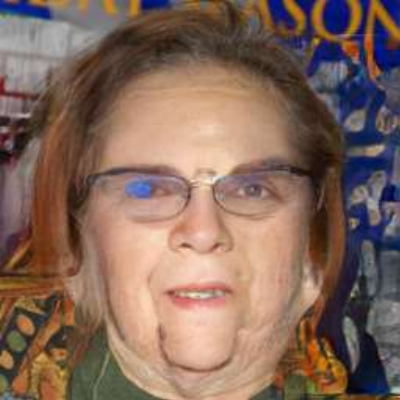}} &
	\frame{\includegraphics[width=\MM\linewidth]{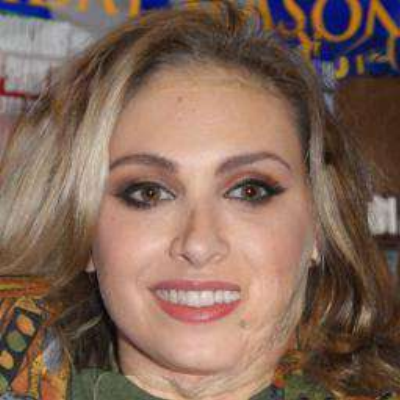}} &
	\frame{\includegraphics[width=\MM\linewidth]{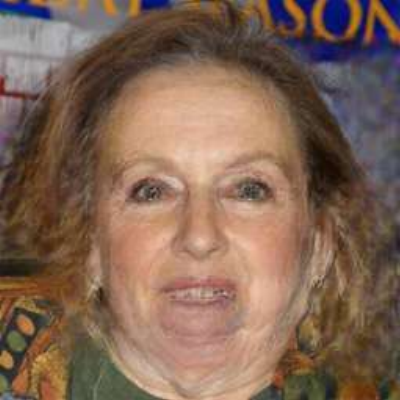}} &
	\frame{\includegraphics[width=\MM\linewidth]{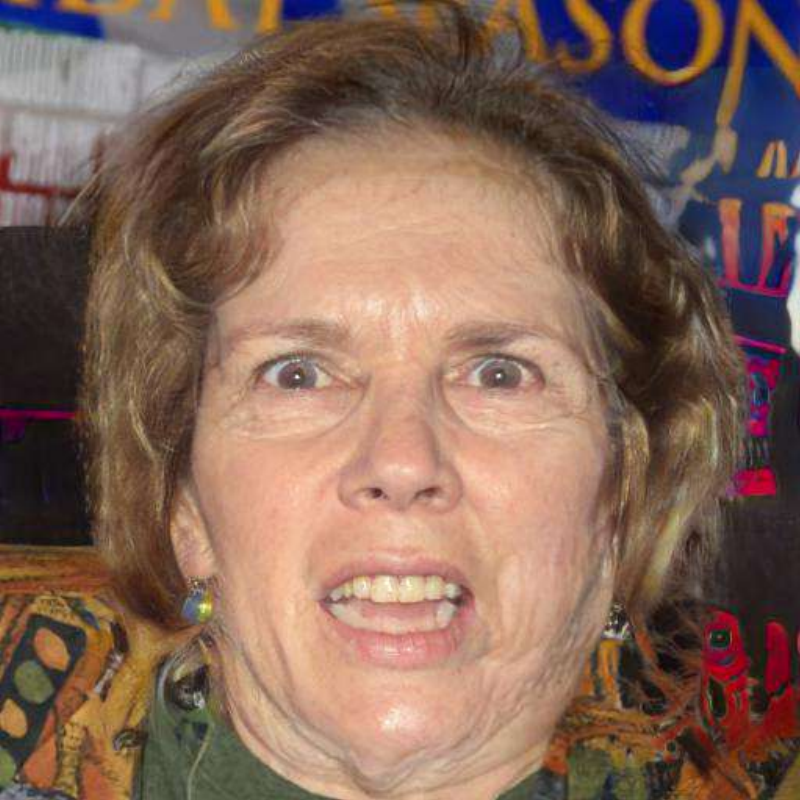}}  \\

\end{tabular}
\caption{Qualitative comparison with the state-of-the-arts on CelebA-HQ dataset. }
\label{fig:visual_compare_chq}
\end{figure*}

\newcommand{\MMM}{0.133}
\renewcommand\arraystretch{0.8}
\begin{figure*}[!th]
	\centering
	\begin{tabular}{c@{\hspace{1mm}}c@{\hspace{1mm}}c@{\hspace{1mm}}c@{\hspace{1mm}}c@{\hspace{1mm}}c@{\hspace{1mm}}c}
	\small{Masked} & \small{DeepFillv2} & \small{LAMA} & \small{MISF} & \small{RePaint} & \small{ZITS} & \small{Ours} \\
    
    \frame{\includegraphics[width=\MMM\linewidth]{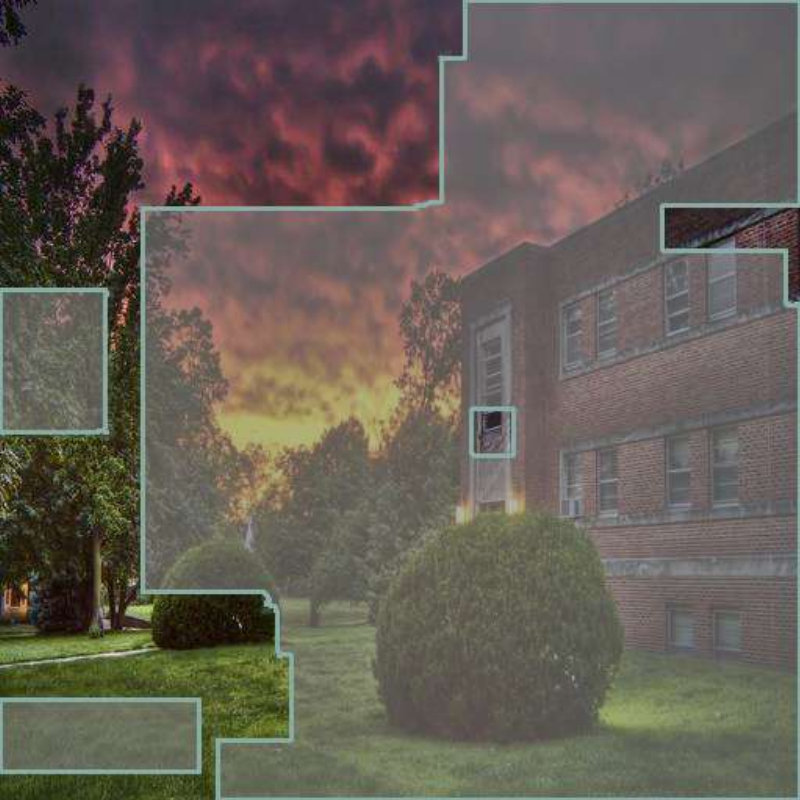}} &
	\frame{\includegraphics[width=\MMM\linewidth]{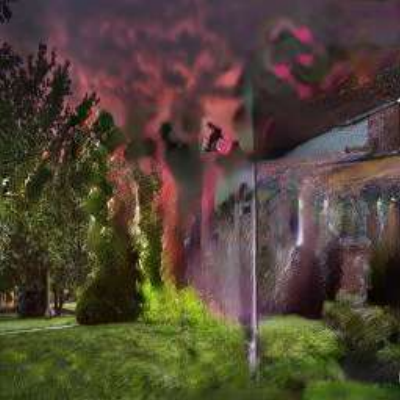}} &
	\frame{\includegraphics[width=\MMM\linewidth]{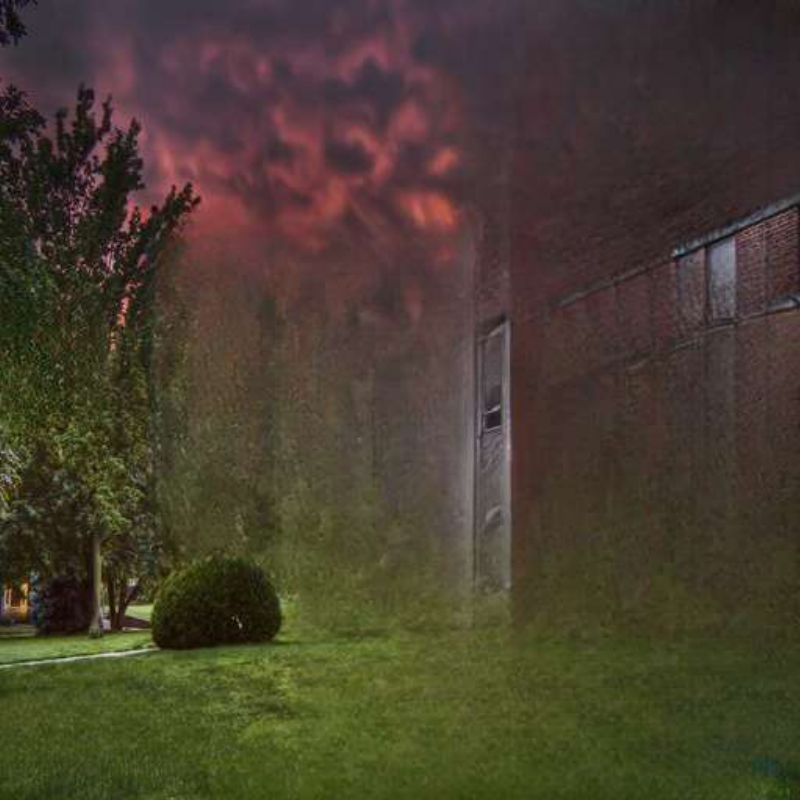}} &
	\frame{\includegraphics[width=\MMM\linewidth]{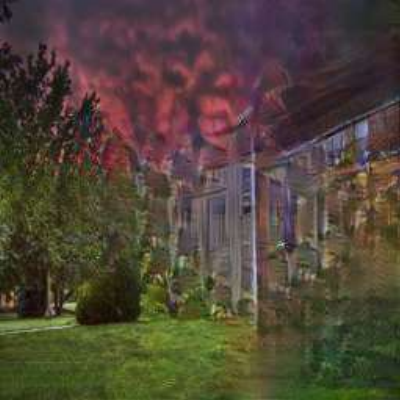}} &
	\frame{\includegraphics[width=\MMM\linewidth]{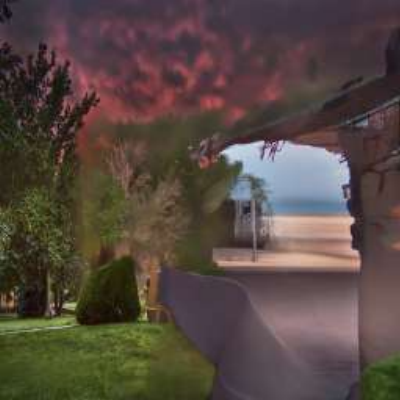}} &
	\frame{\includegraphics[width=\MMM\linewidth]{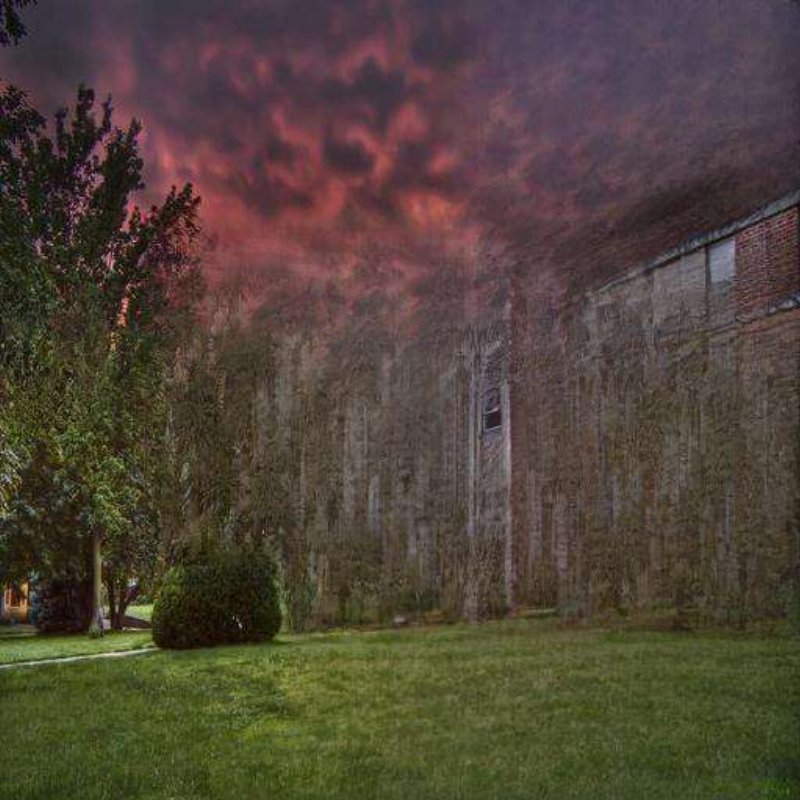}} &
	\frame{\includegraphics[width=\MMM\linewidth]{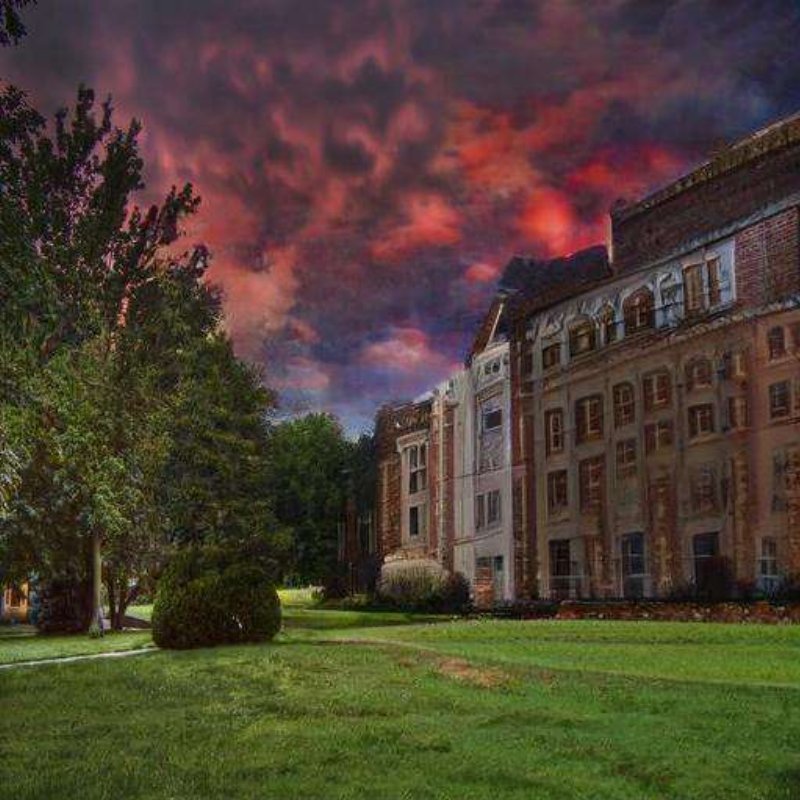}} \\
	
\frame{\includegraphics[width=\MMM\linewidth]{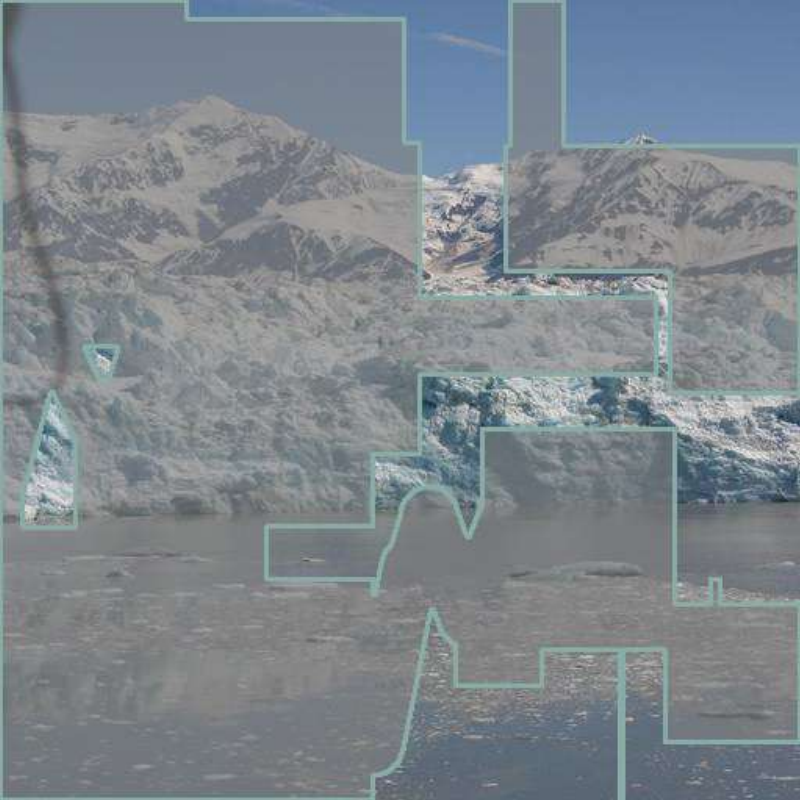}} &
\frame{\includegraphics[width=\MMM\linewidth]{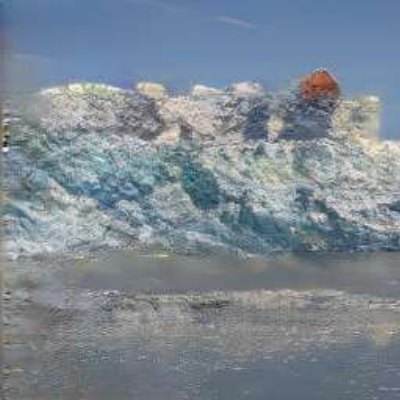}} &
\frame{\includegraphics[width=\MMM\linewidth]{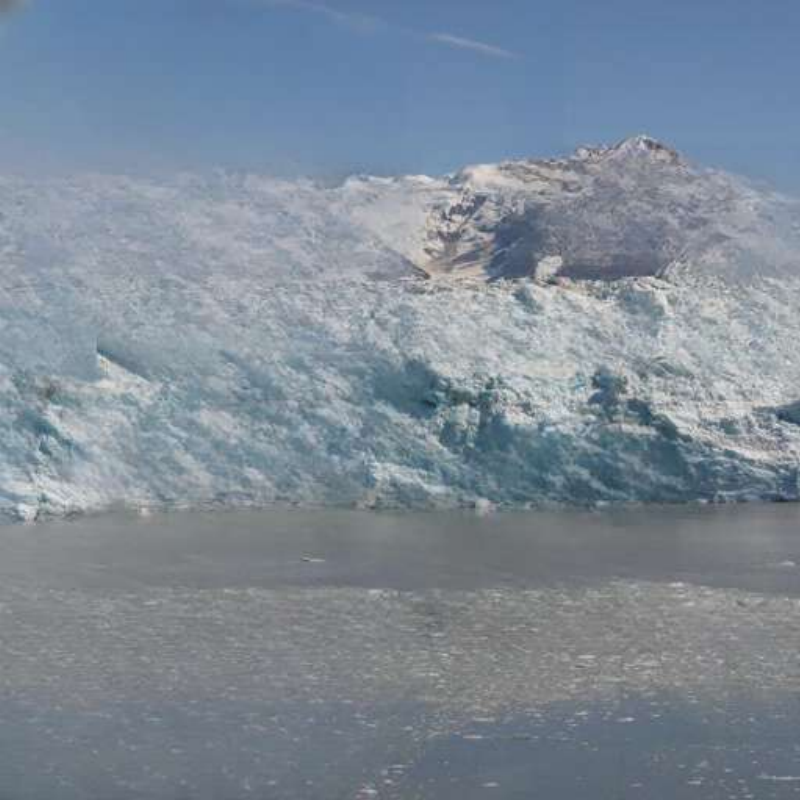}} &
\frame{\includegraphics[width=\MMM\linewidth]{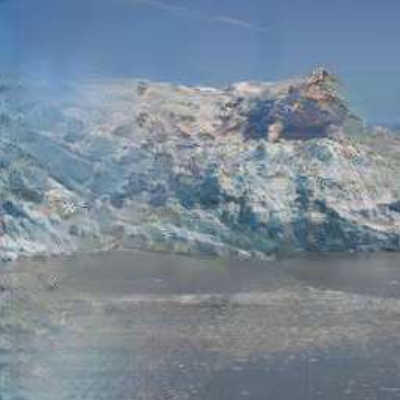}} &
\frame{\includegraphics[width=\MMM\linewidth]{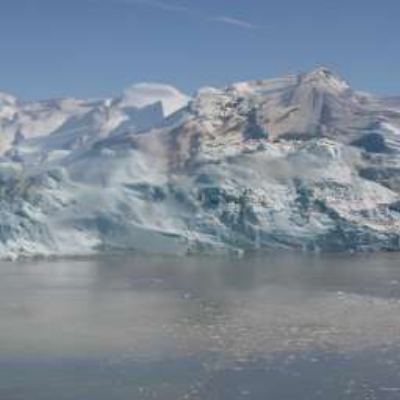}} &
\frame{\includegraphics[width=\MMM\linewidth]{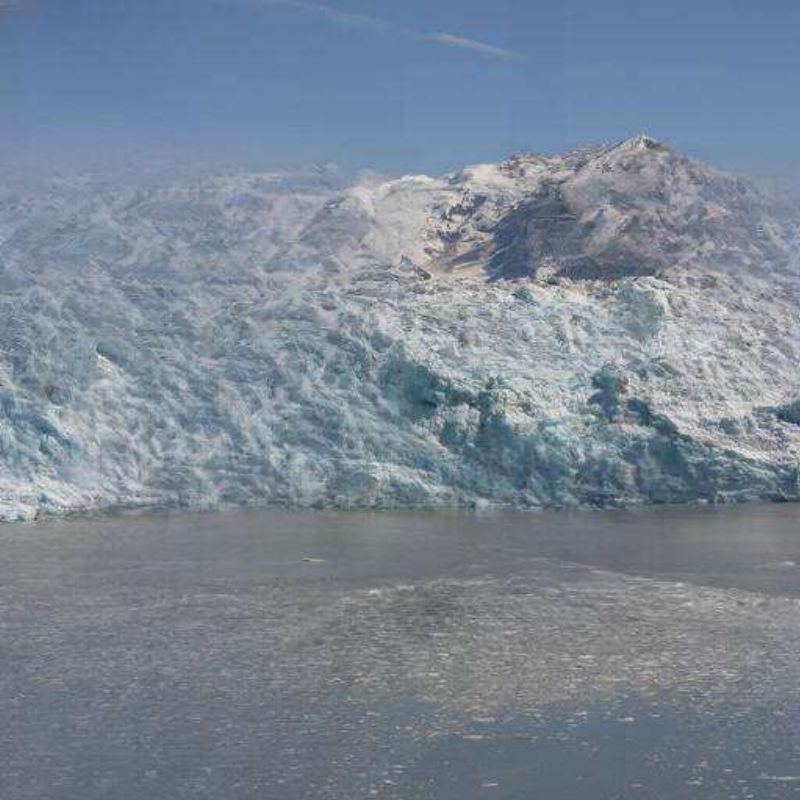}} &
\frame{\includegraphics[width=\MMM\linewidth]{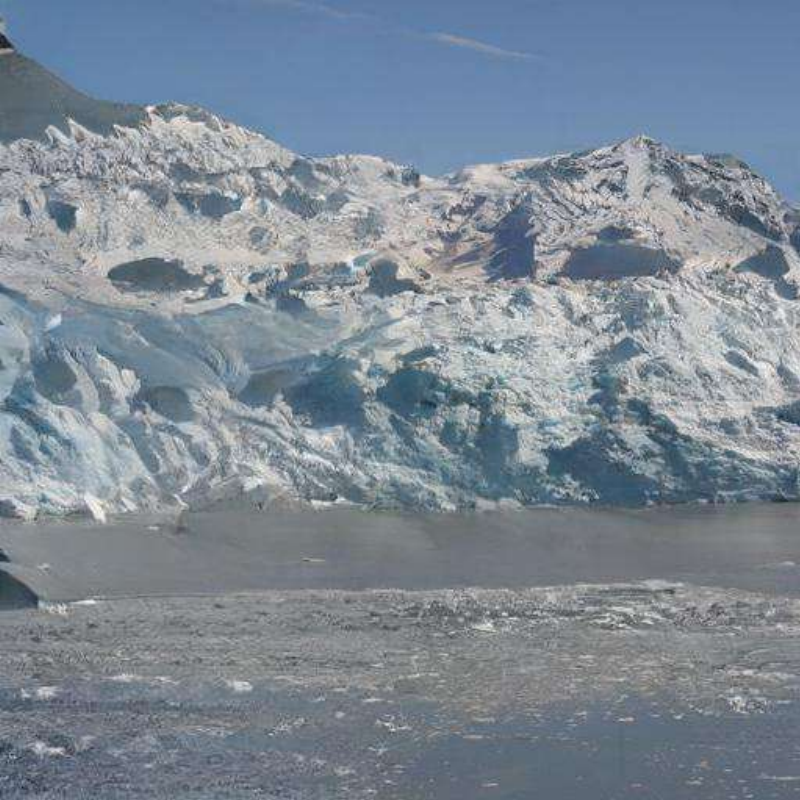}} \\
	
\frame{\includegraphics[width=\MMM\linewidth]{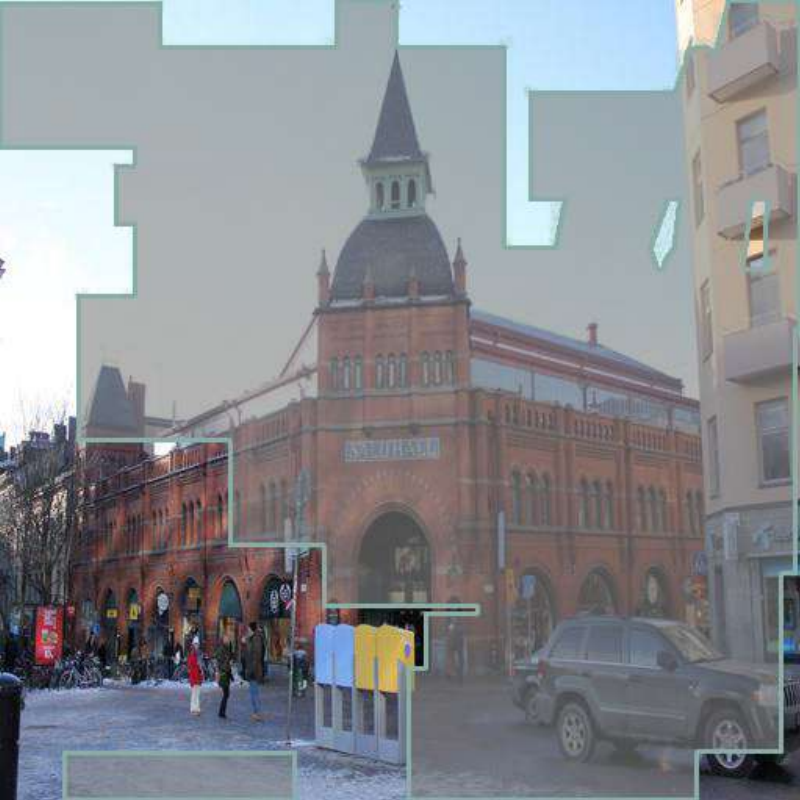}} &
\frame{\includegraphics[width=\MMM\linewidth]{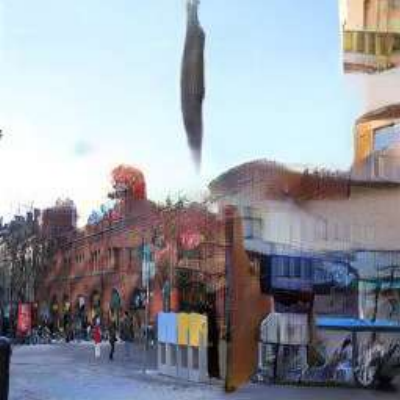}} &
\frame{\includegraphics[width=\MMM\linewidth]{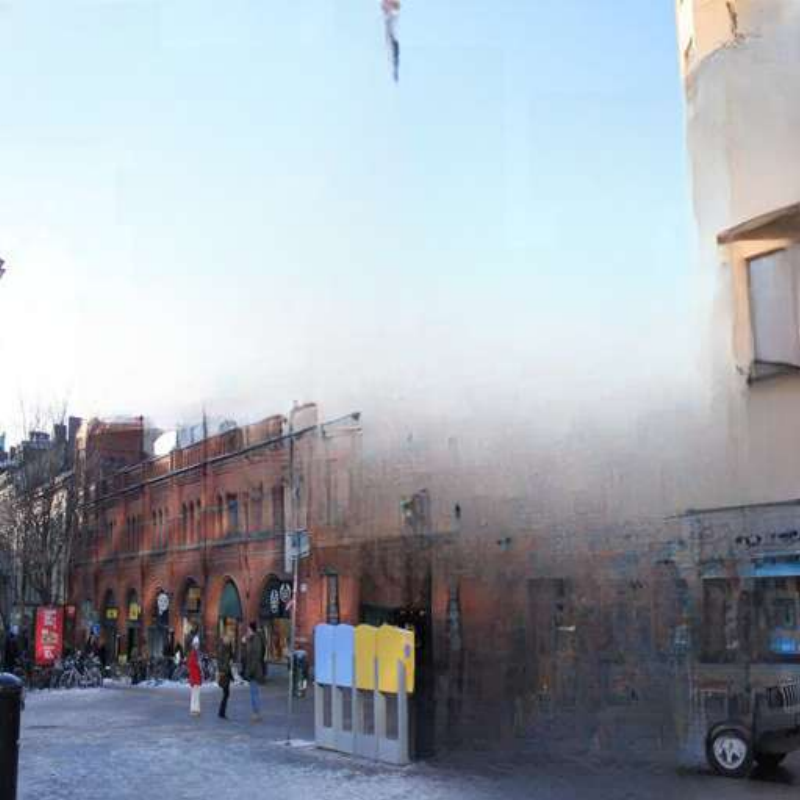}} &
\frame{\includegraphics[width=\MMM\linewidth]{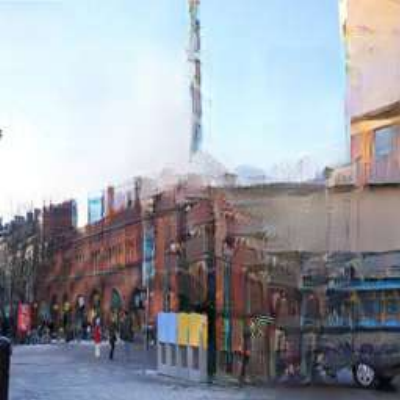}} &
\frame{\includegraphics[width=\MMM\linewidth]{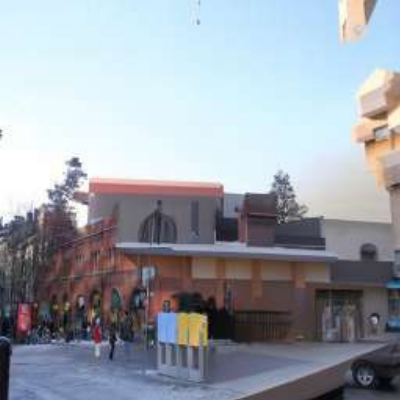}} &
\frame{\includegraphics[width=\MMM\linewidth]{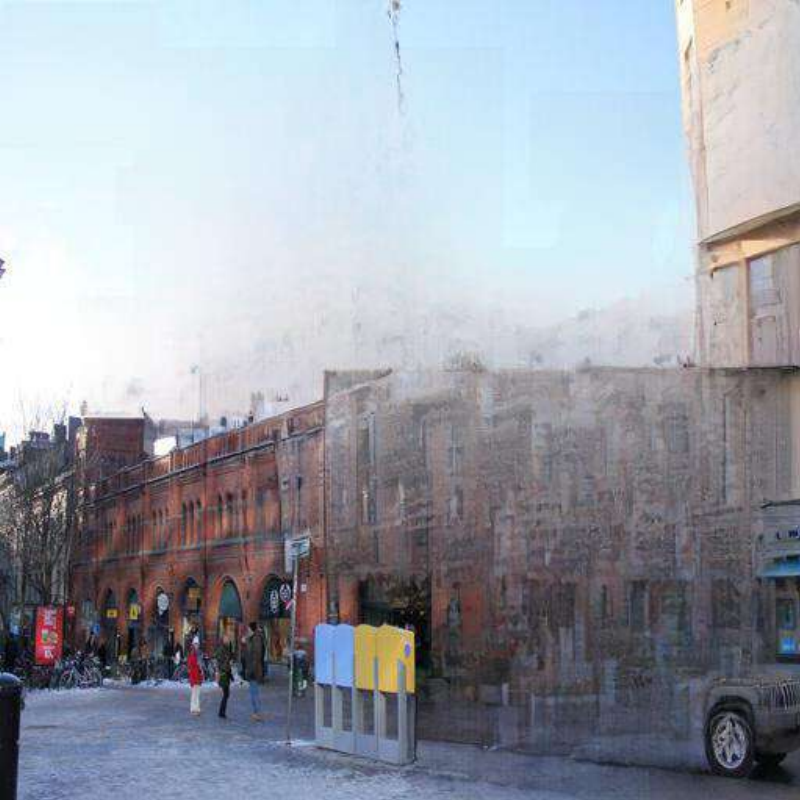}} &
\frame{\includegraphics[width=\MMM\linewidth]{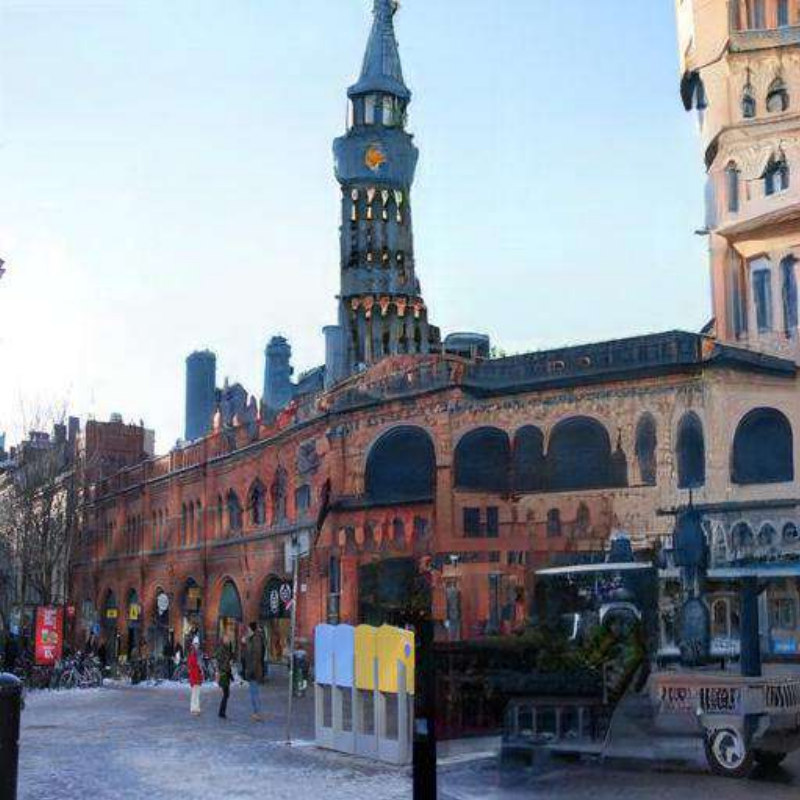}} \\

\end{tabular}
\caption{Qualitative comparison with the state-of-the-arts on Places2 dataset. }
\label{fig:visual_compare_places}
\end{figure*}

Frechet Inception Distance (FID)~\cite{DBLP:conf/nips/HeuselRUNH17}, P-IDS~\cite{DBLP:conf/iclr/ZhaoCSDLCX21}, U-IDS~\cite{DBLP:conf/iclr/ZhaoCSDLCX21} are deep metrics and closer to human perception. It measures the distribution distance with a pre-trained inception model, which better captures distortions. Notably, the Inception network~\cite{DBLP:conf/nips/SalimansGZCRCC16} is used to extract the image features when calculating the FID score, and then calculate its mean and covariance matrix to estimate the distance between the ground-truth and generated data distribution. Due to the limited number of test data images in CelebA-HQ, the SVM-based metrics U-IDS and P-IDS result in all methods scoring zero on these metrics (i.e., SVM underfitting). Therefore, we only tested these two metrics on Places2.

\subsection{Baselines}

To showcase the benefits and advantages of our approach, we meticulously pick baselines from two primary directions: U-Net style methods and inversion-based methods. We begin by comparing our approach to certain established methods including EC~\cite{DBLP:conf/iccvw/NazeriNJQE19}, GC~\cite{DBLP:conf/cvpr/Yu0YSLH18}, RFR~\cite{DBLP:conf/cvpr/LiWZDT20}, CTSDG~\cite{guo2021image}, ICT~\cite{Wan_2021_ICCV}, Score-SDE~\cite{DBLP:conf/iclr/0011SKKEP21},
LAMA~\cite{DBLP:conf/wacv/SuvorovLMRASKGP22},
MISF~\cite{li2022misf}, and
ZITS~\cite{DongZITS}, for the purpose of demonstrating its capacity to fill images hidden by expansive masks.
By contrasting our method to the latest GAN inversion-based inpainting methods mGANprior~\cite{gu2020mganprior} and pSp~\cite{richardson2021encoding}, we show that it effectively resolves the ``gapping'' problem while also improving fidelity. 

We use the official implementations and pre-trained weights of the aforementioned baselines to reproduce.

\newcommand{\M}{0.13}
\renewcommand\arraystretch{0.8}
\begin{figure*}[!th]
	\centering
	\begin{tabular}{c@{\hspace{1.0mm}}c@{\hspace{1.0mm}}c@{\hspace{1.mm}}c@{\hspace{1.0mm}}c@{\hspace{1.0mm}}c@{\hspace{1.0mm}}c}
	\small{Masked} & \small{\makecell[c]{mGANprior \\ w/o Comp.}} & \small{\makecell[c]{mGANprior \\ w/ Comp.}} & \small{pSp w/o Comp.} & \small{pSp w/ Comp.} & \small{\makecell[c]{pSp w/ Blending}} & \small{Ours} \\

	\frame{\includegraphics[width=\M\linewidth]{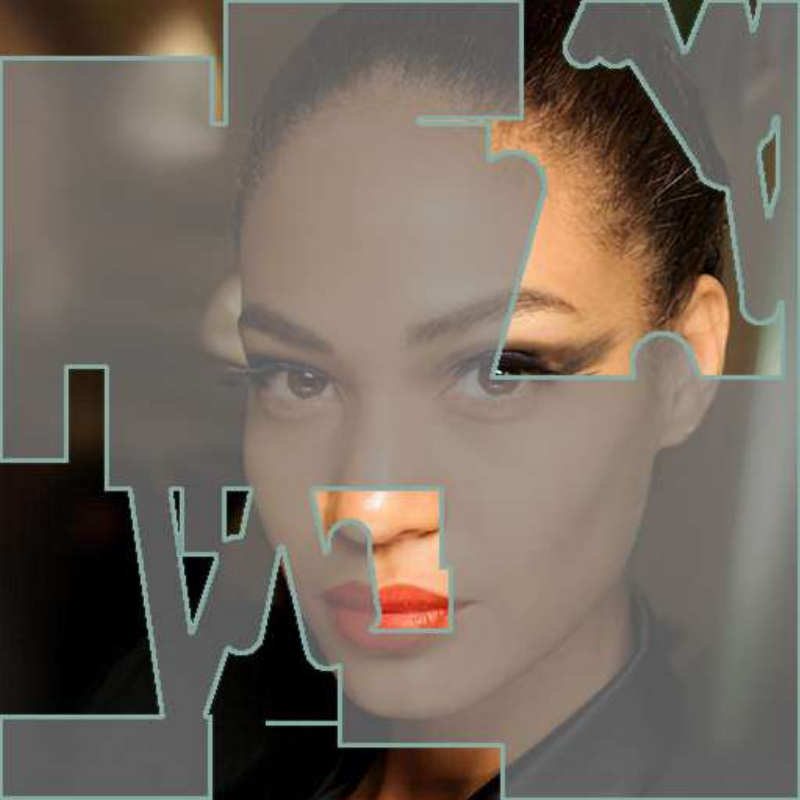}} &
	\frame{\includegraphics[width=\M\linewidth]{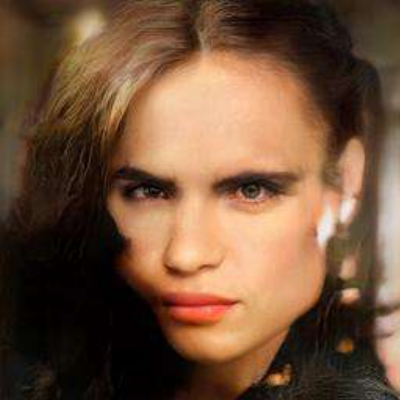}} &
	\frame{\includegraphics[width=\M\linewidth]{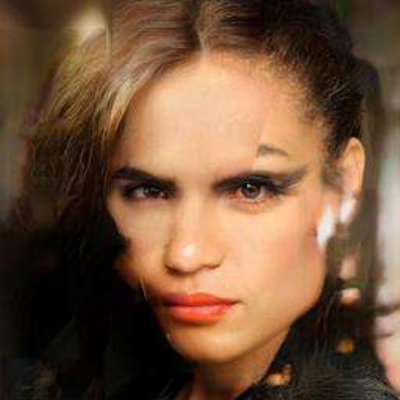}} &
	\frame{\includegraphics[width=\M\linewidth]{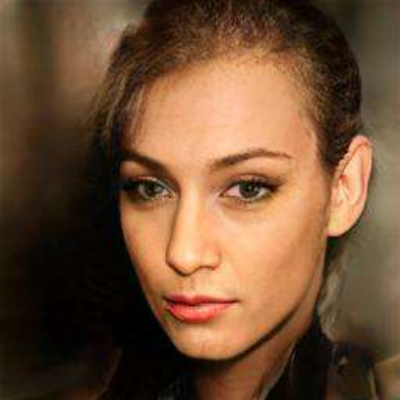}} &
	\frame{\includegraphics[width=\M\linewidth]{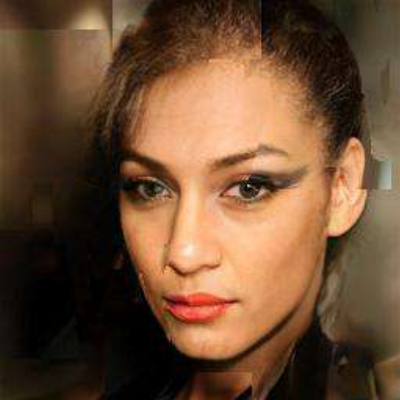}} &
	\frame{\includegraphics[width=\M\linewidth]{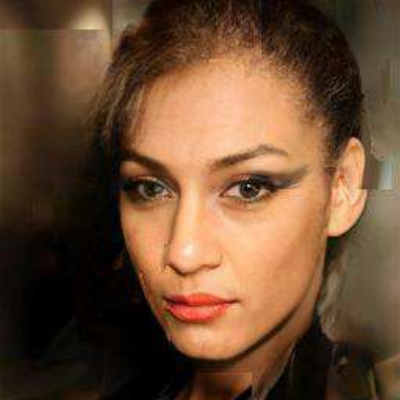}}&
	\frame{\includegraphics[width=\M\linewidth]{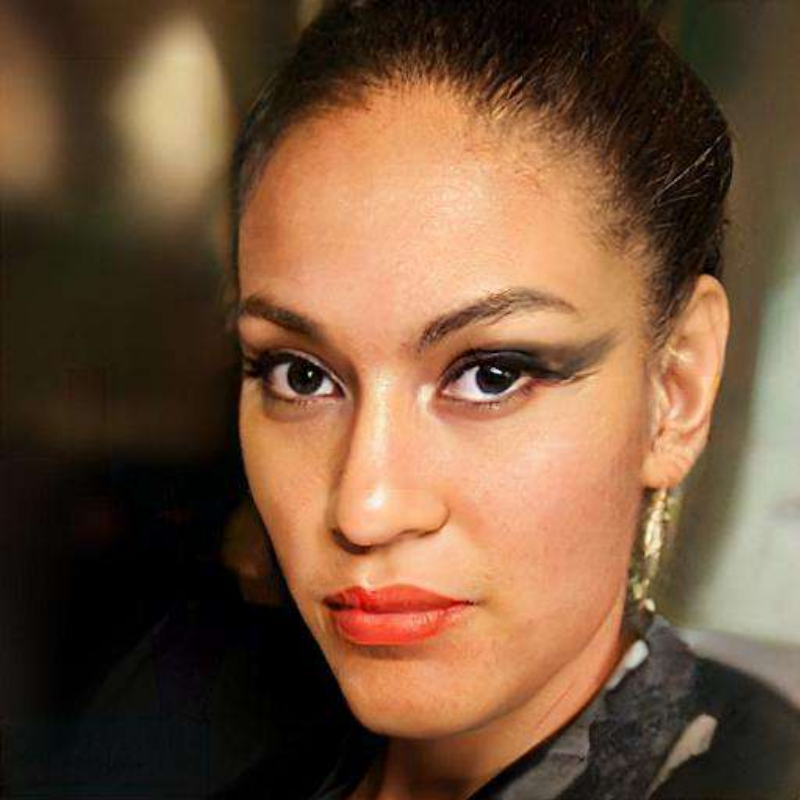}} \\
	
	\frame{\includegraphics[width=\M\linewidth]{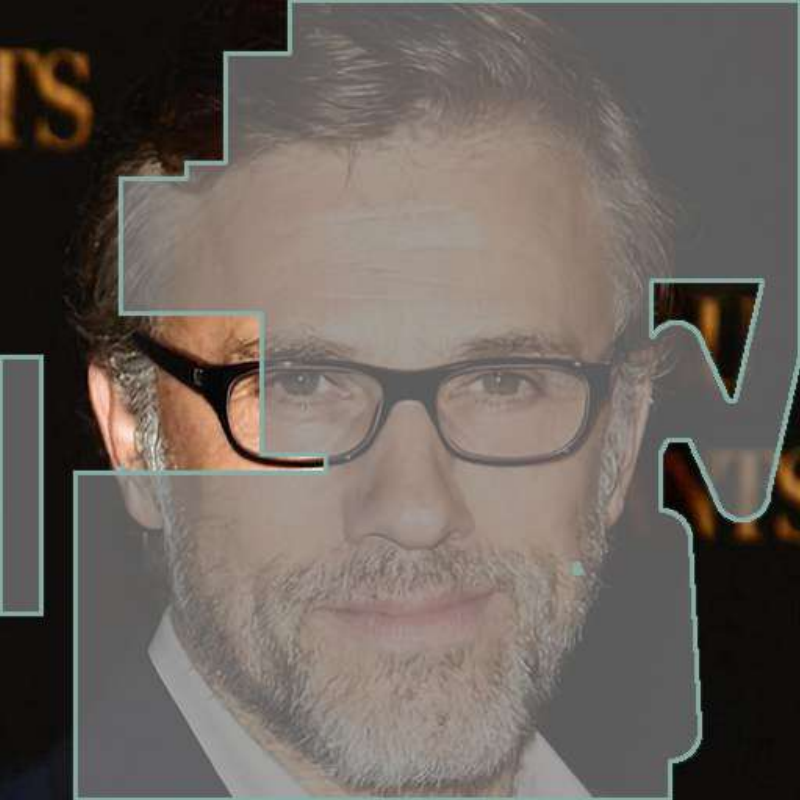}} &
	\frame{\includegraphics[width=\M\linewidth]{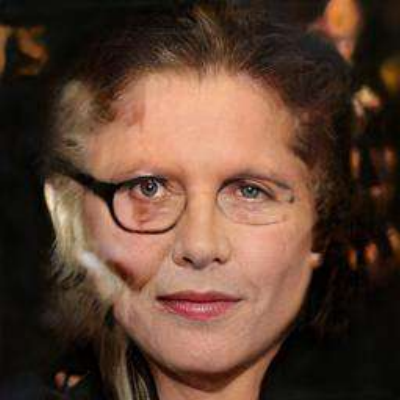}} &
	\frame{\includegraphics[width=\M\linewidth]{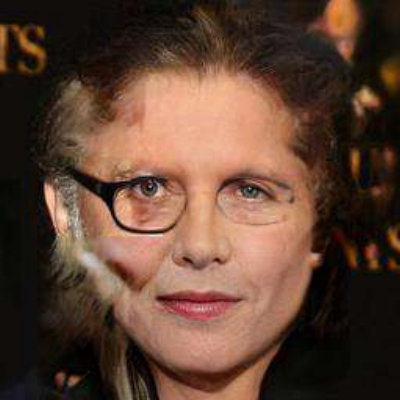}} &
	\frame{\includegraphics[width=\M\linewidth]{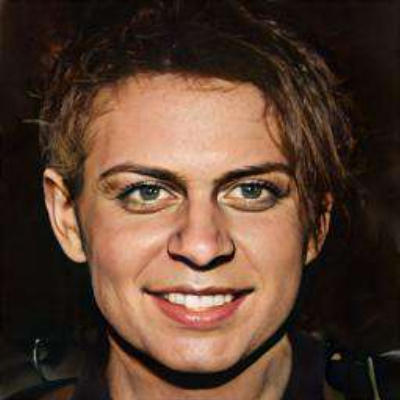}} &
	\frame{\includegraphics[width=\M\linewidth]{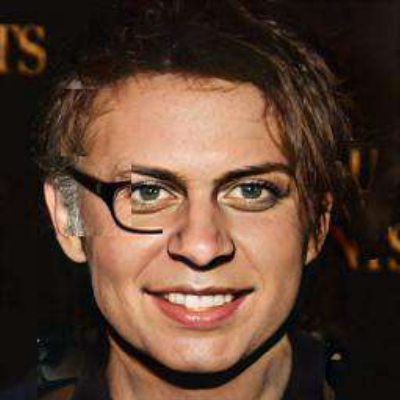}} &
	\frame{\includegraphics[width=\M\linewidth]{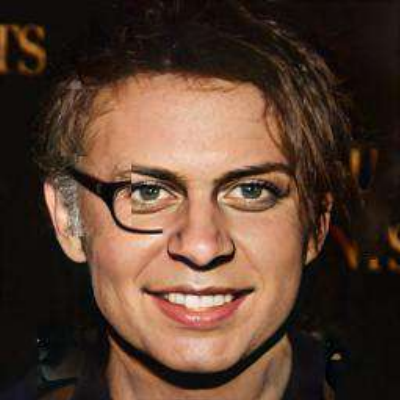}}&
	\frame{\includegraphics[width=\M\linewidth]{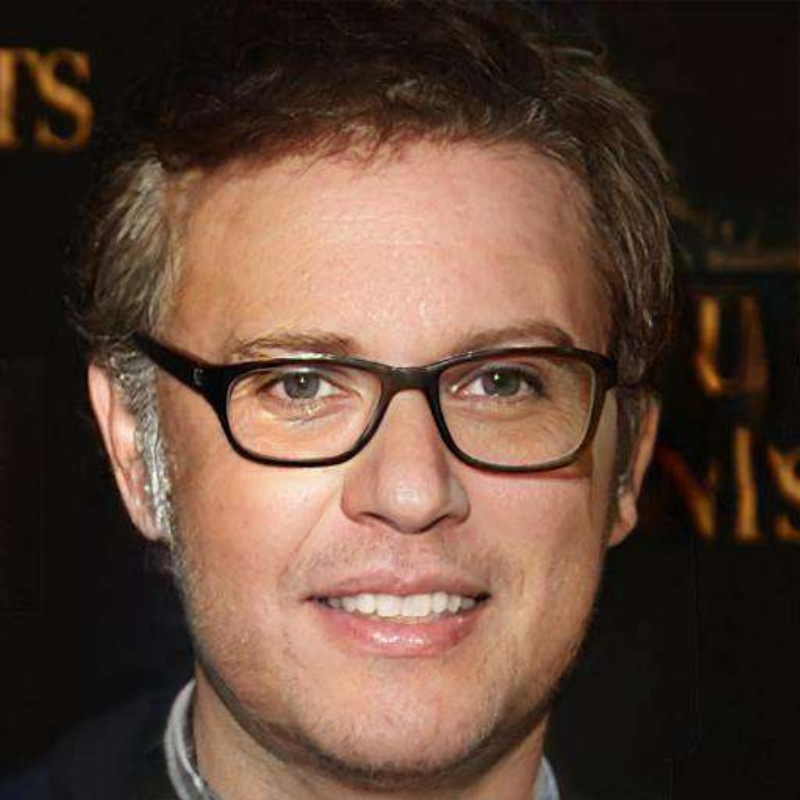}}  \\

\end{tabular}
\caption{Qualitative comparison with advanced GAN inversion state-of-the-art methods on the CelebA-HQ dataset to demonstrate the importance of addressing the "gapping" issue.}
\label{fig:visual_compare_psp}
\end{figure*}

\newcommand{\PLOTB}{0.28}
\renewcommand\arraystretch{0.8}
\begin{figure*}[!th]
	\centering
	\begin{tabular}{c@{\hspace{1.0mm}}c@{\hspace{1.0mm}}c}

	\includegraphics[width=\PLOTB\linewidth]{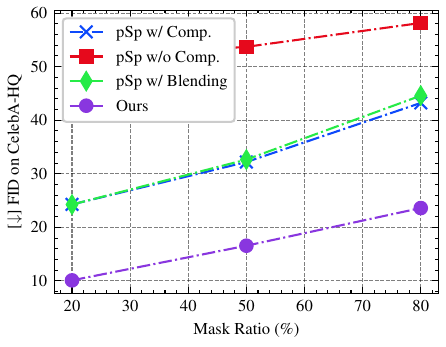} &
	\includegraphics[width=\PLOTB\linewidth]{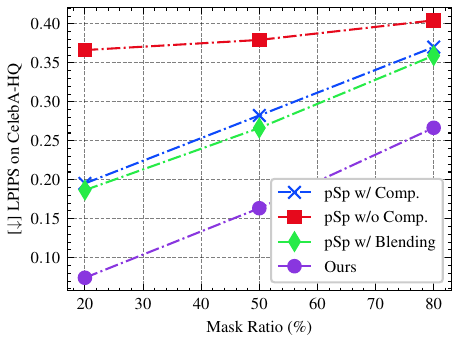} &
	\includegraphics[width=\PLOTB\linewidth]{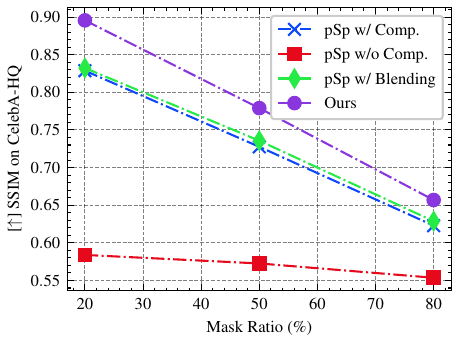} \\
	\small{(a) FID comparison on CelebA-HQ}  & \small{(b) LPIPS comparison on CelebA-HQ} & \small{(c) SSIM comparison on CelebA-HQ} \\

\end{tabular}
\caption{Importance of solving ``gapping'' issue. 'w/ Comp.' means compositing model's generation and unmasked region of ground truth, and 'w/ Blending' means utilizing a image blending post-processing~\cite{wu2021gpgan}. $[\uparrow]$ stands for higher is better, $[\downarrow]$ stands for lower is better.}
\label{fig:plots_compare_psp}
\end{figure*}

\newcommand{\DOMAIN}{0.3}
\renewcommand\arraystretch{0.8}
\begin{figure}[!th]
	\centering
	\begin{tabular}{c@{\hspace{1.0mm}}c@{\hspace{1.0mm}}c@{\hspace{2.mm}}}
	\small{Ground Truth} & \small{Input} & \small{Ours}  \\
	\frame{\includegraphics[width=\DOMAIN\linewidth]{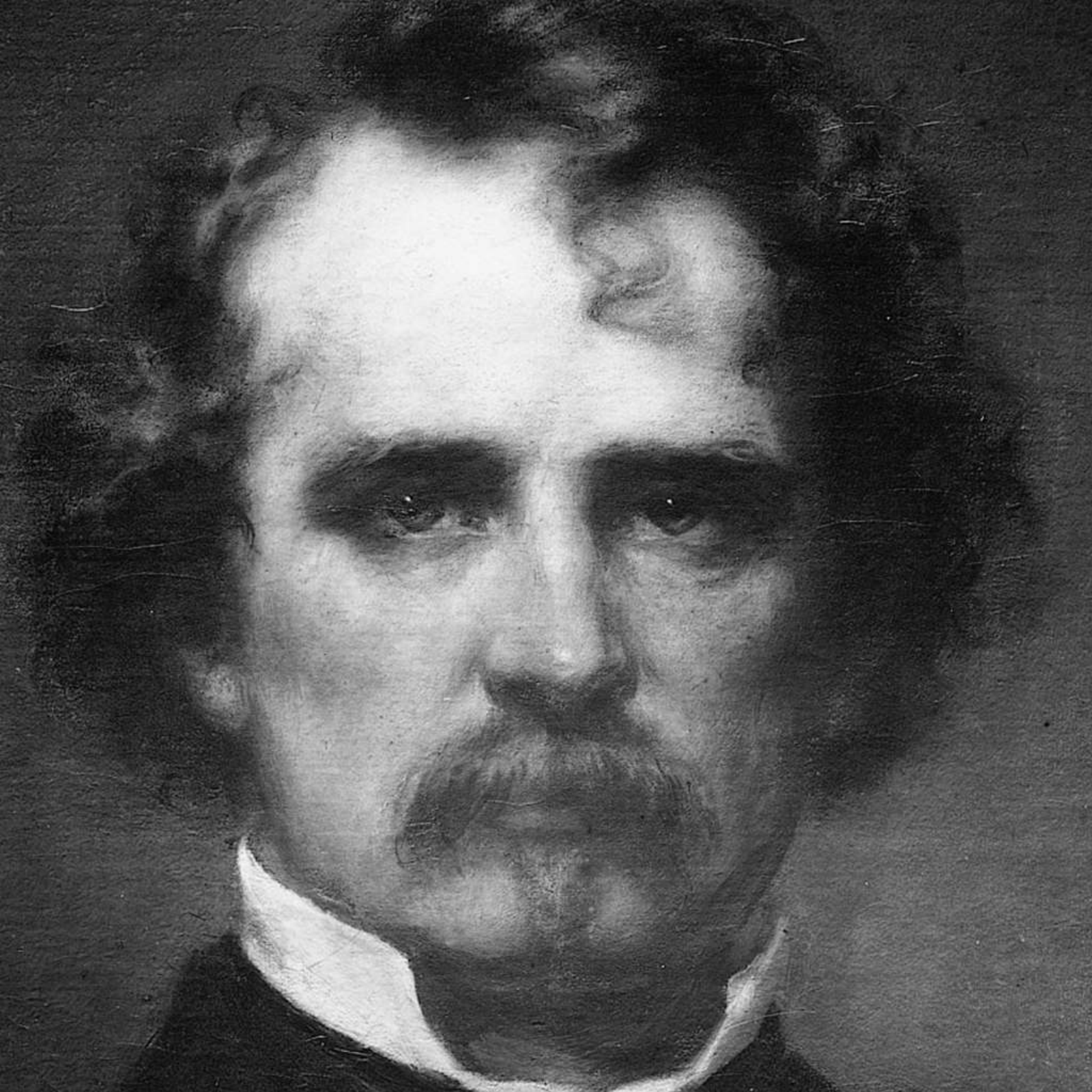}} &
	\frame{\includegraphics[width=\DOMAIN\linewidth]{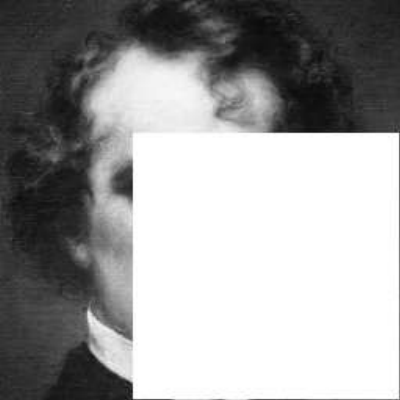}} &
	\frame{\includegraphics[width=\DOMAIN\linewidth]{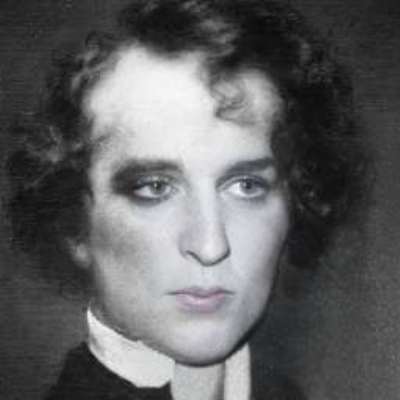}} 
	\\
	
	\frame{\includegraphics[width=\DOMAIN\linewidth]{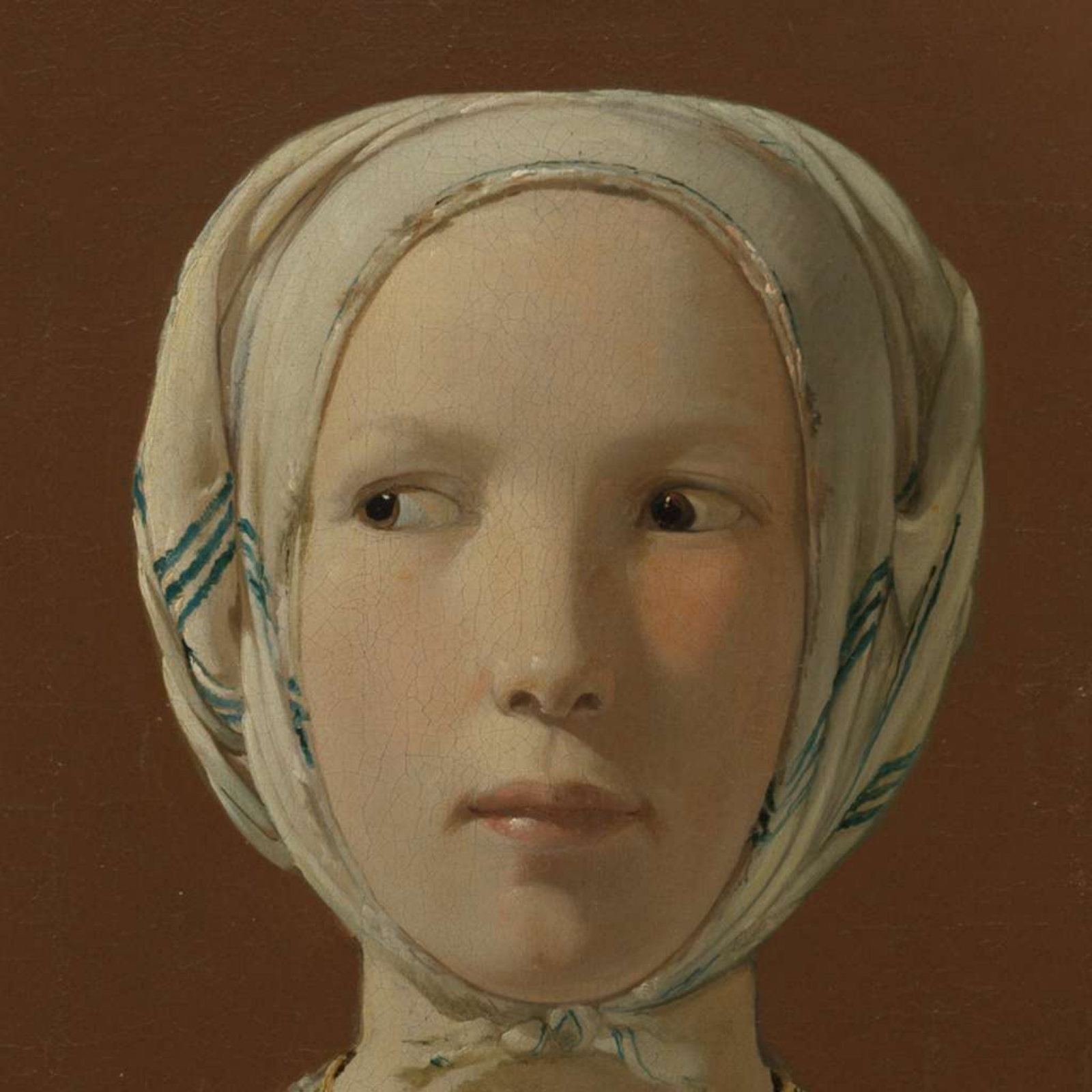}} &
	\frame{\includegraphics[width=\DOMAIN\linewidth]{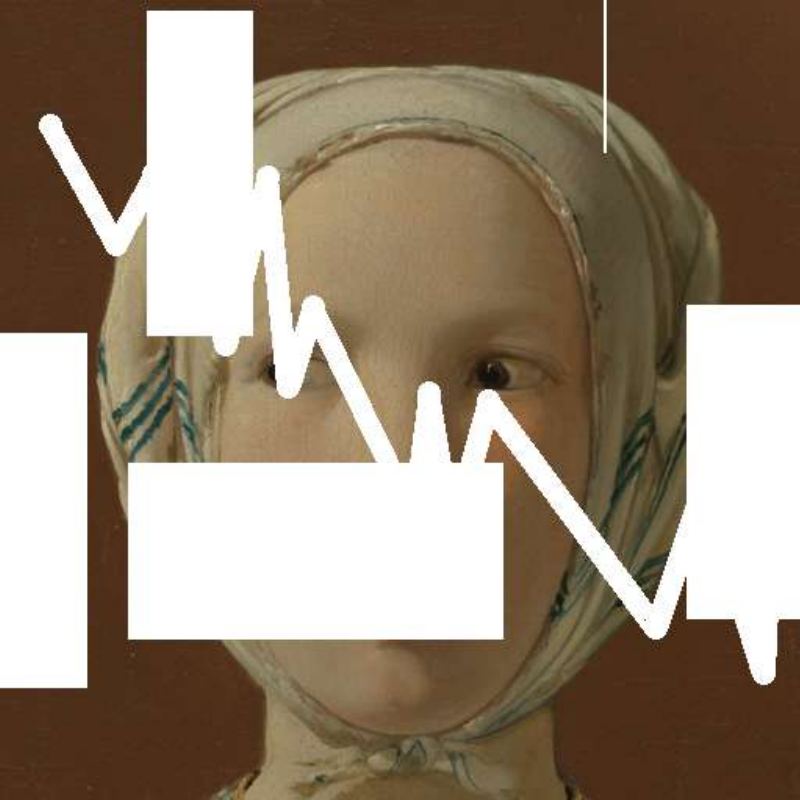}} &
	\frame{\includegraphics[width=\DOMAIN\linewidth]{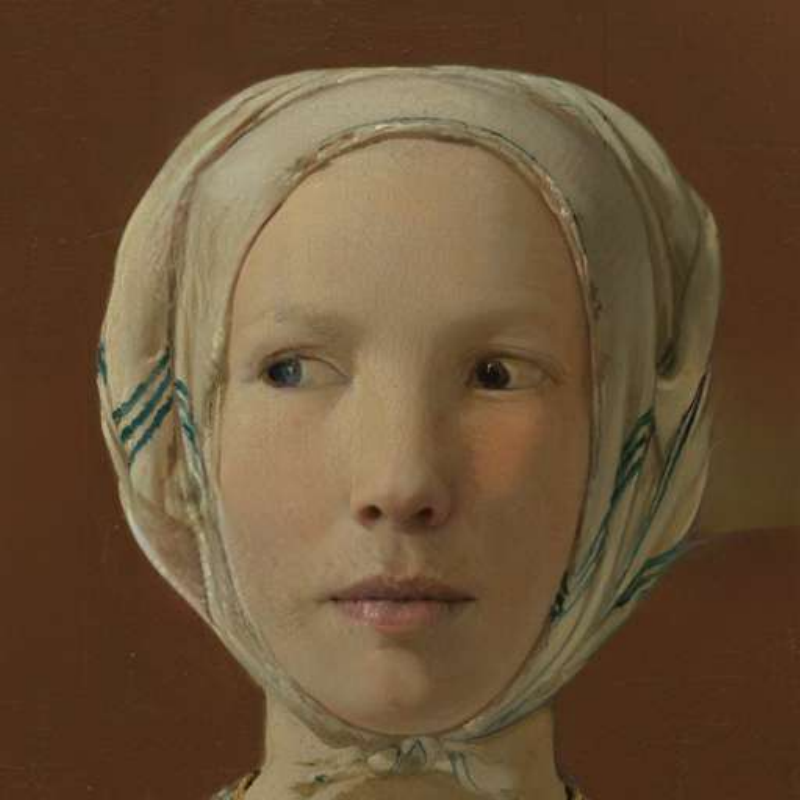}} 
	\\
	
	\frame{\includegraphics[width=\DOMAIN\linewidth]{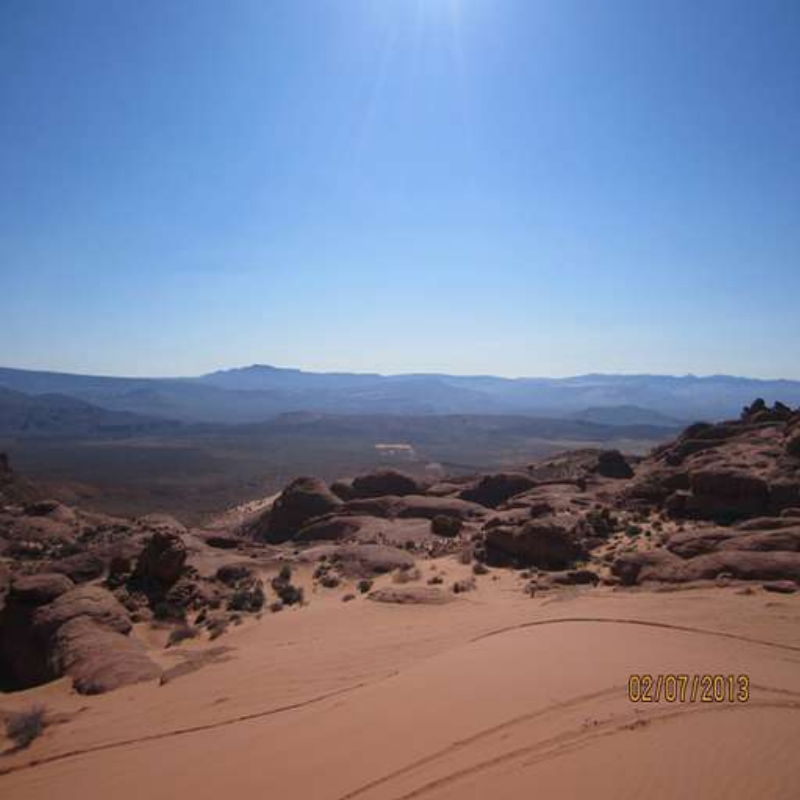}} &
	\frame{\includegraphics[width=\DOMAIN\linewidth]{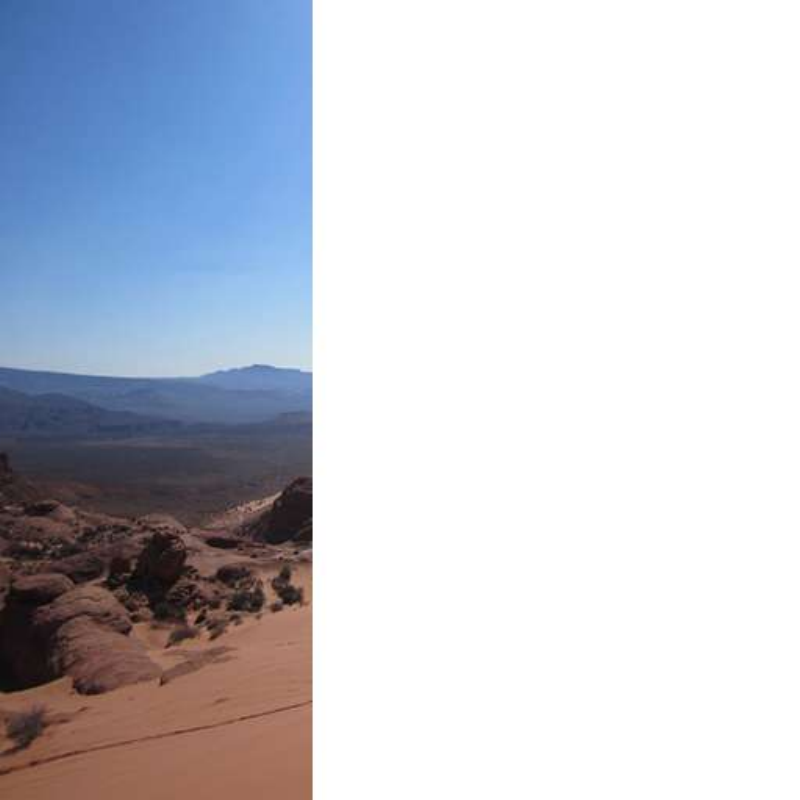}} &
	\frame{\includegraphics[width=\DOMAIN\linewidth]{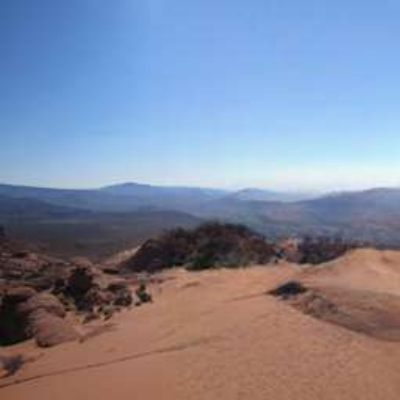}}

\end{tabular}
\caption{Visual results of out-of-domain image inpainting. }
\label{fig:visual_domain}
\end{figure}

\section{Results}\label{sec:results}

\subsection{Comparison to the state-of-the-art}

From Table~\ref{tab:main_compare_1} and Table~\ref{tab:main_compare_2}, our method achieve the best or comparable performance among the state-of-the-art inpainting approaches in terms of FID and LPIPS metrics for all mask settings, indicating that our inpainted results are more realistic. Correspondingly, Fig.~\ref{fig:visual_compare_chq} and ~\ref{fig:visual_compare_places} provide visual inpainting results on CelebA-HQ and Places2 datasets. It can be seen that our method can generate more semantically consistent results compared with other approaches. From the comparison, we can see that our generated images are more semantically consistent while other may produce distortion, blur, and artifacts.

The approaches that do well on pixel-wise similarity are good at employing blurred details, average pixels, and even background-like texture replacement to get higher scores. However, this comes at the expense of fidelity, and our technique performs significantly better at FID than baselines. Further, Fig.~\ref{fig:visual_compare_places} reveals that prior methods struggled to build refined texture if the input image had major corruptions with ambiguity, however our approach was able to create realistic and semantically rich objects such as windows, towers, and mountains.

{\noindent {\bf Fidelity analysis. }} To assess the fidelity of the generated images rather than their similarity to the ground-truth, we further evaluate on Places2 with small ($10\% \sim 30\%$) and large ($70\% \sim 90\%$) mask ratio settings by utilized P-IDS, U-IDS and FID metrics.

We choose Places2 as a benchmark since the P-IDS and U-IDS metrics rely on fitted Support Vector Machine to score the image reality given encoded feature map of images, so we test on the 36,500 images of Places2 to assure distinction. As shown in Table~\ref{tab:main_compare_2}, our method outperforms the state-of-the-arts in all terms, particularly for large mask ratio setting, where our method exceeds the second-best method (except our preliminary ones) by 2037.5\%, 192.2\% and 59.9\% in respect of P-IDS, U-IDS and FID metrics, respectively. The promising results demonstrate that our approach can synthesize high-fidelity images even when faced with large holes.

{\noindent {\bf Gapping issue analysis. }} Because of the "gaping" issue in the image-to-image translation with hard constraints, there is a mismatch in color and a lack of semantic coherence among inversion-based methods. In Fig.~\ref{fig:visual_compare_psp}, mGANprior~\cite{gu2020mganprior} progressively erases the color discrepancy rely on optimized-based inversion but fails to bypass semantic inconsistency. The encoder-based inversion method pSp~\cite{richardson2021encoding} could synthesize realistic pixels for corrupted regions based on the well-trained model, though it still has not resolved the ``gapping'' issue.

At first we expect to address this problem by image post-processing. In particular, we use image blending~\cite{wu2021gpgan}, which does a great job of getting rid of the color discrepancy but it fails to tackle with inconsistent semantic. To further demonstrate the superiority of our method, we construct the `pSp w/ Blending' variant that introduces an image blending~\cite{wu2021gpgan} as image post-processing after generating output images. In Fig.~\ref{fig:visual_compare_psp}, variants with the suffix of `w/o Comp.' stands for the generating images without compositing with ground-truth, and vice versa is the variants with the suffix of `w/ Comp'. 

As depicted in Fig.~\ref{fig:visual_compare_psp}, the color discrepancy occur (zoom in) after compositing, most noticeable at the edges of the mask. The image blending fix the color discrepancy as shown in `pSp w/ Blending' column. Despite this, the post-processing could not help with the semantic inconsistency, as the man's glasses in the second row are still missing some pieces. By merely manipulating the low-level latent code, even with the extended latent space $W^+$, these baselines does not faithfully synthesize the details of the unmasked regions, such as the eye shadow in the first row. And the result of pSp in the second row, it fails to structurally reconstruct glasses.
Hence, previous GAN inversion methods exhibit color and semantic inconsistencies (i.e., visible seams) along the mask boundary after compositing with the ground truth (w/ Comp.).
Instead, the results produced by our method no longer have color discrepancies or semantic inconsistencies of ``gapping'' issue compared to mGANprior, pSp, and pSp with image blending. 

Moreover, we conduct a comparison experiment on CelebA-HQ dataset with the different level of masks. As demonstrated in Fig.~\ref{fig:plots_compare_psp}, our method performs better than previous inversion-based methods mGANprior and pSp w.r.t to FID, LPIPS, and SSIM metrics. That is say, after resolving the "gapping" issue, inversion-based model could get all-round improvement whilst requires no post-processing of images for our method, notably.

\subsection{Out-of-domain Completion}

\begin{table}[t]
\centering
\caption{Comparison with previous inpainting methods on Metfaces. In this experimental setting, the model/generator is only trained on CelebA-HQ.}
\begin{tabular}{lcccc}
\toprule
\multirow{2}{*}{Method} & \multicolumn{2}{c}{Small Mask Ratio} & \multicolumn{2}{c}{Large Mask Ratio} \\ \cmidrule(r){2-3}  \cmidrule(r){4-5} 
                        & FID$\downarrow$    & LPIPS$\downarrow$   & FID$\downarrow$     & LPIPS$\downarrow$   \\ \midrule
RFR~\cite{DBLP:conf/cvpr/LiWZDT20}  & 18.89   & 0.069    & 58.24  & 0.315   \\
CRFill~\cite{zeng2021cr}  & 13.67   & 0.042   & 50.93   &  0.278       \\
LAMA~\cite{DBLP:conf/wacv/SuvorovLMRASKGP22}  & 9.67   & \textbf{0.034}   & 42.93   &  0.255       \\
\midrule
pSp~\cite{richardson2021encoding} & 14.91   & 0.040   & 65.04   & 0.341   \\
Ours  & \textbf{8.84}  & \textbf{0.034}   & \textbf{30.25}   & \textbf{0.231}   \\ \bottomrule
\end{tabular}
\label{tab:compare_metfaces}
\end{table}

In order to validate that our method can reuse the pre-trained GAN generator as priors to tackle images from unseen domains, we conducted two extended tests in which we introduced images or masks from unseen domains and only needed to optimize the lightweight encoder. These tests showed that our method was successful in robustness of out-of-domain completion. The first test is known as image outpainting, which applies our model with outpainting masks~\cite{krishnan2019boundless} on the Scenery dataset. The generator did not retrain on the Scenery dataset and instead kept the weights for Places2. Regarding the second one, we employ MetFaces~\cite{DBLP:conf/nips/KarrasAHLLA20} the face photograph from archaic domain. In the same manner,  we keep the pre-trained weights of the GAN generator that was trained on the CelebA-HQ dataset.

The effects of inpainting an archaic photograph are seen in the top two rows of Fig.~\ref{fig:visual_domain}. It illustrates that our method allows the generator to synthesize semantically coherent style and patches, even in an unknown domain. The outpainting findings on the Scenery dataset are shown in the third and fourth rows of Fig.~\ref{fig:visual_domain}. These results indicate that based merely on the limited corners, our technique can still infer realistic texture and meaningful objects. We only utilize the outpainting masks to train the encoder, and not the GAN generator itself, so that we can guarantee that the masks will not be viewed by the GAN generator.

\begin{table}[!t]
    \centering
    \caption{Comparison with previous outpainting approaches and inpainting baselines on Scenery dataset.}
    \setlength{\tabcolsep}{4pt}
	\begin{tabular}{lclll}
		\toprule
		& where & FID$\downarrow$ & LPIPS$\downarrow$ & SSIM$\uparrow$ \\ \midrule
		LAMA~\cite{DBLP:conf/wacv/SuvorovLMRASKGP22}  & WACV'22    & 138.31  &  0.455 &  0.376  \\
		pSp~\cite{richardson2021encoding}  & CVPR'21 &  49.62  & 0.379 &    0.392   \\ 
		Boundless~\cite{krishnan2019boundless} & ICCV'19   & 45.05  &  0.368 &  0.413 \\
		NS-outpaint~\cite{yang2019very}  &  ICCV'19 & 38.95  &  0.342  &  0.410   \\ 
		Ours        & - & \textbf{16.33}   & \textbf{0.288}  &   \textbf{0.445}  \\ \bottomrule
	\end{tabular}
    \label{tab:compare_scenery}
\end{table}

\newcommand{\SML}{0.3}
\renewcommand\arraystretch{0.8}
\begin{figure}[!t]
	\centering
	\begin{tabular}{c@{\hspace{2.0mm}}c@{\hspace{2.0mm}}c}
	\small{Input} & \small{Ours w/o SML} & \small{Ours w/ SML} \\

	\frame{\includegraphics[width=\SML\linewidth]{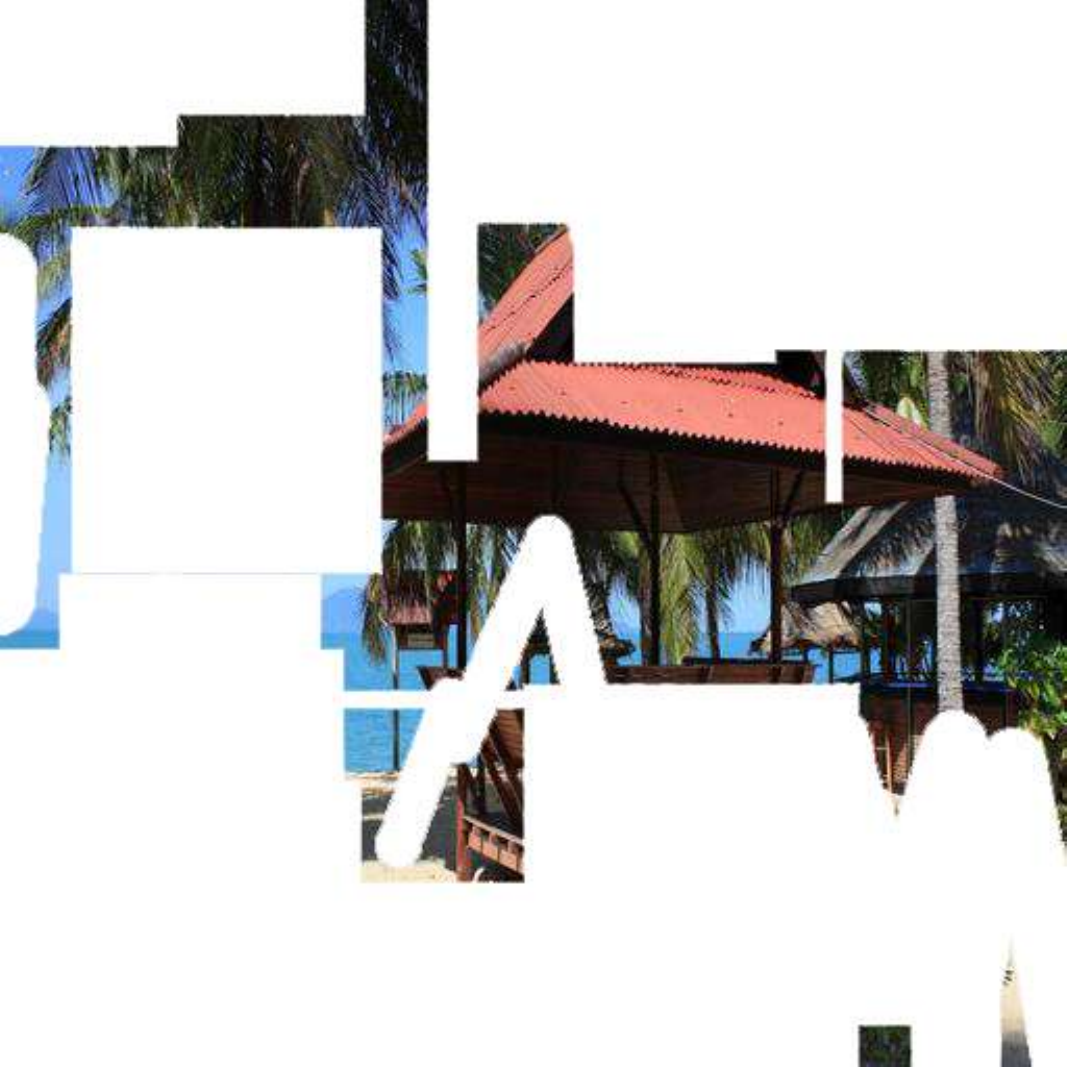}} &
	\frame{\includegraphics[width=\SML\linewidth]{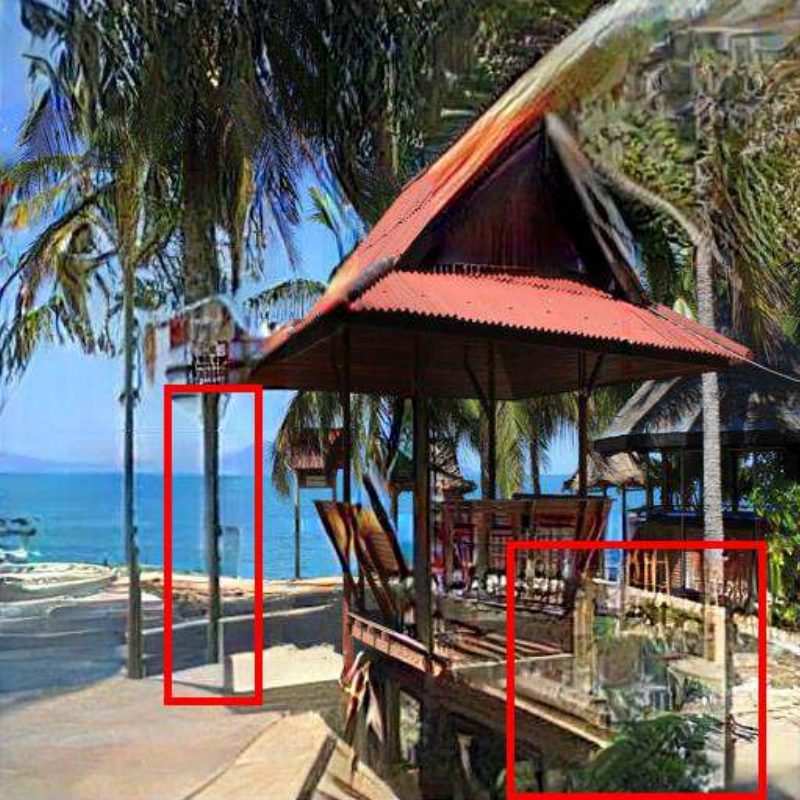}} &
	\frame{\includegraphics[width=\SML\linewidth]{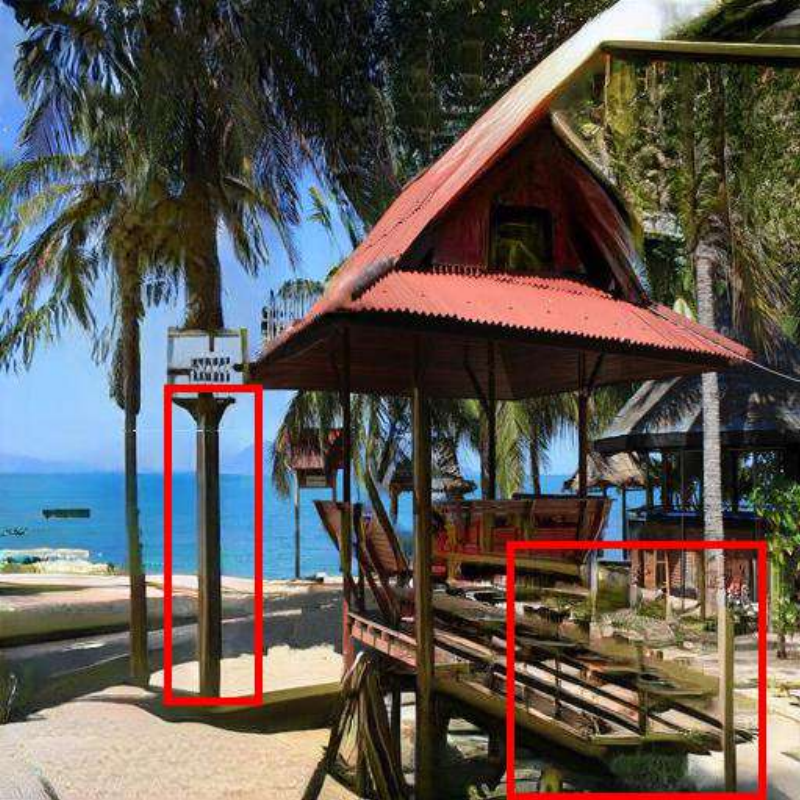}}  \\
	
	\frame{\includegraphics[width=\SML\linewidth]{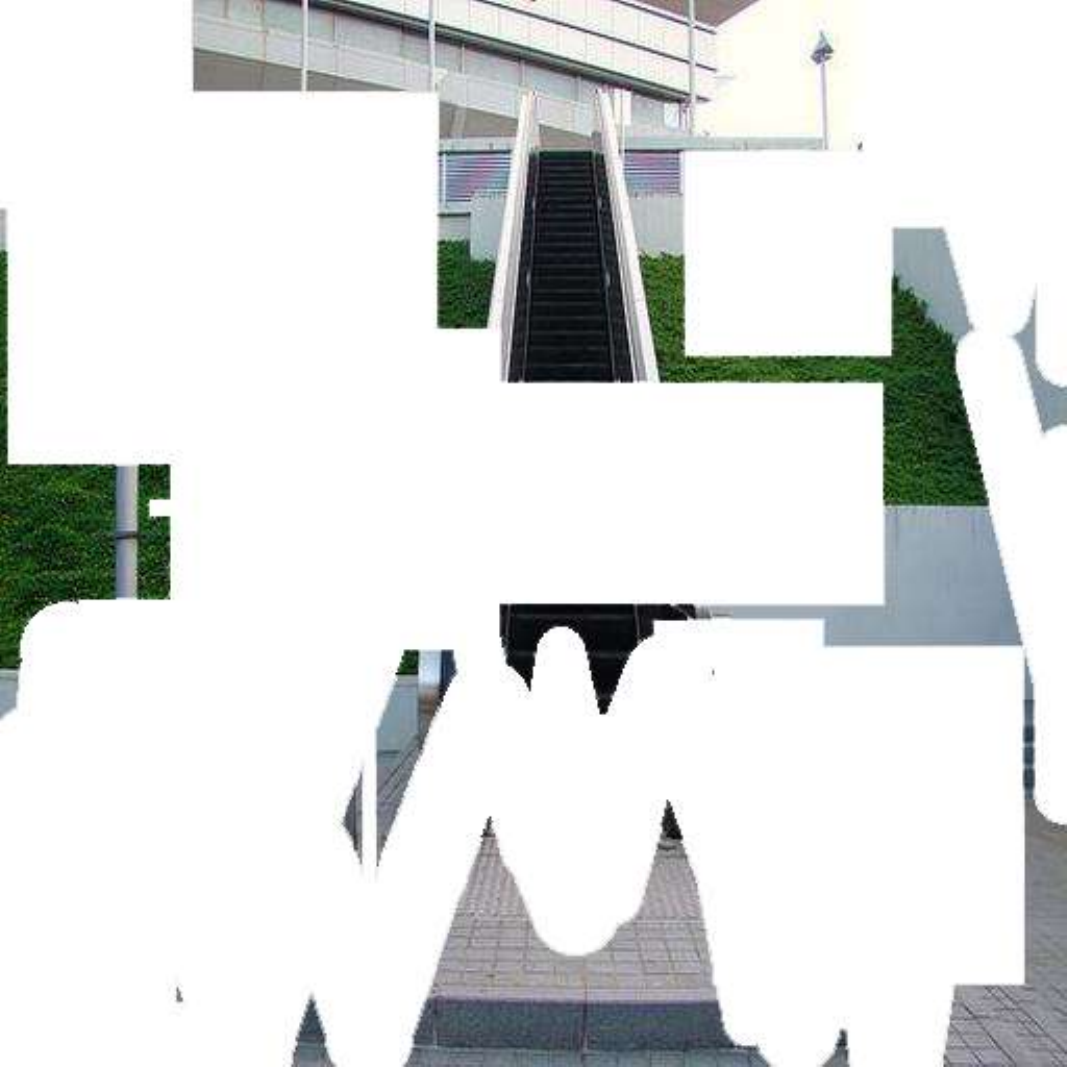}} &
	\frame{\includegraphics[width=\SML\linewidth]{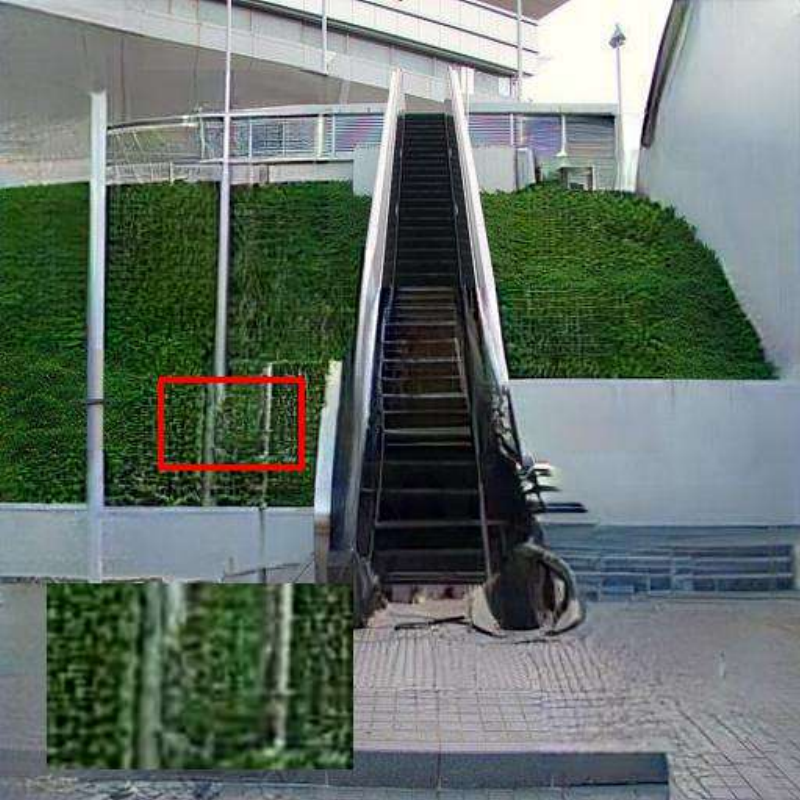}} &
	\frame{\includegraphics[width=\SML\linewidth]{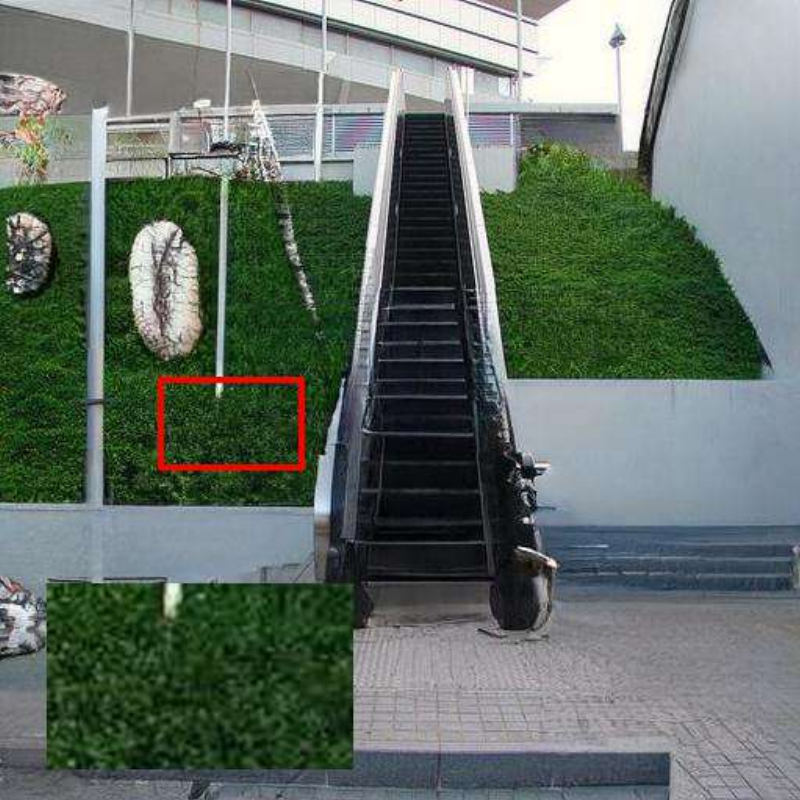}}  \\

\end{tabular}
\caption{The importance of soft-update mean latent. Please zoom in.}
\label{fig:ablation_sml}
\end{figure}

\newcommand{\PLOTA}{0.24}
\renewcommand\arraystretch{0.8}
\begin{figure*}[!th]
	\centering
	\begin{tabular}{c@{\hspace{1.0mm}}c@{\hspace{1.0mm}}c@{\hspace{1.mm}}c}

	\includegraphics[width=\PLOTA\linewidth]{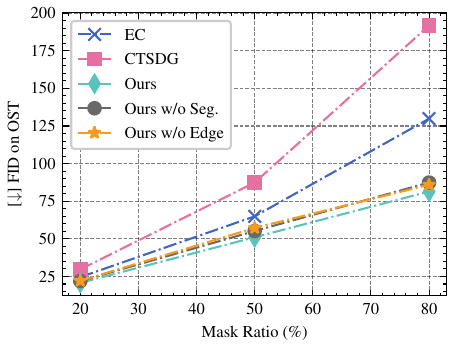} &
	\includegraphics[width=\PLOTA\linewidth]{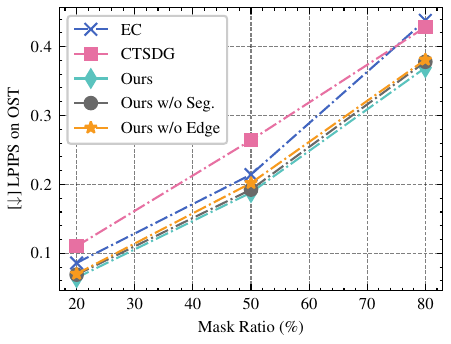} &
	\includegraphics[width=\PLOTA\linewidth]{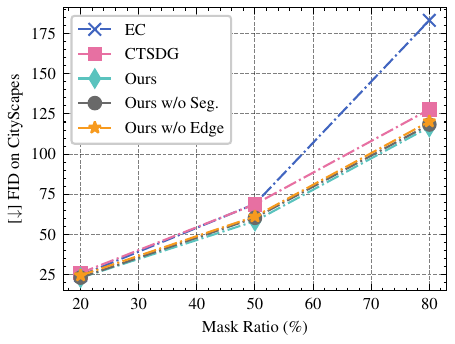} &
	\includegraphics[width=\PLOTA\linewidth]{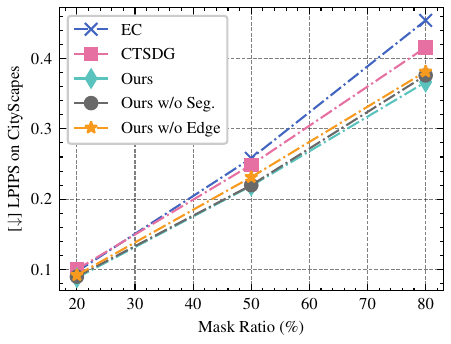}   \\
	\small{(a)} & \small{(b)} & \small{(c)} & \small{(d)} \\

\end{tabular}
\caption{Ablation study to unbiased multi-modal guidance on OST and Cityscapes datasets. }
\label{fig:ablation_struct}
\end{figure*}

In addition, we did a quantitative comparison with the inpainting baselines on the Metfaces, which can be shown in Table~\ref{tab:compare_metfaces}. In conclusion, the findings suggest that the suggested method is reliable and may be extended to other tasks that include inputs from out-of-domain. In a similar vein, we adopt RFR~\cite{DBLP:conf/cvpr/LiWZDT20} and pSp~\cite{richardson2021encoding} as additional baselines, perform tests to assess conventional outpainting techniques, and conduct comparisons between them. According to the results shown in Table~\ref{tab:compare_scenery}, our model achieves much better results than the advanced outpainting baselines~\cite{krishnan2019boundless,yang2019very} when considering FID, LPIPS, and SSIM.

\subsection{Ablation Study}

The ablation experiments are carried out on the Places2 dataset under the large mask ratio setting. 
In Table~\ref{tab:ablation}, we construct four variants to verify the contribution of proposed modules, in which GMA, PM and SML denote gated mask-aware attention, pre-modulation and soft-update mean latent, respectively. 
By learning from these modules, our method considerably outperforms the most naive variants. Additionally, by studying the differences between variants $\mathbb{A}$ and $\mathbb{E}$, it can also be demonstrated that when our proposed $\mathcal{F}\&\mathcal{W}+$ module is adopted, the model's performance is significantly improved quantitatively due to the resolution of the gapping issue.

\begin{table}[t]
    \centering
	\caption{Ablation study comparison on Places2 dataset.}
	\setlength{\tabcolsep}{14pt}
	\begin{tabular}{llcc}
		\toprule
		& & FID$\downarrow$ & LPIPS$\downarrow$ \\ \midrule
		$\mathbb{A }$&  Ours & 10.9116  &  0.3531   \\
		$\mathbb{B }$& - GMA   & 13.1843   & 0.3560 \\ 
		$\mathbb{C }$& - PM    & 15.0592  &  0.3572  \\
		$\mathbb{D }$& - SML  &  21.3145  &  0.3610 \\ 
		$\mathbb{E }$& - $\mathcal{F} \& \mathcal{W}+$   &  31.4058  &  0.4258   \\ \bottomrule
	\end{tabular}
	\label{tab:ablation}
\end{table}

{\noindent {\bf Soft-update mean latent analysis. }}
The soft-update mean latent is motivated by the intuition that fitting diverse domains works better than fitting a preset static domain, especially when the training dataset contains various scenarios such as street and landscape. As shown in Fig.~\ref{fig:ablation_sml}, when we use SML code that dynamically fluctuates during training, the masked region far away from the mask border tends to be reconstructed by explicitly learned semantics instead of repetitive patterns. Notably, `w/o SML' represents using regular static mean latent code.

{\noindent {\bf Multi-modal guidance analysis. }}
We use OST and CityScapes for training and testing in edge and segmentation ablation experiments because they provide ground truth edge and segmentation annotations. In contrast, large-scale dataset Places2 is intended for image classification and lacks such low-level annotations, making it unsuitable for use as an auxiliary structure guidance ablation dataset. In Fig.~\ref{fig:ablation_struct}, we compare several auxiliary prior guided inpainting approaches~\cite{DBLP:conf/iccvw/NazeriNJQE19,guo2021image}. For a fair comparison with the methods relying on only one auxiliary structure, we construct two variants, denoted by ``Ours w/o seg.'' and ``Ours w/o edge''. In particular, on massive corruptions, our technique outperforms FID and LPIPS by a wide margin. This is due to the fact that, unlike competing methods, ours does not directly guide the picture inpainting branch by expected auxiliary structures, but rather emphasizes the interplay representation from three modalities.

In Table~\ref{tab:abla_struct}, we construct several variants to verify the contribution of two auxiliary branches in our method. `w/ b.' and `w/ unb.' stand for our network structure under biased or unbiased guidance, respectively. Specially, `w/ gt' indicates that we utilize the ground truth of auxiliary prior for demonstrating our method could jointly learn
the interplay information of multi-modal features across the three branches and
guide inpainting based on ADN. Notably, we employ `w/ unb.' in the default experimental setting.
Regarding SSIM and FID, our technique performs noticeably better than the non-auxiliary alternative by learning from two auxiliary modalities. Additionally, edge textures contribute slightly less to inpainting than semantic segmentation. In conclusion, our Multi-Modal Mutual Decoder adds semantics to the inpainting branch by cross-attending segmentation and edge structures.

{\noindent {\bf Biased prior guidance. }}
In Table~\ref{tab:abla_att}, MMT-1 represents concatenating predicted structures with feature maps of inpainting branch. MMT-2, MMT-3, and MMT-4 indicate that we employ AdaIN~\cite{DBLP:conf/iccv/HuangB17}, SPADE~\cite{DBLP:conf/cvpr/Park0WZ19}, and MSSA with ADN, respectively, to determine the affine transformation parameters $(\gamma, \beta)$ based on biased structures. Based on biased auxiliary structures, the FID score is much lower as compared to our method without biased prior guidance (\ie, MMT-9). The findings corroborate our hypothesis that biased structures may bring additional noise in image inpainting without ground-truth. Similarly, we can get the same conclusion by comparing variants of `w/ b.' and `w/ unb.' in the Table~\ref{tab:abla_struct}.

\begin{table}[t]
\centering
\caption{Contribution of two auxiliary branches in our method.}
\label{tab:abla_struct}
\setlength{\tabcolsep}{1pt}
\begin{tabular}{cccccccc}
\toprule
\multicolumn{3}{c}{edge branch} & \multicolumn{3}{c}{segmentation branch} &     &      \\ \cmidrule(r){1-3}\cmidrule(r){4-6}
w/ b. & w/ unb. & w/ gt & w/ b.    & w/ unb.    & w/ gt   & FID$\downarrow$ & SSIM$\uparrow$ \\ \midrule
- & - & - & - & - & - &  14.9313   & 0.8815  \\
- & \Checkmark & - & - & - & -  & 13.7514    & 0.8850     \\
- & - & - & - &  \Checkmark & - & 12.8401    & 0.8895     \\
- & \Checkmark & - & - &  \Checkmark & - & \textbf{10.1012}    & \textbf{0.8956} \\
\Checkmark  & - & - & \Checkmark  & - & - & 12.0157  &  0.8910    \\\midrule
-  & - & \Checkmark & - & - & \Checkmark & 8.1642    & 0.9179 \\
\bottomrule
\end{tabular}
\end{table}

{\noindent {\bf Gated mask-aware attention analysis. }}
We build four feature fusion strategies, ranging from MMT-5 to MMT-8 in Table~\ref{tab:abla_att}, to test the efficacy of GMA. To the MMT-5, we directly do element-wise summing on features from three branches before GMA. As with MMT-5, MMT-6 combines data from three different sources using splicing and then fusing it together using two convolutional layers. Our GMA fares best for the three parameters in Table~\ref{tab:abla_att}. Instead of using simple addition or convolution, our GMA perform reliably cross-attention between different modalities to direct high-quality image reconstruction. For MMT-7 and MMT-8, we use AdaIN~\cite{DBLP:conf/iccv/HuangB17} and SPADE~\cite{DBLP:conf/cvpr/Park0WZ19} in place of ADN in GMA. Our ADN outperforms state-of-the-art normalization approaches, proving its worth.

\begin{table}[t]
    \footnotesize
    \caption{Comparison with different feature fusion strategies for three branches.}
    \label{tab:abla_att}
    \centering
    \setlength{\tabcolsep}{3.5pt}
    \begin{tabular}{ccc|cc}
    \toprule
    variant &biased prior & fusion strategy    & FID$\downarrow$ & SSIM$\uparrow$ \\\midrule
    MMT-1 &\Checkmark &concat      &  22.80    & 0.86    \\
    MMT-2 &\Checkmark &AdaIN~\cite{DBLP:conf/iccv/HuangB17}     &  26.71   & 0.85     \\
    MMT-3 &\Checkmark &SPADE~\cite{DBLP:conf/cvpr/Park0WZ19}     &  20.78   & 0.86   \\
    MMT-4 &\Checkmark &GMA+ADN &  18.31    & 0.87      \\
    \midrule
    MMT-5 &\XSolidBrush &GMA+add  &  17.84    & 0.87      \\
    MMT-6 &\XSolidBrush &GMA+conv &  21.05    & 0.86     \\
    MMT-7 &\XSolidBrush &GMA+AdaIN~\cite{DBLP:conf/iccv/HuangB17} &  18.59    & 0.87  \\
    MMT-8 &\XSolidBrush &GMA+SPADE~\cite{DBLP:conf/cvpr/Park0WZ19} &  17.13    & 0.87  \\
    MMT-9 &\XSolidBrush &GMA+ADN &  \bf{15.74}    &   \bf{0.88}   \\
    \bottomrule
    \end{tabular}
\end{table}

{\noindent {\bf Adaptive contextual bottlenecks analysis. }}
Table~\ref{tab:abla_acb} shows how our ACB stacks up against the standard ResNet blocks~\cite{DBLP:conf/cvpr/HeZRS16} and the AOTs~\cite{DBLP:journals/corr/abs-2104-01431}. RES@$8$ and AOT@$8$ refer to $8$ ResNet blocks~\cite{DBLP:conf/cvpr/HeZRS16} and $8$ AOT blocks~\cite{DBLP:journals/corr/abs-2104-01431}, respectively, whereas ACB@$L$ ($L = 2, 4, 6, 8$) indicates $L$ levels of ACB modules. When applied to ResNet blocks, $\dag$ implies duplicating the base feature maps for different paths, and when applied to AOT blocks, $\dag$ means quadrupling the channels of feature maps. It turns out that when you increase the number of blocks in ACB, you enhance its performance. Results are comparable when $8$ ResNet or AOT blocks are used instead of $4$ ACB ones. It's worth noting that each ResNet and AOT block uses less feature map channels. Our ACB blocks are same in number of channels, so we build two variants, $\dag$RES@$8$ and $\dag$AOT@$8$, for a level playing field. 
To the contrary, employing ResNet or AOT blocks does not improve performance when more channels are included in the feature maps. With more channels of feature maps, we hypothesize that our ACB's gating updating scheme can lessen the impact of duplicated noisy context. 
To further verify the impact of bottleneck modules on segmentation inpainting, we can use the mean of category-wise intersection-over-union (mIoU)~\cite{DBLP:conf/eccv/ChenZPSA18}. Still, our ACB module ($L \leq 4$) is superior to the other two kinds of blocks.

\begin{table}[!t]
    \centering
    \caption{Comparison between different bottlenecks.}
    \label{tab:abla_acb}
    \setlength{\tabcolsep}{7pt}
    \begin{tabular}{ccccc}
    \toprule
    bottleneck & FID$\downarrow$ & SSIM$\uparrow$ & LPIPS$\downarrow$  & mIoU\%$\uparrow$\\ \midrule
    RES@8  &  20.71  &  0.87  & 0.16 & 60.95 \\
    $\dag$RES@8  &  21.90    &  0.86    &  0.17 & 60.10  \\
    AOT@8     &   19.05  &  0.87    &  0.13 & 63.88 \\
    $\dag$AOT@8    &   21.22  &  0.87    &  0.14 & 62.83  \\\midrule
    ACB@2     &   19.28   &  0.87    &  0.12 & 63.92 \\
    ACB@4     &   17.34   &  0.87    &  0.11  & 65.03 \\
    ACB@6     &   16.01   &  0.88    &  0.10 & 65.89 \\
    \textbf{ACB@8}     &  \textbf{15.74}    &   \textbf{0.88}   &  \textbf{0.09} & \textbf{66.51} \\
    \bottomrule
    \end{tabular}
\end{table}

\subsection{Discussion}

Fig.~\ref{fig:fail} shows two failure cases.
Since most of the images in the CelebA-HQ dataset are frontal views of celebrities' faces, the left-hand example represents a use case (side face) that is statistically near the long-tail sample point. Even while our approach recovers details to the eye, a global distortion still remains. The right-hand example from Places2, where the model has trouble reconstructing a masked face since the dataset consists mainly of mountainous scenery, buildings, and streets, struggles to do so in the image. Instead of directly operating the input picture, the GAN model is constrained by the pre-trained latent space since it manipulates the latent code of the conditional image. Therefore, it is still a challenge to handle fine-grained from extensive domains or samples with a long tail.

\newcommand{\FAIL}{0.23}
\renewcommand\arraystretch{0.8}
\begin{figure}[t]
	\centering
	\begin{tabular}{c@{\hspace{1.0mm}}c@{\hspace{1.0mm}}c@{\hspace{1.0mm}}c}
	\small{Masked} & \small{Ours} & \small{Masked} & \small{Ours} \\

	\frame{\includegraphics[width=\FAIL\linewidth]{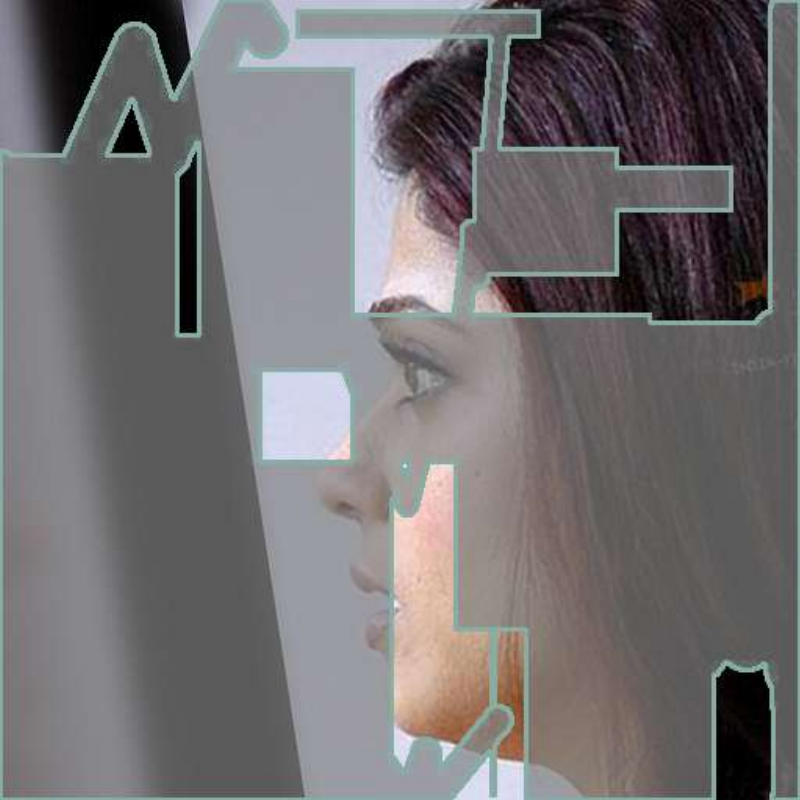}} &
	\frame{\includegraphics[width=\FAIL\linewidth]{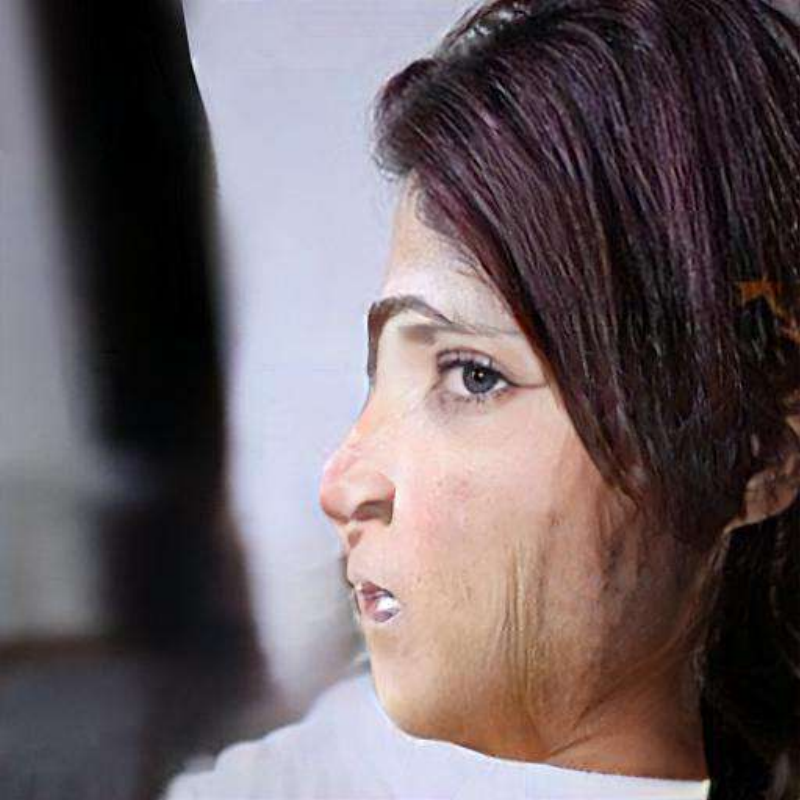}} &
	\frame{\includegraphics[width=\FAIL\linewidth]{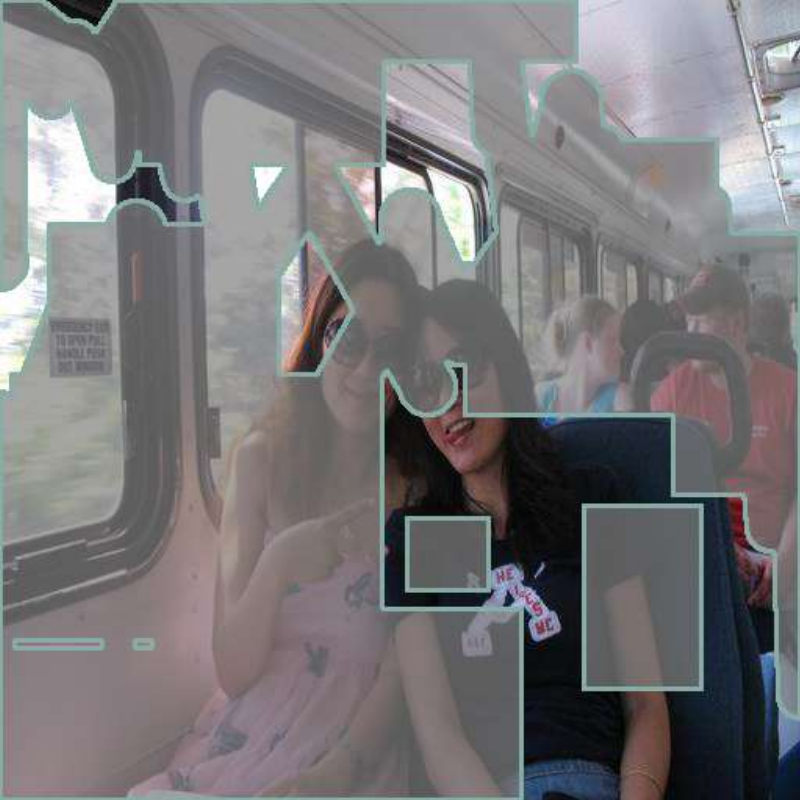}} &
	\frame{\includegraphics[width=\FAIL\linewidth]{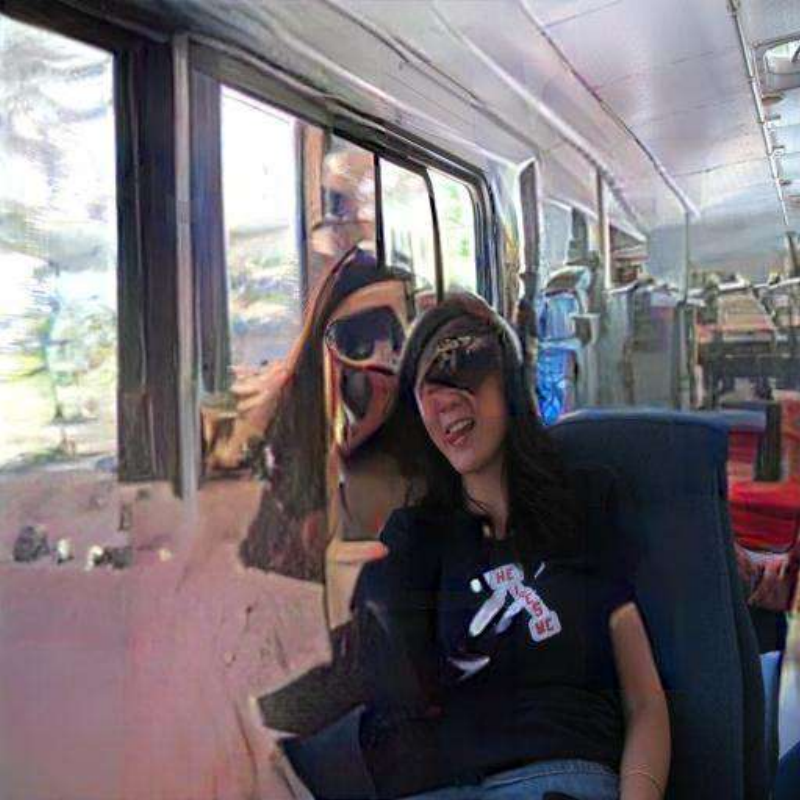}}\\

\end{tabular}
\caption{Illustration of two failure cases of the proposed method.}
\label{fig:fail}
\end{figure}

\section{Conclusion}\label{sec:conclusion}
In this paper, we propose an encoder-based GAN inversion method for image inpainting. The encoder is an end-to-end Multi-modality Guided Transformer which enriches coupled spatial features from shared multi-modal representations (\ie, RGB image, semantic segmentation and edge textures), projects corrupted images into a latent space $\mathcal{F}\&\mathcal{W}^+$ with pre-modulation for learning more discriminative representation. Within the encoder, the suggested Multi-Scale Spatial-aware Attention can integrate compact discriminative features from multiple modalities via Auxiliary DeNormalization. Meanwhile, we introduce the Adaptive Contextual Bottlenecks in the encoder to enhance context reasoning for more semantically consistent inpainted results for the missing region. The novel latent space $\mathcal{F}\&\mathcal{W}^+$ resolves the ``gapping'' issue when applied to GAN inversion in image inpainting. In addition, the soft-update mean latent dynamically samples diverse in-domain patterns, leading to more realistic textures. Extensive quantitative and qualitative comparisons demonstrate the superiority of our model over previous approaches and can cheaply support the semantically consistent completion of images or masks from unseen domains.

\section{Data Availability Statement}
The data supporting the findings of this study are openly available, and our codes will be released at 
\href{https://yeates.github.io/mm-invertfill}{https://yeates.github.io/mm-invertfill}.\\

\noindent \textbf{Acknowledgement.} Libo Zhang is supported by National Natural Science Foundation of China (No. 62476266). Heng Fan has not been supported by any funding for this work at any stage.

\vspace{1em}
\noindent

\bibliographystyle{sn-basic}
\bibliography{sn-bibliography}

\begin{thebibliography}{78}
\providecommand{\natexlab}[1]{#1}
\providecommand{\url}[1]{{#1}}
\providecommand{\urlprefix}{URL }
\providecommand{\doi}[1]{\url{https://doi.org/#1}}
\providecommand{\eprint}[2][]{\url{#2}}
 \bibcommenthead

\bibitem[{Abdal et~al(2019)Abdal, Qin, and Wonka}]{abdal2019image2stylegan}
Abdal R, Qin Y, Wonka P (2019) Image2stylegan: How to embed images into the stylegan latent space? In: ICCV

\bibitem[{Abdal et~al(2020)Abdal, Qin, and Wonka}]{abdal2020image2stylegan}
Abdal R, Qin Y, Wonka P (2020) Image2stylegan++: How to edit the embedded images? In: CVPR

\bibitem[{Ardino et~al(2020)Ardino, Liu, Ricci, Lepri, and Nadai}]{DBLP:conf/icpr/ArdinoL0LN20}
Ardino P, Liu Y, Ricci E, et~al (2020) Semantic-guided inpainting network for complex urban scenes manipulation. In: ICPR

\bibitem[{Ba et~al(2016)Ba, Kiros, and Hinton}]{DBLP:journals/corr/BaKH16}
Ba LJ, Kiros JR, Hinton GE (2016) Layer normalization. arXiv

\bibitem[{Cao and Fu(2021)}]{DBLP:journals/corr/abs-2103-15087}
Cao C, Fu Y (2021) Learning a sketch tensor space for image inpainting of man-made scenes. In: ICCV

\bibitem[{Chen et~al(2018)Chen, Zhu, Papandreou, Schroff, and Adam}]{DBLP:conf/eccv/ChenZPSA18}
Chen L, Zhu Y, Papandreou G, et~al (2018) Encoder-decoder with atrous separable convolution for semantic image segmentation. In: ECCV

\bibitem[{Cheng et~al(2022)Cheng, Lin, Lee, Ren, Tulyakov, and Yang}]{cheng2022inout}
Cheng YC, Lin CH, Lee HY, et~al (2022) Inout: Diverse image outpainting via gan inversion. In: CVPR

\bibitem[{Cordts et~al(2016)Cordts, Omran, Ramos, Rehfeld, Enzweiler, Benenson, Franke, Roth, and Schiele}]{DBLP:conf/cvpr/CordtsORREBFRS16}
Cordts M, Omran M, Ramos S, et~al (2016) The cityscapes dataset for semantic urban scene understanding. In: CVPR

\bibitem[{Criminisi et~al(2003)Criminisi, P{\'{e}}rez, and Toyama}]{DBLP:conf/cvpr/CriminisiPT03}
Criminisi A, P{\'{e}}rez P, Toyama K (2003) Object removal by exemplar-based inpainting. In: CVPR

\bibitem[{Deng et~al(2021)Deng, Hui, Zhou, Meng, and Wang}]{DBLP:conf/mm/DengHZMW21}
Deng Y, Hui S, Zhou S, et~al (2021) Learning contextual transformer network for image inpainting. In: {MM}, pp 2529--2538

\bibitem[{Dong et~al(2022)Dong, Cao, and Fu}]{DongZITS}
Dong Q, Cao C, Fu Y (2022) Incremental transformer structure enhanced image inpainting with masking positional encoding. In: CVPR

\bibitem[{Dosovitskiy et~al(2021)Dosovitskiy, Beyer, Kolesnikov, Weissenborn, Zhai, Unterthiner, Dehghani, Minderer, Heigold, Gelly, Uszkoreit, and Houlsby}]{DBLP:conf/iclr/DosovitskiyB0WZ21}
Dosovitskiy A, Beyer L, Kolesnikov A, et~al (2021) An image is worth 16x16 words: Transformers for image recognition at scale. In: ICLR

\bibitem[{Goodfellow et~al(2014)Goodfellow, Pouget{-}Abadie, Mirza, Xu, Warde{-}Farley, Ozair, Courville, and Bengio}]{DBLP:journals/corr/GoodfellowPMXWOCB14}
Goodfellow IJ, Pouget{-}Abadie J, Mirza M, et~al (2014) Generative adversarial nets. In: NIPS

\bibitem[{Gu et~al(2020)Gu, Shen, and Zhou}]{gu2020mganprior}
Gu J, Shen Y, Zhou B (2020) Image processing using multi-code {GAN} prior. In: CVPR

\bibitem[{Gulrajani et~al(2017)Gulrajani, Ahmed, Arjovsky, Dumoulin, and Courville}]{gulrajani2017improved}
Gulrajani I, Ahmed F, Arjovsky M, et~al (2017) Improved training of wasserstein gans. In: NIPS

\bibitem[{Guo et~al(2021)Guo, Yang, and Huang}]{guo2021image}
Guo X, Yang H, Huang D (2021) Image inpainting via conditional texture and structure dual generation. In: ICCV

\bibitem[{He et~al(2016)He, Zhang, Ren, and Sun}]{DBLP:conf/cvpr/HeZRS16}
He K, Zhang X, Ren S, et~al (2016) Deep residual learning for image recognition. In: CVPR

\bibitem[{Heusel et~al(2017)Heusel, Ramsauer, Unterthiner, Nessler, and Hochreiter}]{DBLP:conf/nips/HeuselRUNH17}
Heusel M, Ramsauer H, Unterthiner T, et~al (2017) Gans trained by a two time-scale update rule converge to a local nash equilibrium. In: NIPS

\bibitem[{Huang and Belongie(2017)}]{DBLP:conf/iccv/HuangB17}
Huang X, Belongie SJ (2017) Arbitrary style transfer in real-time with adaptive instance normalization. In: ICCV

\bibitem[{Iizuka et~al(2017)Iizuka, Simo{-}Serra, and Ishikawa}]{DBLP:journals/tog/IizukaS017}
Iizuka S, Simo{-}Serra E, Ishikawa H (2017) Globally and locally consistent image completion. TOG 36(4):1--14

\bibitem[{Karras et~al(2019)Karras, Laine, and Aila}]{karras2019style}
Karras T, Laine S, Aila T (2019) A style-based generator architecture for generative adversarial networks. In: CVPR

\bibitem[{Karras et~al(2020{\natexlab{a}})Karras, Aittala, Hellsten, Laine, Lehtinen, and Aila}]{DBLP:conf/nips/KarrasAHLLA20}
Karras T, Aittala M, Hellsten J, et~al (2020{\natexlab{a}}) Training generative adversarial networks with limited data. In: NeurIPS

\bibitem[{Karras et~al(2020{\natexlab{b}})Karras, Laine, Aittala, Hellsten, Lehtinen, and Aila}]{karras2020analyzing}
Karras T, Laine S, Aittala M, et~al (2020{\natexlab{b}}) Analyzing and improving the image quality of stylegan. In: CVPR

\bibitem[{Krishnan et~al(2019)Krishnan, Teterwak, Sarna, Maschinot, Liu, Belanger, and Freeman}]{krishnan2019boundless}
Krishnan D, Teterwak P, Sarna A, et~al (2019) Boundless: Generative adversarial networks for image extension. In: ICCV

\bibitem[{Lee et~al(2020)Lee, Liu, Wu, and Luo}]{DBLP:conf/cvpr/Lee0W020}
Lee C, Liu Z, Wu L, et~al (2020) Maskgan: Towards diverse and interactive facial image manipulation. In: CVPR

\bibitem[{Li et~al(2020)Li, Wang, Zhang, Du, and Tao}]{DBLP:conf/cvpr/LiWZDT20}
Li J, Wang N, Zhang L, et~al (2020) Recurrent feature reasoning for image inpainting. In: CVPR

\bibitem[{Li et~al(2022{\natexlab{a}})Li, Lin, Zhou, Qi, Wang, and Jia}]{li2022mat}
Li W, Lin Z, Zhou K, et~al (2022{\natexlab{a}}) Mat: Mask-aware transformer for large hole image inpainting. In: CVPR

\bibitem[{Li et~al(2022{\natexlab{b}})Li, Guo, Lin, Li, Feng, and Wang}]{li2022misf}
Li X, Guo Q, Lin D, et~al (2022{\natexlab{b}}) Misf: Multi-level interactive siamese filtering for high-fidelity image inpainting. In: CVPR

\bibitem[{Liao et~al(2020)Liao, Xiao, Wang, Lin, and Satoh}]{DBLP:conf/eccv/LiaoXWLS20}
Liao L, Xiao J, Wang Z, et~al (2020) Guidance and evaluation: Semantic-aware image inpainting for mixed scenes. In: ECCV

\bibitem[{Liao et~al(2021)Liao, Xiao, Wang, Lin, and Satoh}]{DBLP:conf/cvpr/Liao00L021}
Liao L, Xiao J, Wang Z, et~al (2021) Image inpainting guided by coherence priors of semantics and textures. In: CVPR

\bibitem[{Lillicrap et~al(2016)Lillicrap, Hunt, Pritzel, Heess, Erez, Tassa, Silver, and Wierstra}]{DBLP:journals/corr/LillicrapHPHETS15}
Lillicrap TP, Hunt JJ, Pritzel A, et~al (2016) Continuous control with deep reinforcement learning. In: ICLR

\bibitem[{Liu et~al(2018)Liu, Reda, Shih, Wang, Tao, and Catanzaro}]{DBLP:conf/eccv/LiuRSWTC18}
Liu G, Reda FA, Shih KJ, et~al (2018) Image inpainting for irregular holes using partial convolutions. In: ECCV

\bibitem[{Liu et~al(2019)Liu, Jiang, Xiao, and Yang}]{DBLP:conf/iccv/LiuJX019}
Liu H, Jiang B, Xiao Y, et~al (2019) Coherent semantic attention for image inpainting. In: ICCV

\bibitem[{Liu et~al(2020)Liu, Jiang, Song, Huang, and Yang}]{liu2020rethinking}
Liu H, Jiang B, Song Y, et~al (2020) Rethinking image inpainting via a mutual encoder-decoder with feature equalizations. In: ECCV

\bibitem[{Mao et~al(2017)Mao, Li, Xie, Lau, Wang, and Smolley}]{mao2017least}
Mao X, Li Q, Xie H, et~al (2017) Least squares generative adversarial networks. In: ICCV

\bibitem[{Miyato et~al(2018)Miyato, Kataoka, Koyama, and Yoshida}]{DBLP:conf/iclr/MiyatoKKY18}
Miyato T, Kataoka T, Koyama M, et~al (2018) Spectral normalization for generative adversarial networks. In: ICLR

\bibitem[{Nazeri et~al(2019)Nazeri, Ng, Joseph, Qureshi, and Ebrahimi}]{DBLP:conf/iccvw/NazeriNJQE19}
Nazeri K, Ng E, Joseph T, et~al (2019) Edgeconnect: Structure guided image inpainting using edge prediction. In: ICCVW

\bibitem[{Park et~al(2019)Park, Liu, Wang, and Zhu}]{DBLP:conf/cvpr/Park0WZ19}
Park T, Liu M, Wang T, et~al (2019) Semantic image synthesis with spatially-adaptive normalization. In: CVPR

\bibitem[{Pathak et~al(2016)Pathak, Kr{\"{a}}henb{\"{u}}hl, Donahue, Darrell, and Efros}]{DBLP:conf/cvpr/PathakKDDE16}
Pathak D, Kr{\"{a}}henb{\"{u}}hl P, Donahue J, et~al (2016) Context encoders: Feature learning by inpainting. In: CVPR

\bibitem[{Richardson et~al(2021)Richardson, Alaluf, Patashnik, Nitzan, Azar, Shapiro, and Cohen{-}Or}]{richardson2021encoding}
Richardson E, Alaluf Y, Patashnik O, et~al (2021) Encoding in style: {A} stylegan encoder for image-to-image translation. In: CVPR

\bibitem[{Ronneberger et~al(2015)Ronneberger, Fischer, and Brox}]{DBLP:conf/miccai/RonnebergerFB15}
Ronneberger O, Fischer P, Brox T (2015) U-net: Convolutional networks for biomedical image segmentation. In: MICCAI

\bibitem[{Salimans et~al(2016)Salimans, Goodfellow, Zaremba, Cheung, Radford, and Chen}]{DBLP:conf/nips/SalimansGZCRCC16}
Salimans T, Goodfellow IJ, Zaremba W, et~al (2016) Improved techniques for training gans. In: NIPS

\bibitem[{Shetty et~al(2018)Shetty, Fritz, and Schiele}]{DBLP:conf/nips/ShettyFS18}
Shetty R, Fritz M, Schiele B (2018) Adversarial scene editing: Automatic object removal from weak supervision. In: NIPS

\bibitem[{Song et~al(2019)Song, Cao, Song, Hu, and He}]{DBLP:conf/aaai/SongCSHH19}
Song L, Cao J, Song L, et~al (2019) Geometry-aware face completion and editing. In: AAAI

\bibitem[{Song et~al(2018)Song, Yang, Shen, Wang, Huang, and Kuo}]{DBLP:conf/bmvc/SongYSWHK18}
Song Y, Yang C, Shen Y, et~al (2018) Spg-net: Segmentation prediction and guidance network for image inpainting. In: BMVC

\bibitem[{Song et~al(2021)Song, Sohl{-}Dickstein, Kingma, Kumar, Ermon, and Poole}]{DBLP:conf/iclr/0011SKKEP21}
Song Y, Sohl{-}Dickstein J, Kingma DP, et~al (2021) Score-based generative modeling through stochastic differential equations. In: {ICLR}

\bibitem[{Suvorov et~al(2022)Suvorov, Logacheva, Mashikhin, Remizova, Ashukha, Silvestrov, Kong, Goka, Park, and Lempitsky}]{DBLP:conf/wacv/SuvorovLMRASKGP22}
Suvorov R, Logacheva E, Mashikhin A, et~al (2022) Resolution-robust large mask inpainting with fourier convolutions. In: WACV

\bibitem[{Ulyanov et~al(2016)Ulyanov, Vedaldi, and Lempitsky}]{ulyanov2016instance}
Ulyanov D, Vedaldi A, Lempitsky VS (2016) Instance normalization: The missing ingredient for fast stylization. arXiv

\bibitem[{Wan et~al(2021)Wan, Zhang, Chen, and Liao}]{Wan_2021_ICCV}
Wan Z, Zhang J, Chen D, et~al (2021) High-fidelity pluralistic image completion with transformers. In: ICCV

\bibitem[{Wang et~al(2018{\natexlab{a}})Wang, Chen, Yuan, Liu, Huang, Hou, and Cottrell}]{DBLP:conf/wacv/WangCYLHHC18}
Wang P, Chen P, Yuan Y, et~al (2018{\natexlab{a}}) Understanding convolution for semantic segmentation. In: WACV

\bibitem[{Wang et~al(2018{\natexlab{b}})Wang, Yu, Dong, and Loy}]{DBLP:conf/cvpr/WangYDL18}
Wang X, Yu K, Dong C, et~al (2018{\natexlab{b}}) Recovering realistic texture in image super-resolution by deep spatial feature transform. In: CVPR

\bibitem[{Wang et~al(2018{\natexlab{c}})Wang, Tao, Qi, Shen, and Jia}]{DBLP:conf/nips/WangTQSJ18}
Wang Y, Tao X, Qi X, et~al (2018{\natexlab{c}}) Image inpainting via generative multi-column convolutional neural networks. In: NIPS

\bibitem[{Wang et~al(2004)Wang, Bovik, Sheikh, and Simoncelli}]{DBLP:journals/tip/WangBSS04}
Wang Z, Bovik AC, Sheikh HR, et~al (2004) Image quality assessment: from error visibility to structural similarity. IEEE transactions on image processing 13(4):600--612

\bibitem[{Woo et~al(2018)Woo, Park, Lee, and Kweon}]{DBLP:conf/eccv/WooPLK18}
Woo S, Park J, Lee J, et~al (2018) {CBAM:} convolutional block attention module. In: ECCV

\bibitem[{Wu et~al(2019)Wu, Zheng, Zhang, and Huang}]{wu2021gpgan}
Wu H, Zheng S, Zhang J, et~al (2019) {GP-GAN:} towards realistic high-resolution image blending. In: ACM MM

\bibitem[{Xia et~al(2022)Xia, Zhang, Yang, Xue, Zhou, and Yang}]{xia21inversionsurvey}
Xia W, Zhang Y, Yang Y, et~al (2022) Gan inversion: A survey. IEEE Transactions on Pattern Analysis and Machine Intelligence

\bibitem[{Xiong et~al(2019)Xiong, Yu, Lin, Yang, Lu, Barnes, and Luo}]{DBLP:conf/cvpr/XiongYLYLBL19}
Xiong W, Yu J, Lin Z, et~al (2019) Foreground-aware image inpainting. In: CVPR

\bibitem[{Xu et~al(2021)Xu, Shen, Zhu, Yang, and Zhou}]{xu2021generative}
Xu Y, Shen Y, Zhu J, et~al (2021) Generative hierarchical features from synthesizing images. In: CVPR

\bibitem[{Yan et~al(2018)Yan, Li, Li, Zuo, and Shan}]{DBLP:conf/eccv/YanLLZS18}
Yan Z, Li X, Li M, et~al (2018) Shift-net: Image inpainting via deep feature rearrangement. In: ECCV

\bibitem[{Yang et~al(2020)Yang, Qi, and Shi}]{DBLP:conf/aaai/YangQS20}
Yang J, Qi Z, Shi Y (2020) Learning to incorporate structure knowledge for image inpainting. In: AAAI

\bibitem[{Yang et~al(2022)Yang, Jiang, Liu, and Loy}]{yang2022vtoonify}
Yang S, Jiang L, Liu Z, et~al (2022) Vtoonify: Controllable high-resolution portrait video style transfer. TOG

\bibitem[{Yang et~al(2019)Yang, Dong, Liu, Yang, and Yan}]{yang2019very}
Yang Z, Dong J, Liu P, et~al (2019) Very long natural scenery image prediction by outpainting. In: ICCV

\bibitem[{Yu et~al(2017)Yu, Koltun, and Funkhouser}]{DBLP:conf/cvpr/YuKF17}
Yu F, Koltun V, Funkhouser TA (2017) Dilated residual networks. In: CVPR

\bibitem[{Yu et~al(2018)Yu, Lin, Yang, Shen, Lu, and Huang}]{DBLP:conf/cvpr/Yu0YSLH18}
Yu J, Lin Z, Yang J, et~al (2018) Generative image inpainting with contextual attention. In: CVPR

\bibitem[{Yu et~al(2019)Yu, Lin, Yang, Shen, Lu, and Huang}]{DBLP:conf/iccv/YuLYSLH19}
Yu J, Lin Z, Yang J, et~al (2019) Free-form image inpainting with gated convolution. In: ICCV

\bibitem[{Yu et~al(2021)Yu, Zhan, Wu, Pan, Cui, Lu, Ma, Xie, and Miao}]{DBLP:conf/mm/YuZWPCLMXM21}
Yu Y, Zhan F, Wu R, et~al (2021) Diverse image inpainting with bidirectional and autoregressive transformers. In: {MM}

\bibitem[{Yu et~al(2022{\natexlab{a}})Yu, Du, Zhang, and Luo}]{yu2022unbiased}
Yu Y, Du D, Zhang L, et~al (2022{\natexlab{a}}) Unbiased multi-modality guidance for image inpainting. In: ECCV

\bibitem[{Yu et~al(2022{\natexlab{b}})Yu, Zhang, Fan, and Luo}]{yu2022high}
Yu Y, Zhang L, Fan H, et~al (2022{\natexlab{b}}) High-fidelity image inpainting with gan inversion. In: ECCV

\bibitem[{Zeng et~al(2020)Zeng, Lin, Yang, Zhang, Shechtman, and Lu}]{zeng2020high}
Zeng Y, Lin Z, Yang J, et~al (2020) High-resolution image inpainting with iterative confidence feedback and guided upsampling. In: ECCV

\bibitem[{Zeng et~al(2021)Zeng, Lin, Lu, and Patel}]{zeng2021cr}
Zeng Y, Lin Z, Lu H, et~al (2021) Cr-fill: Generative image inpainting with auxiliary contextual reconstruction. In: ICCV

\bibitem[{Zeng et~al(2022)Zeng, Fu, Chao, and Guo}]{DBLP:journals/corr/abs-2104-01431}
Zeng Y, Fu J, Chao H, et~al (2022) Aggregated contextual transformations for high-resolution image inpainting. IEEE Transactions on Visualization and Computer Graphics

\bibitem[{Zhang et~al(2018)Zhang, Isola, Efros, Shechtman, and Wang}]{DBLP:conf/cvpr/ZhangIESW18}
Zhang R, Isola P, Efros AA, et~al (2018) The unreasonable effectiveness of deep features as a perceptual metric. In: CVPR

\bibitem[{Zhao et~al(2021)Zhao, Cui, Sheng, Dong, Liang, Chang, and Xu}]{DBLP:conf/iclr/ZhaoCSDLCX21}
Zhao S, Cui J, Sheng Y, et~al (2021) Large scale image completion via co-modulated generative adversarial networks. In: ICLR

\bibitem[{Zheng et~al(2019)Zheng, Cham, and Cai}]{DBLP:conf/cvpr/ZhengCC19}
Zheng C, Cham T, Cai J (2019) Pluralistic image completion. In: CVPR

\bibitem[{Zhou et~al(2017{\natexlab{a}})Zhou, Lapedriza, Khosla, Oliva, and Torralba}]{DBLP:journals/pami/ZhouLKO018}
Zhou B, Lapedriza A, Khosla A, et~al (2017{\natexlab{a}}) Places: A 10 million image database for scene recognition. IEEE Transactions on Pattern Analysis and Machine Intelligence 40(6):1452--1464

\bibitem[{Zhou et~al(2017{\natexlab{b}})Zhou, Zhao, Puig, Fidler, Barriuso, and Torralba}]{zhou2017scene}
Zhou B, Zhao H, Puig X, et~al (2017{\natexlab{b}}) Scene parsing through ade20k dataset. In: CVPR

\bibitem[{Zhu et~al(2016)Zhu, Kr{\"{a}}henb{\"{u}}hl, Shechtman, and Efros}]{zhu16generative}
Zhu J, Kr{\"{a}}henb{\"{u}}hl P, Shechtman E, et~al (2016) Generative visual manipulation on the natural image manifold. In: ECCV

\bibitem[{Zhu et~al(2020)Zhu, Shen, Zhao, and Zhou}]{zhu2020domain}
Zhu J, Shen Y, Zhao D, et~al (2020) In-domain {GAN} inversion for real image editing. In: ECCV

\end{thebibliography}

\end{document}